\documentclass[twoside,11pt]{article}
\usepackage{jair, theapa, rawfonts}

\usepackage{natbib}
\usepackage{pgffor}
\NewDocumentCommand{\mycite}{m}{%
    \begingroup
    \def\citepunct{; }
    \def\last{}
    (
    \foreach \x in {#1} {%
        \ifx\last\empty
            \citeauthor{\x} \citeyear{\x}%
        \else
            \citepunct \citeauthor{\x} \citeyear{\x}%
        \fi
        \global\let\last\x
    }%
    )
    \endgroup
}

\usepackage{setspace}
\usepackage{sectsty}
\subsubsectionfont{\bfseries} 
\usepackage{xparse}
\usepackage{url}
\usepackage{graphicx}
\usepackage{multirow}
\usepackage{amsmath,amssymb,amsfonts}
\usepackage{amsthm}
\usepackage{mathrsfs}
\usepackage[title]{appendix}
\usepackage{xcolor}
\usepackage{textcomp}
\usepackage{manyfoot}
\usepackage{booktabs}
\usepackage{algorithm}
\usepackage{algpseudocode}
\usepackage{listings}
\usepackage{soul}  
\usepackage{comment}
\usepackage{tocloft}
\usepackage{dirtytalk}
\usepackage{bm}
\usepackage{color}
\usepackage{array}
\usepackage{graphicx}
\usepackage[center]{caption}
\usepackage{subcaption}
\usepackage{makecell}
\setcellgapes{2pt}
\newcommand{\specialcell}[2][c]{
\begin{tabular}[#1]{@{}c@{}}#2\end{tabular}}
\usepackage{adjustbox}
\usepackage{soul}
\setstcolor{red}
\usepackage{mdframed}
\usepackage{arydshln}
\usepackage{hyperref}
\usepackage{float}

\usepackage{tikz}

\ShortHeadings{Robust Deep Reinforcement Learning Through Adversarial Attacks and Training}
{Schott, Delas, Hajri, Gherbi, Yaich, Boulahia-Cuppens, Cuppens \& Lamprier}
\firstpageno{1}

\begin{document}

\title{Robust Deep Reinforcement Learning Through Adversarial Attacks and Training: A Survey}


\author{\name Lucas Schott \email lucas.schott@irt-systemx.fr \\ \addr Institut de Recherche Technologique SystemX \\ MLIA, ISIR, Sorbonne Université
\AND
        \name Joséphine Delas \email josephine.delas@polymtl.ca \\ \addr Polytechnique Montréal
\AND
        \name Hatem Hajri \email hatem.hajri@safrangroup.com \\ \addr Safran Tech
\AND
        \name Elies Gherbi \email elies.gherbi@irt-systemx.fr \\ \addr Institut de Recherche Technologique SystemX
\AND
        \name Reda Yaich \email reda.yaich@irt-systemx.fr \\ \addr Institut de Recherche Technologique SystemX
\AND
        \name Nora Boulahia-Cuppens \email nora.boulahia-cuppens@polymtl.ca \\ \addr Polytechnique Montréal
\AND
        \name Frederic Cuppens \email frederic.cuppens@polymtl.ca \\ \addr Polytechnique Montréal
\AND
        \name Sylvain Lamprier \email sylvain.lamprier@univ-angers.fr \\ \addr LERIA, Université d'Angers \\ MLIA, ISIR, Sorbonne Université
}

\maketitle


\begin{abstract}
Deep Reinforcement Learning (RL) is a subfield of Machine Learning (ML) that trains autonomous agents to take sequential actions in complex environments. Despite its significant performance in well-known environments, it remains susceptible to minor condition variations, raising concerns about its reliability in real-world applications. To improve usability, deep RL must demonstrate trustworthiness and robustness. A way to improve the robustness of deep RL agents to unknown changes in the environment conditions and possible perturbations is through Adversarial Training, by training the agent against well-suited adversarial attacks on observations and the environments dynamics. Addressing this critical issue, our work presents an in-depth analysis of contemporary adversarial attack and training methods, systematically categorizing them and comparing their objectives and operational mechanisms.
\end{abstract}

\setcounter{secnumdepth}{4} 

\setcounter{tocdepth}{1}

\newpage

{
\setstretch{0.5}
\tableofcontents
}

\section{Introduction}

    The advent of deep RL has marked a significant shift in various fields, including games \mycite{mnih2015human,silver2016mastering,vinyals2019grandmaster}, autonomous robotics \mycite{levine2016end}, autonomous driving \mycite{kiran2021deep}, and energy management \mycite{zhang2018review}. By combining reinforcement learning with neural networks, deep RL leverages high-dimensional observations and action spaces to learn effective neural policies for complex tasks without expert supervision. Throughout this paper, we use RL to refer to deep reinforcement learning unless otherwise specified. 

    However, while RL achieves remarkable performance in well-known controlled environments, it also struggles to maintain robust performance amid diverse condition changes and real-world perturbations. It particularly struggles to bridge the reality gap (or sim-to-real problem) \mycite{hofer2020perspectives,collins2019quantifying}: RL agents are usually trained in simulation, which remains an imitation of the real world, resulting in a gap between the performance of a trained agent in the simulation and its performance once transferred to the real-world application. This poses the problem of robustness facing distribution shift, which refers to the agent's ability to maintain performance in deployment despite changes in the environment dynamics.
    
    Moreover, the emergence of adversarial attacks that generate perturbations in the inputs and in the environment dynamics, which are deliberately designed to mislead neural network decisions, poses unique challenges in RL \mycite{chen2019adversarial,ilahi2021challenges,moos2022robust}. But it can also serve as an opportunity to improve robustness by applying adversarial training to agents that aim to maintain good performance on the task at hand despite powerful alterations to their environment.
    
    This survey aims to review methods in this critical area by presenting a comprehensive framework for understanding the concept of robustness of RL agents. It covers both robustness to perturbed inputs and robustness to altered environment dynamics. Additionally, it introduces a new classification system that organizes the different types of perturbations affecting robustness into a unified model. It also provides a review of the existing literature on adversarial methods for robust RL agents and classifies these methods according to the proposed taxonomy. The goal is to provide a deeper understanding of various adversarial techniques, including their strengths, limitations, and impacts on the performance, robustness, and generalization capabilities of RL agents.
    
    The key contributions of this work include:
    \begin{itemize}
        \item Unifying the various formulations of adversarial learning for RL in a general framework.
        \item Developing a taxonomy and classification for adversarial attacks in RL.
        \item Reviewing existing adversarial attacks, characterized using our proposed taxonomy.
        \item Reviewing how adversarial attacks can be used to improve the robustness of RL agents.
    \end{itemize}
    
\begin{figure*}[!ht]
        \centering
        \includegraphics[width=\textwidth]{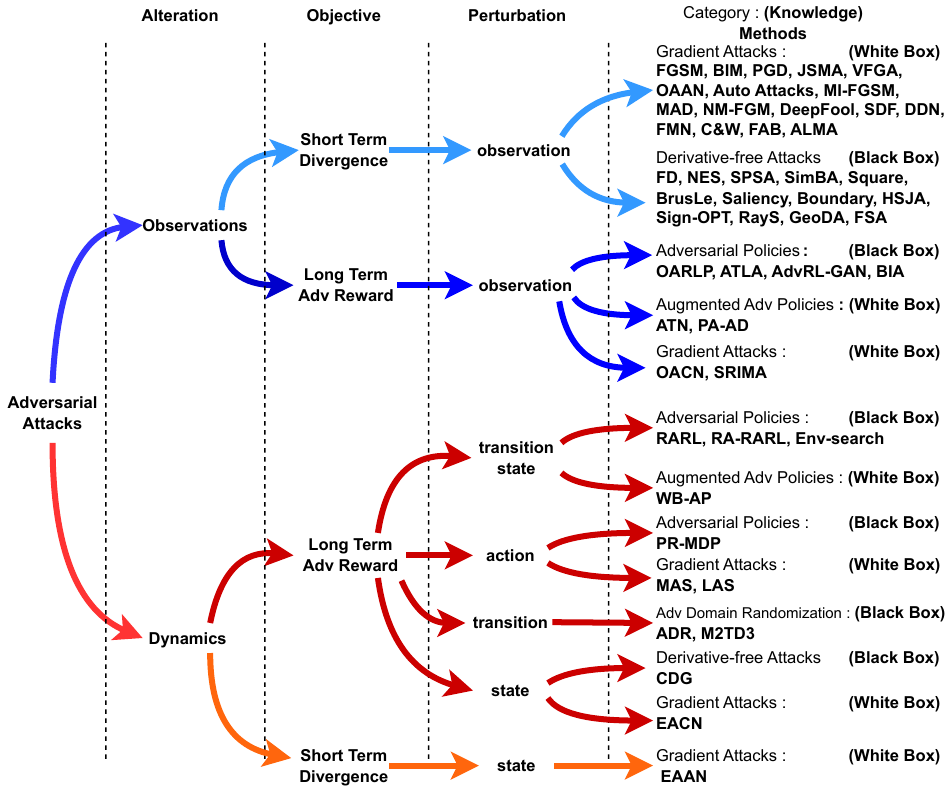}
        \caption{Categorization of the adversarial attacks of the literature as described in Section \ref{sec:adv_attacks} with the taxonomy introduced in Section \ref{sec:taxonomy} of this survey.}
        \label{fig:categorization}
    \end{figure*}

    The structure of the survey is organized as follows:
    \begin{itemize}
        \item Section \ref{sec:background} introduces the fundamentals of RL, neural networks, and Deep RL. It also discusses the importance of robustness of neural networks and outlines the formal prerequisites for analyzing robustness in RL.
        \item Section \ref{sec:formalization} introduces a formalization of the notion of adversarial robustness in RL.
        \item Section \ref{sec:taxonomy} presents a taxonomy for categorizing adversarial attack methods as shown in Figure \ref{fig:categorization}.
        \item Section \ref{sec:adv_attacks} gathers the different adversarial attack techniques on RL agents existing in the literature.
        \item Section \ref{sec:adv_training} focuses on the adversarial training techniques that can be employed to improve the robustness of RL agents.
        \item Section \ref{sec:tools} provides an overview of tools and libraries commonly used for developing and testing adversarial robustness in RL.
        \item Section \ref{sec:discussion} discusses further steps for robustness in RL and Section \ref{sec:conclusion} concludes the survey.
    \end{itemize}

\section{Background}
    \label{sec:background}

    \subsection{Reinforcement Learning}
    
        RL is a training framework in the Machine Learning (ML) family for sequential decision-making agents that interact with an environment. Agents take actions in the environment and receive numerical rewards as feedback for how well those actions comply with the task at hand. The objective of an RL agent is to learn a policy, a mapping from states to actions, which maximizes the expected cumulative reward through time.
        
        \subsubsection{Partially Observable Markov Decision Process}

            A Markov Decision Process (MDP) 
            is a mathematical framework for modeling decision-making problems where an agent interacts with an environment over discrete time steps. In most real-world applications, the agent may not have access to the environment's complete state and instead receives partial observations. This scenario is known as a Partially Observable Markov Decision Process (POMDP)
            , which is a generalization of the MDP framework, represented by the tuple \mbox{$\Omega = (S, A, T, R, X, O)$}, where:
            \begin{itemize}
                \item $S$ is the set of states in the environment,
                \item $A$ is the set of actions available to the agent,
                \item $T : S \times A \times S \rightarrow [0,1]$ is the stochastic transition function, with $T(s_+|s,a)$ denoting the probability of transitioning to state $s_+$ given state $s$ and action $a$,
                \item $R : S \times A \times S \rightarrow \mathbb{R}$ is the reward function. $R(s,a,s_+)$ is received by the agent for taking action $a$ in state $s$ and moving to state $s_+$,
                \item $X$ is the set of observations as perceived by the agent,
                \item $O : S \times X \rightarrow [0,1]$ is the observation function, with $O(x|s)$ denoting the probability of observing $x$ given state $s$.
            \end{itemize}
            A step in the environment represented by the POMDP $\Omega$ is represented by the transition $(s_t,x_t,a_t,s_{t+1})$, where $s_t$ stands for the state, $x_t$ the observation of this state, $a_t$ the action applied by the agent, $s_{t+1}$ the next state after the transition. In this paper, we will use the POMDP framework as a general model, even though some environments could be described as MDPs.

        \subsubsection{Fundamentals of Reinforcement Learning}
    
            In RL, the goal is to learn a policy $\pi : S \to \Delta(A)$ where $\Delta(A)$ denotes the probability distribution over actions, and $\pi(a\mid s)$ denote the probability of selecting the action $a$ given state $s$, with $\sum_{a\in A}\pi(a\mid s)=1$. The optimal policy, denoted as $\pi^*$, therefore maximizes the expected cumulative discounted reward:
            \begin{equation}
                \pi^*=\arg\max_{\pi} \mathbb{E}_{\tau \sim \pi^\Omega} [R(\tau)]
            \end{equation}
            with
            \begin{equation}
                R(\tau) = \sum_{t=0}^{|\tau|-1} \gamma^{t}R(s_t, a_t, s_{t+1})
            \end{equation}            
            where $\tau=\big((s_0,a_0),(s_1,a_1),...,(s_{|\tau|},\textunderscore{})\big)$ is sampled from the distribution \(\pi^\Omega\) of trajectories obtained by executing policy \(\pi\) in environment \(\Omega\). The discount factor \(\gamma\), ranging from 0 to 1, weights the importance of future rewards.
            
            An important criterion for defining optimality is the state value function, denoted as $V^\pi: S \rightarrow \mathbb{R}$. For a state $s$, the value $V^\pi(s)$ represents the expected cumulative discounted reward starting from $s$ and following the policy $\pi$ thereafter. This can be formally expressed as:
            \begin{equation}
                V^\pi(s) = \mathbb{E}_{\tau \sim \pi^\Omega}[R(\tau)|s_0 = s]
                \label{eq:v_function}
            \end{equation}
            
            It can be expressed recursively with the Bellman equation \mycite{bellman1957dynamic}:
            \begin{equation}
                V^\pi(s) = \; \sum_{a} \pi(a|s) \sum_{s_+} T(s_+|s,a) \Big( R(s,a,s_+) + \gamma V^\pi(s_+) \Big)
            \end{equation}  
            Finally, the state-action value function $Q^\pi : S \times A \rightarrow \mathbb{R}$ is used in many algorithms as an alternative to $V^\pi$. The Q-value of a state $s$ and action $a$ is the expected cumulative discounted reward, starting from $s$, taking $a$, and following $\pi$: 
            \begin{equation}
                Q^\pi(s,a) = \mathbb{E}_{\tau \sim \pi^\Omega}[R(\tau)|s_0 = s, a_0=a]
                \label{eq:q_function}
            \end{equation}
            It can be expressed recursively with the equation:
            \begin{equation}
                Q^\pi(s,a) = \; \sum_{s_{+}} T(s_+|s,a) \Big( R(s,a,s_+) + \gamma \sum_{a_+} \pi(a_+|s_+)\big[ Q^\pi(s_+,a_+) \big] \Big)
            \end{equation}
            
\begin{figure}[!ht]
                \centering
                \includegraphics[scale=1]{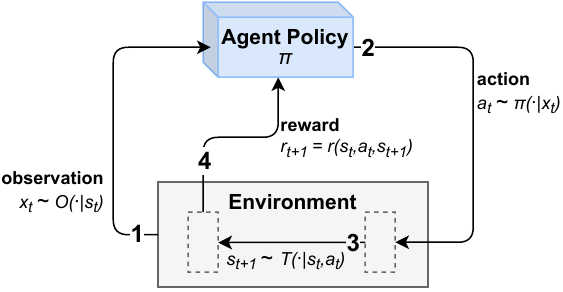}
                \caption{Flowchart of an agent with a policy function $\pi$ interacting with a POMDP environment}
                \label{fig:reinforcement_learning}
            \end{figure}

            In the POMDP setting, since states are not directly observable by agents, the practice is to base policies and value functions on the history of observations \mbox{(i.e., $x_{0:t}$ at step $t$)} in place of the actual state of the system (i.e., $s_t$). For ease of notation, we consider policies and value functions defined with only the last observation as input (i.e., $x_t$), while every approach presented below can be extended to methods leveraging full histories of observations. More specifically, we consider policies defined as $\pi: X \to \Delta(A)$ and action-value functions as $Q: X \times A \rightarrow \mathbb{R}$. Figure \ref{fig:reinforcement_learning} shows the flowchart of an agent with a policy function $\pi$ interacting with a POMDP environment.

    \subsection{Neural Networks and Deep Reinforcement Learning }
    
        To solve the complex task of RL problems with large input spaces and enable generalization, RL methods are usually combined with neural networks.
        
        \subsubsection{Deep Neural Networks}
            
            A neural network is a system of interconnected nodes (neurons) that process and transmit signals. Deep neural networks are models utilizing multiple layers of neurons, featuring varying degrees of architecture complexity, to analyze intricate data patterns.
            Training involves adjusting inter-neuron weight parameters to reduce errors (measured by a loss function) between the network’s predictions and actual outcomes, often employing Stochastic Gradient Descent (SGD) inspired algorithms. This training refines the network's ability to accurately recognize and respond to input data. The update rule of the parameters $\theta$ of the model $f_\theta$ in this context, given inputs $x$, labels $y$, learning rate $\alpha$ and loss function $\mathcal L$, is expressed as:
            \begin{equation}
                \theta = \theta - \alpha \cdot \nabla_{\theta} \mathcal L(f_{\theta}(x),y)
            \end{equation}

        \subsubsection{Deep Reinforcement Learning }
            
            Deep RL combines the principles of RL with the capabilities of neural networks.
            The central concept in Deep RL is to construct a policy $\pi$ using a neural network. This can be achieved either by approximating the Q-function (Equation \eqref{eq:q_function}) or the V-function (Equation \eqref{eq:v_function}), or by directly inferring the policy from experience. There are several popular Deep RL algorithms, each with its specific strengths and weaknesses. Approaches can be model-based or model-free, can be designed for particular contexts like discrete or continuous action spaces, or depend on the possibility to train the model on- or off-policy. The fundamental model-free Deep RL algorithms are PG \mycite{williams_simple_1992}, DQN \mycite{mnih2013playing} and DDPG \mycite{lillicrap2015continuous}, but the most effective contemporary algorithms are Rainbow \mycite{hessel2018rainbow}, PPO \mycite{schulman2017proximal}, SAC \mycite{haarnoja2018soft} or TQC \mycite{kuznetsov2020controlling} depending on the context of application. As noted in the introduction, we use \textbf{RL} throughout this paper to refer to deep reinforcement learning, unless otherwise specified.

    \subsection{Robustness Issues in RL}
        
        RL enables agents to learn complex behaviors by interacting with their environment. This poses unique security challenges, both due to interactions with the environment and to the agent's deep learning nature. The dynamic nature of RL, combined with the necessity for long-term strategic decision-making, exposes RL systems to a range of security threats that can compromise their learning process, decision-making integrity, and overall effectiveness. These challenges are further exacerbated by the adversarial landscape, where attackers can manipulate the environment or the agent's perception to induce faulty learning or decision-making. Addressing these challenges is crucial for deploying robust RL in security-sensitive applications.

        \subsubsection{Uncertainties in the Environment}
            
            RL systems, while powerful in optimizing decision-making processes, face significant challenges when dealing with unknown uncertainties in deployment environments. One of the primary robustness issues arises from the discrepancy between the training environment and the real-world conditions, often referred to as the \textit{reality gap} (or sim-to-real problem). This gap can lead to suboptimal or unsafe behavior when the agent encounters unanticipated situations that were not represented during training. Additionally, RL models are generally sensitive to the stochastic nature of real-world environments, where noise and dynamic changes can degrade performance. These uncertainties can severely impact the learning process, making the agent less reliable and predictable.

        \subsubsection{Adversarial Attacks of Neural Networks}
        \label{sec:dnn_attacks}
        
            Neural networks are a powerful tool now used in numerous real-world applications. However, their complex and highly non-linear structure makes them hard to control, raising growing concerns about their reliability. Adversarial ML recently emerged to exhibit vulnerabilities of neural networks by producing attacks on their inputs that modify outcomes. The National Institute of Standards and Technology (NIST)’s National Cybersecurity Center of Excellence (NCCE) \mycite{vassilev2024adversarial} and its European counterpart, ETSI \mycite{dahmen-lhuissier2022etsi}, provide terminologies and ontologies to frame the study of these adversarial methods.
            
            Neural networks are sensitive to minor perturbations in their vast input dimensions, leading to vulnerabilities. For example, in classification tasks, adversarial examples (slight, undetectable data alterations) deceive models into misclassification \mycite{yuan2019adversarial,vassilev2024adversarial,nguyen2015deep,sun2018survey}. The techniques that create these examples range from manual modifications \mycite{barreno2010security} to algorithm-generated perturbations.
            
            Usually the objective of adversarial attacks can be defined as described in Equation \eqref{eq:adv_ex_min_norm}: from an original instance $x$, find the nearest altered instance $x'$, according to a chosen metric $\|\cdot\|_p$ (e.g., $L_0$, $L_1$ or $L_\infty$ norms), that changes the output of the model $f_{\theta}$.
            \begin{equation}
                \min_{x'} \|x-x'\|_p \quad s.t. \quad f_{\theta}(x) \neq f_{\theta}(x')
                \label{eq:adv_ex_min_norm}
            \end{equation}
            
            Alternatively, it can be defined as described in Equation \eqref{eq:adv_ex_bounded_norm}: from an original instance $x$, find an altered instance $x'$ that changes the most the output of the model $f_{\theta}$ according to a metric $\mathcal{L}$ (e.g., Kullback-Leibler divergence) while satisfying the perturbation budget $\|x-x'\|_p \le \epsilon$.
            \begin{equation}
                \max_{x'} \mathcal{L}\big(f_{\theta}(x),f_{\theta}(x')\big)  \quad \text{s.t.} \quad \|x-x'\|_p \le \epsilon
                \label{eq:adv_ex_bounded_norm}
            \end{equation}

            In that context, when the model to attack is observable (white-box setting), gradient attacks are the greatest threat to neural networks. Such attacks exploit the knowledge of the model to discover the most impactful perturbations of the input given the attack objective. The most fundamental operation for these attacks is the computation of the gradient of a specified adversarial loss with respect to the input fed to the model.  
            The simplest gradient attack method, called  FGSM \mycite{goodfellow2014explaining},  relies on one single gradient step to craft a perturbation as follows:
            \begin{equation}
            \label{eq:supervised_grad}
                x' = x + \eta^{\xi} \quad \text{with} \quad  \eta^{\xi} \propto \nabla_x \mathcal{L}\big(f(x), y\big)
            \end{equation}
            where $f$ is the neural network to be attacked, $f(x)$ is the model output, $y$ is the ground truth label or a target decision, and $\mathcal{L}$ is an adversarial loss (e.g., cross entropy). In more advanced techniques, the basic gradient information $\nabla_x\mathcal L$ can be processed (signed, projected, accumulated, line-searched, etc.) to produce an update tailored to the chosen threat model.  These attacks have proven highly effective in practice across many pointwise decision-making tasks. For instance, the \citet{carlini2017towards} method can generate $L_2$-bounded perturbations that remain visually indistinguishable while bypassing many existing defense mechanisms.
                        
            Beyond pointwise classification, the vulnerability of neural networks to adversarial examples also extends to deep RL agents, posing critical risks in safety-sensitive applications such as autonomous driving and medical diagnosis. Initially developed for image classification, adversarial attacks have been shown to be equally effective against RL agents. For example, \citet{behzadan2017vulnerability} demonstrated that Deep Q-Networks (DQNs) are susceptible to such attacks, a finding later supported by \citet{huang2017adversarial}. The RL framework, being more flexible than supervised learning, exposes additional adversarial opportunities through its various components, further challenging the long-term security of RL systems.

            This survey examines adversarial attacks in RL and explores defense strategies to enhance agent robustness by simulating worst-case scenarios essential for Robust RL.

    \subsection{Enhancing Robustness in RL}

        There are several approaches to improve the robustness of RL agents, which we divide into three categories. First, Safe RL approaches aim at keeping the agent out of danger zones predefined by experts. Second, General Purpose Robust RL approaches try to improve robustness by leveraging different properties of neural networks. Third, Adversarial Specific RL approaches use adversarial attacks either to make the agent more robust or to detect and defend against attacks.

        \subsubsection{Safe RL}
        
            Formulating the challenge of safe control in RL \mycite{brunke2022safe} merges insights from both optimal control theory and RL, aiming to optimize a solution that balances task achievement with stringent safety standards in environments with uncertain dynamics. 

            \paragraph*{Safety Shields}\ \\
                The first approach to safe RL involves implementing hard or soft safety constraints that restrict the agent's actions under certain conditions. These are safety shields \mycite{garg2024learning,jansen2020safe,hsu2023safety} that block the action chosen by the agent if safety is violated, replacing it with a safe alternative.

            \paragraph*{Safety Constraint Loss}\ \\
                A second approach encourages learning safe behaviors by incorporating safety constraints into the optimization's loss function or reward system. These methods are based on the Constrained Markov Decision Process (CMDP) paradigm and often add a constraint safety loss to be minimized, in addition to the reward to be maximized \mycite{yang2021wcsac,bai2022achieving,wang2023enforcing}.
                        
            All these three approaches designed for safe RL require known uncertainties that can be easily measured, and for which constraints and solutions can be hard-coded. These techniques are suitable for improving robustness to the environment and interaction uncertainties in well-controlled environments.

        \subsubsection{General Purpose Robust RL}
        \label{sec:gen_purpose}

            General Purpose Robust RL explores methods to strengthen agents against perturbations and uncertainties without explicitly using adversarial attacks. This part emphasizes techniques like robust architectures, exploration strategies, randomized smoothing, and distillation to ensure consistent performance against perturbations and in varied and unpredictable environments.

            \paragraph*{Robust Architecture}\ \\
                The right network architecture can significantly improve the resilience of agent policies. \mycite{huang2021exploring} shows that larger models can be less robust, while reducing the capacity of deeper layers enhances robustness. Moreover, \cite{wierstra2007solving} and \cite{zhang2021robust} have shown that recurrent architectures improve resilience by limiting the impact of perturbations in single-timeframe observations since a sequence is used at each inference, and changes in environment dynamics can also be detected by such a recurrent architecture. \mycite{nie2024improve} employs a novel architecture for the policy network that incorporates global $L_\infty$ Lipschitz continuity and provides a convenient method to enhance policy robustness based on the output margin. In the broader vision domain, \mycite{bai2021transformers} demonstrates that transformers are not inherently more adversarially robust than CNNs. Still, they exhibit superior robustness to distribution shifts, suggesting architectural choices strongly influence robustness properties.

            \paragraph*{Defensive Distillation}\ \\
                Defensive Distillation is a technique initially proposed as a universal defense mechanism against potential adversarial attacks on neural networks. It is a process in which a smaller, simpler model (the student model) is trained to mimic the behavior of a larger, more complex model (the teacher model). The idea is that the student model learns not just the final predictions but also the \textit{soft} output probabilities of the teacher model. By training on soft targets, the student model learns smoother decision boundaries between classes. This makes it harder for small perturbations to push inputs across these boundaries \mycite{papernot2016distillation}. Defensive distillation has been applied to RL in \mycite{rusu2015policy,czarnecki2019distilling}. However, such an approach still leaves models vulnerable to effective gradient, black-box, and gradient-free attacks \mycite{carlini2016defensive}.
                
            \paragraph*{Noising Exploration}\ \\                
                To improve robustness, the agent may also enhance regularization via noisy training rewards \mycite{kumar2019enhancing,wang2020reinforcement}. Or increase exploration with noisy actions \mycite{hollenstein2022action}. Also \mycite{eysenbach2022maximum} theoretically proves that Maximum Entropy Reinforcement Learning inherently solves a class of robust RL problems by encouraging stochastic policies that perform well under a range of possible dynamics perturbations. These approaches are suitable to improve robustness to smooth environments and interaction uncertainties, but are not usually suited for preparing defense against specific attacks of abrupt dynamics alterations.

            \paragraph*{Ensemble of Agents}\ \\
                Ensembling in RL refers to combining multiple policies (or value functions) to improve robustness and overall performance. By aggregating decisions from several agents, ensembling mitigates weaknesses of individual learners—overfitting to specific environments or attacks—and reduces estimation variance. For example, Ensemble Proximal Policy Optimization (EPPO) learns a set of diverse policies and aggregates them for stronger generalization in high-variance settings \mycite{yang2022towards}. Classic value-based ensembles such as Bootstrapped DQN and Averaged-DQN reduce uncertainty and stabilize learning by maintaining multiple heads/networks and aggregating their predictions \mycite{osband2016deep,anschel2017averaged}. In continuous control, REDQ maintains an ensemble of Q-functions and uses randomized in-target minimization to curb overestimation while enabling a high update-to-data ratio, yielding robust, data-efficient learning \mycite{chen2021randomized}. Beyond variance reduction, ensembles can serve as defenses; they can be formed from different random seeds, training curricula, and then combined by majority vote, weighted averaging, or a meta-controller; surveys of ensemble RL summarize these design choices and their benefits for exploration, robustness, and transfer \mycite{yang2023survey}.
    
            \paragraph*{Domain Randomization}\ \\
                Domain Randomization (DR) is a widely used technique to improve the robustness and transferability of reinforcement learning policies. In DR, environment parameters such as friction, mass, or sensor noise are uniformly randomized during training, enabling the agent to learn behaviors that remain consistent across a wide range of dynamic variations~\mycite{tobin2017domain,peng2018sim}. By exposing the policy to diverse conditions, DR promotes invariance to environment-specific features and mitigates overfitting to the training simulator.
                Assuming potential interactions with the target real-world environment, \cite{du2021auto} extends this approach with Auto-Tuned Sim-to-Real, which automatically adjusts the randomization distributions to better align with realistic observations. This method belongs to the class of model-based approaches that leverage learned models to refine environment parameterization, a topic that lies beyond the scope of this survey.
            
            \paragraph*{Randomized Smoothing}\ \\
                Randomized Smoothing \mycite{cohen2019certified} is a technique that, applied to RL \mycite{kumar2022policy}, adds a high amount of different noises to an observation to create an ensemble of noisy observations centered on the initial one. Then all these observations are given to the agent, and the action selected is an aggregation of the outputs of the agent for all the noisy observations. This enables the smooth output of the policy of the agent, improving its robustness to adversarial attacks of neural networks.

        \subsubsection{Robust RL through Adversarial Attacks}

            Adversarial RL focuses on enhancing the robustness of RL agents by incorporating adversarial perturbations into training, either to detect the perturbations or to be directly robust to them. This approach improves resilience to observation and environment dynamics perturbations, directly addressing the challenges posed by hostile and uncertain conditions.

            \paragraph*{Adversarial Detection}\ \\
                Adversarial detection aims at identifying modified observations for removal. Modifications of inputs that can arise from attacks or simply sensor faults can indeed be highly problematic for critical settings. Their early detection is of crucial importance. Adversarial detection methods were first developed for supervised learning \mycite{metzen2017detecting,pang2018towards}, and then adapted to RL, through successor representation \mycite{lin2017detecting}, using a separately trained model \mycite{hickling2023robust}, or via local quadratic approximation of the deep neural policy loss \mycite{korkmaz2023detecting}. The goal of adversarial detection is to improve robustness to adversarial attacks of neural networks by detecting and skipping perturbed samples. It can be used in an environment where observations are sent continuously to the agent, and an action is not strictly required, given each observation \mycite{carlini2017adversarial}.

            \paragraph*{Adversarial Training}\ \\
                Adversarial Training incorporates perturbations into the training process to robustify agents against potential adversarial attacks or real-world uncertainties \mycite{moos2022robust}. It follows the principles of robust control \mycite{dorato1987historical,morimoto2005robust} by employing a min-max optimization strategy, essentially preparing the system to handle the worst-case scenario efficiently:
                \begin{equation}
                    \min_{\pi} \max_{\delta \in\Delta} J(\pi,\delta)
                \end{equation}
                where $J(\pi,\delta)$ represents the expected cost for policy $\pi$ under perturbations $\delta$ within the set $\Delta$.
                
                By training with adversarially altered experiences, agents learn to uphold performance despite manipulations in inputs or environment dynamics. This methodology ensures that the control system remains effective and reliable even when faced with unpredictable changes or adverse conditions, thereby enhancing its robustness and resilience in uncertain environments. This approach is suitable for improving robustness to environment and interaction uncertainties as well as adversarial attacks on neural networks.\\

        There is a wide spectrum of defense methods for RL algorithms, each with its benefits and limitations. Combining several methods allows one to cover different aspects of the adversarial threat, but the defendant must keep in mind that simply stacking defense layers does not necessarily improve robustness \mycite{he2017adversarial}: each method must be analyzed and selected with care according to the given context.
        
        In this survey, we focus on adversarial training as a method for improving robustness. It is a general framework that focuses specifically on improving robustness, and it can be used both as a defense against adversarial attacks as well as against environment and interaction uncertainties. Leveraging adversarial training for more robust and reliable RL algorithms is the most used defense against adversarial attacks, and the variety of methods available fits each specific use case.

\subsection{Motivating Applications}
\label{sec:motivating_applications}

    Beyond their theoretical interest, adversarial attacks and robust training in RL are directly motivated by deployment scenarios where reliability under perturbation is a safety-critical requirement. We briefly survey key domains to ground the rest of this survey.

    \paragraph*{Autonomous Driving}\ \\
%
        Autonomous driving is one of the most compelling settings for adversarial robustness in RL. Perception modules can be fooled by physical adversarial attacks, such as adversarial patches applied to road signs \mycite{eykholt2018robust}, or by sensor noise and camera occlusions that shift the observation distribution at test time. Beyond perception, the driving policy must handle dynamic and unpredictable behaviors from other vehicles, pedestrians, or road conditions that were not encountered during training. Adversarial training, by exposing the agent to worst-case perturbations on observations and dynamics during training, directly addresses the sim-to-real gap that is a central challenge in deploying autonomous systems \mycite{kiran2021deep}. In safety-critical settings where a single failure can have severe consequences, robust RL methods offer a principled approach to certify that a policy performs acceptably even under adversarial or unexpected conditions.


        \paragraph*{Robotics and Sim-to-Real Transfer}\ \\ Robotic control policies are typically trained in simulation and then deployed on physical hardware. The discrepancy between simulated and real dynamics, variations in friction, mass, actuator response, or contact models, can be modeled as perturbations on the environment dynamics. Robust RL methods, particularly those based on adversarial dynamics perturbations \mycite{peng2018sim, pinto2017robust}, directly target this challenge by training policies against adversarially chosen dynamic parameters within a controlled uncertainty set. This produces policies that generalize to the physical world without costly re-training on the real system, which is especially valuable when real-world data collection is expensive or dangerous.

\paragraph*{Cyber-Physical and Security-Critical Systems}\ \\
        In cyber-physical systems and network security, RL agents are increasingly considered for tasks such as intrusion detection, automated penetration testing, and resource allocation under adversarial conditions \mycite{nguyen2021deep}. In these environments, adversaries actively probe and manipulate the system, making worst-case robustness a primary design objective rather than an afterthought. Adversarial training in RL is well-suited to this context since it explicitly optimizes policies against strategic adversaries, mirroring the red-team/blue-team dynamic of real security operations \mycite{apruzzese2020deep}. The MARL perspective on adversarial robustness \mycite{standen2025adversarial} is particularly relevant here, as network defenders and attackers are naturally modeled as competing agents.


\paragraph*{Energy Management and Industrial Control}\ \\
RL has demonstrated strong potential for optimizing energy consumption 
    in smart grids and industrial processes \mycite{zhang2018review,cao2025deep}. 
    These systems face multiple adversarial threats: sensor readings can be 
    falsified to destabilize grid frequency control \mycite{sun2026cyber}, 
    reward signals can be poisoned in energy pricing mechanisms 
    \mycite{gunn2022adversarial}, energy uncertainty models  can be subject to distributional shifts on dynamics \mycite{yaseen2026resilient} and RL-based grid controllers, despite 
    achieving state-of-the-art performance in operational challenges such as 
    the L2RPN benchmark, have been shown to be vulnerable to adversarial 
    attacks, with adversarial training offering an effective defense 
    \mycite{pan2021improving,pan2022characterizing,omnes2021adversarial}. More broadly, in 
    cyber-physical industrial control systems, adversarial perturbations 
    on actuator commands represent a realistic and critical threat that 
    adversarial training can directly address \mycite{tan2020robustifying}. \\


        Across these domains, a common thread emerges: the deployment environment is always more uncertain and more adversarial than the training environment. Adversarial training in RL provides a principled and theoretically grounded framework to close this gap, making it a central tool for building reliable autonomous agents in the real world.
    In the next section, we introduce the formalization of Robust RL via adversarial training as a framework for improving the robustness of agents to new elements of uncertainty and adversarial attacks.


\section{Formalization and Scope}

    \label{sec:formalization}
    
    The aim of this section is to unify the various formulations of adversarial robust learning for RL in a general framework.

    \subsection{The problem of Robustness in RL}
        
        \label{sec:formalization_problem}

        Generally speaking, we are interested in the following optimization problem: 
        \begin{equation}
            \pi^*=\arg\max_{\pi} \mathbb{E}_{\Omega \sim \Phi(\cdot|\pi)} \mathbb{E}_{\tau \sim \pi^\Omega} [R(\tau)]
        \end{equation}
        where $\Phi$ corresponds to the distribution of environments to which the agent is likely to be confronted when deployed (whether it adversarially considers $\pi$ or not at test time), $\pi^\Omega$ is the distribution of trajectories using the policy $\pi$ and the dynamics from $\Omega$, and $R(\tau)$ is the cumulative reward collected in $\tau$. While this formulation suggests meta-RL, in this setting $\Phi(\Omega|\pi)$ is unknown at train time. The training setup is composed of a unique MDP on which the policy can be learned, which is usually the case for many applications. 
        
        Given a unique training POMDP $\Omega$, the problem of robustness we are interested in can be reformulated by means of an alteration distribution  $\Phi(\phi|\pi)$:
        \begin{equation}
            \pi^*=\arg\max_{\pi}\mathbb{E}_{\phi \sim \Phi(\cdot|\pi)} \mathbb{E}_{\tau \sim \pi^{\phi,{\Omega}}} [R(\tau)]
        \end{equation}
        where $\pi^{\phi,\Omega}$ is the distribution of trajectories using policy $\pi$ on $\phi(\Omega)$, standing as the MDP $\Omega$ altered by $\phi$. Generally speaking, we can set $\phi$ as a function that can alter any component of $\Omega$ as $\phi(\Omega)= (\phi_S(S), \phi_A(A), \phi_T(T), \phi_R(R), \phi_X(X), \phi_O(O))$. While $\phi$ can simultaneously affect any of these components, we particularly focus on two crucial components for robustness:
        \begin{itemize}  
            \item Observation alterations: $\phi_O$ denotes alterations of the observation function of $\Omega$. In the corresponding altered environment $\widetilde{\Omega}=(S,A,T,R,X,\phi_O(O))$, the observation obtained from a state $s \in S$ could differ from that in $\Omega$. This can result from an adversarial attacker that perturbs signals from sensors to induce failures, observation abilities from the real world that might be different than those in simulation, or even unexpected failures of some sensors. These perturbations only induce perception alterations for $\pi$, without any effect on the true internal state of the system in the environment. Occurring at a specific step $t$ of a trajectory $\tau$, such an alteration thus only impacts the future of $\tau$ if it induces changes in the policy decision at $t$.
            \item Dynamics alterations: $\phi_T$ denotes alterations of the transition function of $\Omega$. In the corresponding altered environment $\widetilde{\Omega}=(S,A,\phi_T(T),R,X,O)$, dynamics are modified, such that actions do not have exactly the same effect as in $\Omega$. This can result from an adversarial attacker that modifies components of the environment to induce failures, from real-world physics, which might be different than those in the training simulator, or from external events that can incur unexpected situations. Dynamics alterations act on trajectories by modifying the resulting state $s_{t+1}$ emitted by the transition function $T$ at any step $t$. Even when localized at a single specific step $t$ of a trajectory, they thus impact its whole future.
        \end{itemize}
        In this work, we do not explicitly address variations of other components \mbox{($S$, $A$, $R$ and $X$)}, as they usually pertain to different problem areas. $\phi_S$ (resp. $\phi_A$) denotes alterations of the state (resp. action) set, where states (resp. actions) can be removed or introduced in $S$ (resp. $A$). $\phi_X$ denotes alterations of the observation support $X$. While some perturbations of dynamics $\phi_T$ or observations $\phi_O$ can lead the agent to reach new states or observations never considered during training (which corresponds to an implicit $\phi_S$ or $\phi_X$ perturbation), $\phi_S$, $\phi_A$, and $\phi_X$ all correspond to support shifts, related to static Out-Of-Domain issues, which we do not specifically focus on in this work. $\phi_R$ denotes alterations of the reward function $R$, which does not pose the problem of robustness in usage, since the reward function is only used during training.

    \subsection{Adversarial Attacks for Robust RL}

        \label{sec:formalization_dro_adv_training}

        Following distributionally robust optimization (DRO) principles \mycite{rahimian2019distributionally,queeney2023risk}, unknown distribution shifts can be anticipated by considering worst-case settings in some uncertainty sets ${\cal R}$. In our robust RL setting, this reduces to the following max-min optimization problem:
\begin{equation}
    \pi^*=\arg\max_{\pi} \min_{\widetilde{\Phi} \in {\cal R}}  \mathbb{E}_{\phi \sim \widetilde{\Phi}(\cdot|\pi)} \mathbb{E}_{\tau \sim \pi^{\phi,{\Omega}}} [R(\tau)]
    \label{eq:dro}
\end{equation}
where ${\cal R}$ is a set of perturbation distributions. As well-known in DRO literature for supervised regression problems, the shape of ${\cal R}$ has a strong impact on the corresponding optimal decision system. In our RL setting, increasing the level of disparities allowed by the set ${\cal R}$ constrains the resulting policy $\pi$ to have to perform simultaneously over a broader spectrum of environmental conditions. 

A common structural assumption on uncertainty sets is \emph{state–action rectangularity}. An uncertainty set ${\cal R}$ is SA-rectangular when it factorizes independently over all state–action pairs, i.e.,
\[
{\cal R}=\prod_{s\in S}\prod_{a\in A(s)} {\cal R}_{s,a},
\]
meaning that adversarial perturbations for each $(s,a)$ can be chosen independently from the others. 
This assumption is often adopted to ensure tractability of the worst-case optimization. 
For example, the seminal Robust Value Iteration algorithms of \cite{iyengar2005robust,nilim2005robust,wiesemann2013robust} rely on SA-rectangular uncertainty sets to obtain a contractive Bellman operator and a solvable dynamic program in tabular settings.

While this enables better generalization for environmental shifts, it also implies dealing with various highly unrealistic scenarios if the set $\cal R$ is not restricted on reasonable levels of perturbations. With extremely large sets ${\cal R}$, the policy $\pi$ is expected to be equally effective for any possible environment; eventually converging to a trivial uniform policy, that allocates equal probability to every action for any state from $S$. The shape of $\cal R$ has thus to be controlled to find an accurate trade-off between generalization and effectiveness \mycite{iyengar2005robust, nilim2005robust}.

Dealing with worst-case distributions of perturbations defined over full supports of $\Omega$ is highly intractable in most realistic applications. In this survey, we rather focus on adversarial training that leverages the simultaneous optimization of an attacker agent $\xi$, that produces perturbations for situations reached by the protagonist agent $\pi$, by acting on the adversarial actions $A^{\xi,\Omega}$ that the environment $\Omega$ permits:
        \begin{align}
        \label{obj_robusttl}
            \pi^* = \arg\max_\pi
            \mathbb{E}_{\tau \sim \pi^{\xi^*,\Omega}}[R(\tau)] \nonumber \\
            s.t.\qquad \xi^*=\arg\min_\xi  \Delta^{\pi,\Omega}(\xi)
        \end{align}
        where $\Delta^{\pi,\Omega}(\xi)$ stands as the optimization objective of the adversarial agent given $\pi$ and the training environment $\Omega$, which ranges from adverse reward functions to divergence metrics (c.f., Section \ref{sec:objective}), and $\pi^{\xi,\Omega}(\tau)$ corresponds to the probability of a trajectory following policy $\pi$ in a POMDP dynamically modified by an adversarial agent $\xi$, given a set of actions $A^\xi = (A^{\xi,X},A^{\xi,A},A^{\xi,S},A^{\xi,T},A^{\xi,S+})$. The action \mbox{$a^\xi_t=(a^{\xi,X}_t,a^{\xi,A}_t,a^{\xi,S}_t,a^{\xi,T}_t,a^{\xi,S+}_t)$} of adversary $\xi$ can target any element of any transition $\tau_t=(s_t,x_t,a_t,s_{t+1})$ of trajectories in $\Omega$. While any perturbation of $x_t$ induces an alteration of the observation function $O$, any perturbation of $s_t$, $a_t$ or $s_{t+1}$ induces an alteration of the transition function $T$ (either directly, through its internal dynamics or indirectly via the modification of its inputs or outputs).
        
        In this setting, any trajectory is composed as a sequence of adversary-augmented transitions $\widetilde{\tau}_t=(s_t,x_t,a_t,a^\xi_t,x'_t,a'_t,\widetilde{s}_t,\widetilde{x}_t,\widetilde{s}_{t+1},\widetilde{x}_{t+1},s_{t+1})$, where the elements $x'_{t}$ (resp. $a'_{t}$) stands for the perturbed observation (resp. action) produced by the application of the adversary action $a^{\xi,X}_t$ (resp. $a^{\xi,A}_t$) at step $t$. $\widetilde{s}_t$ (resp. $\widetilde{s}_{t+1}$) stands for the intermediary state produced by the application of the adversary action $a^{\xi,S}_t$ (resp. $a^{\xi,T}_t$) at step $t$ before (resp. during) the transition function, and $\widetilde{x}_t$ (resp. $\widetilde{x}_{t+1}$) is the observation of this state. Finally, $s_{t+1}$ stands for the final next state produced by the application of the adversary action $a^{\xi,S+}_t$ after the transition function, and its observation is $x_{t+1}$. The support and scope of adversarial actions define the level of perturbations allowed in the corresponding uncertainty set ${\cal R}$ from Equation \eqref{eq:dro}, with impacts on the generalization/accuracy trade-off of the resulting policy $\pi$. While the protagonist agent $\pi$ acts from $x_t$ with $a_t \sim \pi(\cdot|x_t)$, in the following, we consider the general case of adversaries $\xi$ that act from $s_t$, $x_t$ and $a_t$, that is $\xi: A^\xi \times S \times X \times A \rightarrow [0,1]$ where $a^\xi_t \sim \xi(\cdot|s_t,x_t,a_t)$. By doing this we consider adversaries $\xi$ that have full knowledge of the environment state, observation, and action, although this could easily be limited to adversarial policies $\xi$ that act only from partial information.

\section{Taxonomy of Adversarial Attacks in RL}
    \label{sec:taxonomy}
    
    We conduct a systematic analysis of adversarial attacks for RL agents, with a focus on their purposes and applications. To better grasp the variety of methods available, together with their specificities, we propose a taxonomy of adversarial attacks for RL. This taxonomy is used to categorize the adversarial attack as previously shown in Figure \ref{fig:categorization} and later described in Table \ref{tab:adv_attacks}. This section discusses the different components of adversarial approaches for robust RL, before developing the main approaches in the next section.
    
    Here and in the following, we differentiate between \textit{perturb} and \textit{alter}. The term perturb refers to modifying an element within the transition tuple, such as an observation, state, or action. Conversely, alter is used to describe changes to a component of the agent's POMDP. For instance, an adversarial attack that perturbs the observations results in an alteration of the observation function $O$ in the agent's POMDP $\Omega$. Similarly, any adversarial attack that perturbs actions, states, or adds an adversarial action in the transition functions, constitutes an alteration of the transition function $T$ in the POMDP $\Omega$ (which defines the environment dynamics) of the agent.

     \subsection{Perturbed Element}
        
        \label{sec:perturbation}
        
        An adversarial attack is a method that uses an adversarial action $a^\xi_t \in A^\xi$ emitted by the adversary agent $\xi$ at step $t$, to produce a perturbation in the simulation during the trajectory of an agent. Given the type of attack, an action $a^\xi_t$ can directly perturb different elements:

        \begin{itemize}
        
            \item \textbf{The observations $x_t$:} Via a perturbation function $\Psi^X : X \times X \times A^{\xi,X} \rightarrow [0,1]$, where $x'_t \sim \Psi^X(\cdot| x_t,a^{\xi,X}_t)$.
            
            \item \textbf{The actions $a_t$:} Via a perturbation function $\Psi^A : A \times A \times A^{\xi,A} \rightarrow [0,1]$ where $a'_t \sim \Psi^A(\cdot|a_t,a^{\xi,A}_t)$.
            
            \item \textbf{The current state $s_t$ (before transition):} Via an additional transition function $T_{\Psi}^S : S \times S \times A^{\xi,S} \rightarrow [0,1]$ where \mbox{$\widetilde{s}_t \sim T_\Psi^S(\cdot|s_t,a^{\xi,S}_t)$} is applied before the main transition function of the environment is applied, so on the current state $s_t$, but after the decision $a_t$ of the agent is taken.
            
            \item \textbf{The transition function $T$:} Via an adversarially augmented transition $T_\Psi : S \times S \times A \times A^{\xi,T} \rightarrow [0,1]$ where $\widetilde{s}_{t+1} \sim T_\Psi(\cdot|s_t,a_t,a^{\xi,T}_t)$ is applied as a substitute of the main transition function of the environment $T$.
            
            \item \textbf{The next state $s_{t+1}$ (after transition):} Via an additional transition function $T_\Psi^{S+} : S \times S \times A^{\xi,S+} \rightarrow [0,1]$ where $s_{t+1} \sim T_\Psi^{S+}(\cdot|s_{t+1},a^{\xi,S+}_t)$ is applied after the main transition function of the environment is applied, so on the next state $s_{t+1}$, but before the next decision $a_{t+1}$ of the agent is taken.
            
        \end{itemize}
            
        The perturbations on the first two types of elements (observation and action) require just the modification of a vector that will be fed as input to another function, so they are easy to implement in any environment. The perturbations on the last three types of elements (state, transition function, and next state) are more complex and require modifying the environment itself, either by being able to modify the state with an additional transition function or being able to modify the main transition function itself by incorporating the effect of the adversary action.

     \subsection{Altered POMDP Component}
        
        \label{sec:altered_component}

        Following the two main types of alterations $\phi$ that are discussed in Section \ref{sec:formalization}, the main axis of the taxonomy of approaches concerns the impact on the POMDP of actions that adversary agents emit during training of $\pi$. Given the adversarial elements defined in the previous section, we specify each possible perturbation independently to discuss each specific adversarial impact on the POMDP.

        \subsubsection{Alteration of the Observation Function $O$}
            
            \label{sec:altered_observation}

            The first type of component alteration is the alteration of the observation function $O$ of the POMDP $\Omega$. Directly inspired by adversarial attacks in supervised machine learning, many methods are designed to modify the inputs that are perceived by the protagonist agent $\pi$. The principle is to modify the input vector of an agent, which can correspond, for instance, to the outputs of a sensor of a physical agent, like an autonomous vehicle. The observation is perturbed before the agent takes any decision, so that the agent gets the perturbed observation and can be fooled.
            
            More formally, in the setting of an observation attack, the adversary $\xi$ acts to produce a perturbed observation $x'_t$ before it is fed as input to $\pi$, via the specific perturbation function $\Psi^X(x'_t|x_t,a^{\xi,X}_t)$ applied to the observation $x_t$.
            
            In that case, $\xi$ can be regarded as an adversary agent that acts by emitting adversarial actions $a^{\xi,X}_t \sim \xi(\cdot|s_t,x_t)$ with $a^{\xi,X}_t \in A^{\xi,X}$ in a POMDP given $\pi$ defined as\\ \mbox{$\Omega^{\pi}=(S,A^{\xi,X},T^\pi,R_\xi,X,O)$}. Here $T^\pi(s_{t+1}|s_t,a^{\xi,X}_t)$ is defined as the transition function of the environment, considering $\pi$, from the adversary's perspective. Consequently, the policy $\pi$ is incorporated into the environment dynamics as observed by the adversary. Sampling $s_{t+1} \sim T^\pi(\cdot|s_t,a^{\xi,X}_t)$ is performed as described in Algorithm \ref{algo:obs_adversary_pov}.
            
            \vspace{5pt}
            \begin{algorithm}[H]
                \caption{Transition $T^\pi$, using the observation perturbation function $\Psi^X$, in the POMDP $\Omega^{\pi}$ of the adversary.}
                \label{algo:obs_adversary_pov}
                \begin{algorithmic}[1]
                    \State sample $x_t \sim O(\cdot|s_t)$                  \Comment{observation}
                    \State sample $x'_t \sim \Psi^X(\cdot|x_t,a^{\xi,X}_t)$ \Comment{perturbed observation}
                    \State sample $a_t \sim \pi(\cdot|x'_t)$                \Comment{agent action}
                    \State sample $s_{t+1} \sim T(\cdot|s_t, a_t)$          \Comment{next state after transition}
                \end{algorithmic}
            \end{algorithm}
            \vspace{5pt}

            Reversely, agent $\pi$ acts on an altered POMDP $\Omega^{\xi}=(S,A,T,R,X,O^\xi)$. Here $O^\xi(x_t|s_t)$ is defined as the observation function of the environment, considering $\xi$ from the agent's perspective. Consequently, the adversary $\xi$ is incorporated into the observation function as observed by the agent. Sampling $x'_t \sim O^\xi(\cdot|s_t)$ is performed as described in Algorithm \ref{algo:obs_agent_pov}.

            \vspace{5pt}
            \begin{algorithm}[H]
                \caption{Observation $O^\xi$, using a observation perturbation function $\Psi^X$, in the POMDP $\Omega^{\xi}$ of the agent. Also illustrated in Figure \ref{fig:observation_perturbation}.}
                \label{algo:obs_agent_pov}
                \begin{algorithmic}[1]
                    \State sample $x_t \sim O(\cdot|s_t)$                   \Comment{observation}
                    \State sample $a^{\xi,X}_t \sim \xi(\cdot|s_t,x_t)$     \Comment{adversary action}
                    \State sample $x'_t \sim \Psi^X(\cdot|x_t,a^{\xi,X}_t)$ \Comment{perturbed observation}
                \end{algorithmic}
            \end{algorithm}
            \vspace{5pt}
                        
            \begin{figure}[!ht]
                \centering
                \includegraphics[scale=1.2]{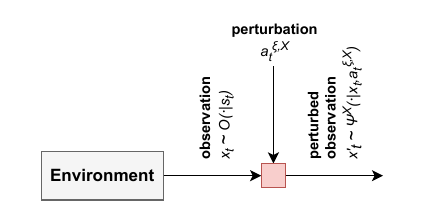}
                \caption{Integration of the observation perturbation function $\Psi^xi$ in the environment.}
                \label{fig:observation_perturbation}
            \end{figure}
            
            From a broader perspective, the adversary $\xi$ and the agent $\pi$ act simultaneously in a single environment $\Omega^{\pi,\xi}=(\Omega^\pi,\Omega^\xi)$ that combines the perspectives of both the adversary and the agent. Following this, the probability $P^{\Omega,\pi,\xi}(\widetilde{\tau}_t|s_t)$ of an adversary-augmented transition \mbox{$\widetilde{\tau}_t = (s_t,x_t,a_t,a^\xi_t,x'_t,a'_t,\widetilde{s}_t,\widetilde{x}_t,\widetilde{s}_{t+1},\widetilde{x}_{t+1},s_{t+1})$} given current state $s_t$ is given by:
            \begin{multline}
                P^{\Omega,\pi,\xi}(\widetilde{\tau}_t|s_t) =
                O(x_{t}|s_t)
                \xi(a^{\xi,X}_{t}|s_t,x_t)
                \Psi^X(x'_t|x_t,a^{\xi,X}_t)
                \pi(a_t|x'_t)
                \delta_{a_t}(a'_t)\\
                \delta_{s_t}(\widetilde{s}_t)
                O(\widetilde{x}_{t}|\widetilde{s}_t)
                T(\widetilde{s}_{t+1}|\widetilde{s}_t,a'_t)
                O(\widetilde{x}_{t+1}|s_{t+1})
                \delta_{\widetilde{s}_{t+1}}(s_{t+1})
                \label{eq:prob_obs}
            \end{multline}
            
            where $\delta_x$ stands for a Dirac distribution centered on $x$. In the case of observation perturbation, neither the states nor the action are perturbed. In Equation~\eqref{eq:prob_obs}, we therefore have $\delta_{a_t}(a'_t)$, $\delta_{s_t}(\widetilde{s}_t)$, and $\delta_{\widetilde{s}_{t+1}}(s_{t+1})$, which enforces the following identities for sampled components: $a'_t = \mathrm{Id}(a_t), \; \widetilde{s}_t = \mathrm{Id}(s_t), \; s_{t+1} = \mathrm{Id}(\widetilde{s}_{t+1})$, with $\mathrm{Id}(.)$ the identity function.

        \subsubsection{Alteration of the Environment Dynamics (Transition Function $T$)}
            
            \label{sec:altered_dynamics}

            The other type of component alteration is altering the transition function $T$ of the POMDP (i.e., the environment dynamics). The principle is to modify the effects of the actions of the protagonist in the environment. For example, this can include moving or modifying the behavior of some physical objects in the environment, like modifying the positions or speed of some vehicles in an autonomous driving simulator or modifying the way the protagonist's actions are affecting the environment (e.g. by amplifying or reversing actions). 
            
            This is done by emitting adversarial actions $A^\xi$, that are allowed by the environment $\Omega$ through a specific adversary function $\Psi^A$, $T_\Psi^S$, $T_\Psi$ or $T_\Psi^{S+}$, creating an altered transition function $T^\xi$ for the protagonist agent. In that setting, four types of adversaries can be considered:

             \paragraph{Transition Perturbation}\ \\
             
                In this setting, the process begins with the agent in a given state. The agent chooses an action, which is applied to the environment. This should lead to transition to a new state, according to the environment's transition function. However, this transition function is altered, effectively altering the environment dynamics, resulting in a different new subsequent state than if the transition had not been altered.
                
                For instance, in the context of an autonomous vehicle, the vehicle might decide to change lanes (action) based on the existing traffic setup (state). The application of this action should generally lead to transition to a specific next state, following the action chosen by the vehicle and the behavior of the other vehicles. But the behavior of surrounding vehicles is modified (altered transition), for instance modifying their speed. Consequently, the vehicle emerges in a new traffic configuration (next state) that is different from what would typically result from the chosen action if the behavior of the surrounding vehicles had not been modified.

                This process introduces variability into the environment dynamics by directly changing the transition function of the environment.
             
                More formally, the adversary $\xi$ acts to induce an altered next state $s_{t+1}$ by modifying the transition function itself, replacing it with the altered transition function $T^\Psi(s_{t+1}|(s_t,a_t),a^{\xi,T}_t)$ introduced in Section \ref{sec:perturbation}. In that case, $\xi$ can be regarded as an agent that acts by emitting adversarial actions $a^{\xi,T}_t \sim \xi(\cdot|s_t,x_t,a_t)$, given $\pi$ in a POMDP defined as  $\Omega^{\pi}=((S,A),A^{\xi,T},T^\pi,R_\xi,X,O)$. Here $T^\pi((s_{t+1},a_{t+1})|(s_t,a_t),a^{\xi,T}_t)$ is defined as the transition function of the environment, considering $\pi$, from the adversary's perspective. Consequently, the policy $\pi$ is incorporated into the environment dynamics as observed by the adversary. Sampling $(s_{t+1},a_{t+1}) \sim T^\pi(\cdot|(s_t,a_t),a^{\xi,T}_t)$ is performed as described in Algorithm \ref{algo:transition_adversary_pov}.

                \vspace{5pt}
                \begin{algorithm}[H]
                    \caption{Transition $T^\pi$, using the altered transition function $T_\Psi$, in the POMDP $\Omega^{\pi}$ of the adversary.}
                    \label{algo:transition_adversary_pov}
                    \begin{algorithmic}[1]
                        \State sample $s_{t+1} \sim T_\Psi(\cdot|s_t,a_t,a^{\xi,T}_t)$ \Comment{next state after altered transition}
                        \State sample $x_{t+1} \sim O(\cdot|s_{t+1})$                  \Comment{next observation}
                        \State sample $a_{t+1} \sim \pi(\cdot|x_{t+1})$                \Comment{next agent action}
                    \end{algorithmic}
                \end{algorithm}
                \vspace{5pt}

                Reversely, from the perspective of the protagonist agent $\pi$, it acts in an altered POMDP \mbox{$\Omega^{\xi}=(S,A,T^\xi,R,X,O)$}, where $s_{t+1} \sim T^\xi(\cdot|s_t,a_t)$ is performed as described in Algorithm \ref{algo:transition_agent_pov}.

                \vspace{5pt}
                \begin{algorithm}[H]
                    \caption{Transition $T^\xi$, using a altered transition function $T_\Psi$, in the POMDP $\Omega^{\xi}$ of the agent. Also illustrated in Figure \ref{fig:transition_perturbation}.}
                    \label{algo:transition_agent_pov}
                    \begin{algorithmic}[1]
                        \State sample $x_t \sim O(\cdot|s_t)$                               \Comment{observation}
                        \State sample $a^{\xi,T}_t \sim \xi
                        (\cdot|s_t,x_t,a_t)$                                                \Comment{adversary action T}
                        \State sample $s_{t+1} \sim T_\Psi(\cdot|s_t, a_t, a^{\xi,T}_t)$    \Comment{next state after altered transition}
                    \end{algorithmic}
                \end{algorithm}
                \vspace{5pt}
                
                \begin{figure}[!ht]
                    \centering
                    \includegraphics[scale=1.2]{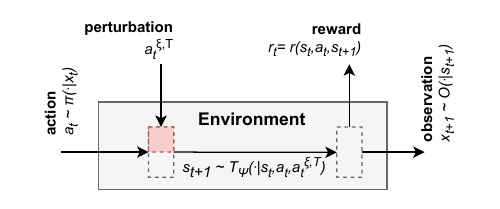}
                    \caption{Integration of the altered transition function $T_\Psi$ in the environment.}
                    \label{fig:transition_perturbation}
                \end{figure}

            \paragraph{Current State Perturbation}\ \\
            
                The process begins with the agent in a given state. The agent chooses an action to be applied within the environment. However, before this action is applied, the current state is perturbed. This perturbation modifies the initial state in which the chosen action is applied. The application of the action in this perturbed state leads to a transition to a new state according to the environment's transition function.
                
                For example, consider an autonomous vehicle deciding to change lanes (action) based on the prevailing traffic configuration (state). Before executing this maneuver, the traffic configuration is altered (perturbed state), such as by adjusting the positions of nearby vehicles. Consequently, when the vehicle performs its lane change, it does so in this adjusted traffic scenario, leading to a different traffic configuration (next state) than if the original state had not been modified.
                
                This process introduces variability into the environment dynamics without necessitating a direct modification of the transition function of the environment.
            
                More formally, the adversary $\xi$ acts to induce an perturbed next state $s_{t+1}$ by perturbing the state before the environment transition function, via the additional transition function $T_\Psi^{S}(\widetilde{s}_t|s_t,a^{\xi,S}_t)$ introduced in Section \ref{sec:perturbation}. In that case, $\xi$ can be regarded as an agent that acts by emitting adversarial actions $a^{\xi,S}_t \sim \xi(\cdot|s_t,x_t,a_t)$, given $\pi$ in a POMDP defined as  $\Omega^{\pi}=((S,A),A^{\xi,S},T^\pi,R_\xi,X,O)$. Here $T^\pi((s_{t+1},a_{t+1})|(s_t,a_t),a^{\xi,S}_t)$ is defined as the transition function of the environment, considering $\pi$, from the adversary's perspective. Consequently, the policy $\pi$ is incorporated into the environment dynamics as observed by the adversary. Sampling $(s_{t+1},a_{t+1}) \sim T^\pi(\cdot|(s_t,a_t),a^{\xi,S}_t)$ is performed as described in Algorithm \ref{algo:state_adversary_pov}.

                \vspace{5pt}
                \begin{algorithm}[H]
                    \caption{Transition $T^\pi$, using the current state perturbation function $T_\Psi^S$, in the POMDP $\Omega^{\pi}$ of the adversary.}
                    \label{algo:state_adversary_pov}
                    \begin{algorithmic}[1]
                        \State sample $\widetilde{s}_t \sim T_\Psi^S(\cdot|s_t,a^{\xi,S}_t)$ \Comment{perturbed state}
                        \State sample $s_{t+1} \sim T(\cdot|\widetilde{s}_t, a_t)$           \Comment{next state after transition}
                        \State sample $x_{t+1} \sim O(\cdot|s_{t+1})$                        \Comment{next observation}
                        \State sample $a_{t+1} \sim \pi(\cdot|x_{t+1})$                      \Comment{next agent action}
                    \end{algorithmic}
                \end{algorithm}
                \vspace{5pt}

                Reversely, from the perspective of the protagonist agent $\pi$, it acts in an altered POMDP \mbox{$\Omega^{\xi}=(S,A,T^\xi,R,X,O)$}, where $s_{t+1} \sim T^\xi(\cdot|s_t,a_t)$ is performed as described in Algorithm \ref{algo:state_agent_pov}.

                \vspace{5pt}
                \begin{algorithm}[H]
                    \caption{Transition $T^\xi$, using the current state perturbation function $T_\Psi^S$, in the POMDP $\Omega^{\xi}$ of the agent. Integration illustrated in Figure \ref{fig:state_perturbation}.}
                    \label{algo:state_agent_pov}
                    \begin{algorithmic}[1]
                        \State sample $x_t \sim O(\cdot|s_t)$                                                       \Comment{observation}
                        \State sample $a^{\xi,S}_t \sim \xi
                        (\cdot|s_t,x_t,a_t)$                                                                        \Comment{adversary action S}
                        \State sample $\widetilde{s}_t \sim T_\Psi^{S}(\cdot|s_t,a^{\xi,S}_t)$                      \Comment{perturbed state}
                        \State sample $s_{t+1} \sim T(\cdot|\widetilde{s}_t, a_t)$                                  \Comment{final next state after transition}
                    \end{algorithmic}
                \end{algorithm}
                \vspace{5pt}
                
                \begin{figure}[!ht]
                    \centering
                    \includegraphics[scale=1.2]{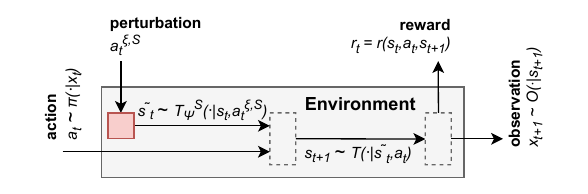}
                    \caption{Integration of the current state perturbation function $T_\Psi^{S}$ in the environment.}
                    \label{fig:state_perturbation}
                \end{figure}

            \paragraph{Next State Perturbation}\ \\
            
                The process begins with the agent in a given state. The agent chooses an action that is then applied in the environment. This leads to transition to a new subsequent state, according to the environment's transition function. However, before the agent can choose its next action, this new state is perturbed.
                
                For instance, in the case of an autonomous vehicle, the vehicle might choose to change lanes (action) based on the current traffic configuration (state). After the action is executed, the vehicle finds itself in a new traffic configuration (next state). Before choosing the next action, this new state is perturbed, for example, by modifying the positions of surrounding vehicles. This means the vehicle now faces a modified traffic configuration (perturbed next state) from which it must decide its next move.
                
                This process introduces variability into the environment dynamics without necessitating a direct modification of the transition function of the environment.
                
                The key difference between perturbing the current state and perturbing the next state lies in the agent's awareness of the situation. In \textit{current state perturbation}, the agent lacks actual knowledge of its precise state when choosing the action because this state is modified just before the action is applied. However, in the \textit{next state perturbation}, the agent has full awareness of its current state when choosing the action.
                
                More formally, the adversary $\xi$ acts to produce an modified next state $s_{t+1}$ by perturbing the next state after the transition function via the posterior transition function $T_\Psi^{S+}(\widetilde{s}_{t+1}|s_{t+1},a^{\xi,S+}_t)$ introduced in Section \ref{sec:perturbation}.
                
                In that case, $\xi$ can be regarded as an agent that acts by emitting adversarial actions $a^{\xi,S+}_t \sim \xi(\cdot|s_t,x_t)$, given $\pi$ in a POMDP defined as  $\Omega^{\pi} = (S,A^{\xi,S+},T^\pi,R_\xi,X,O)$. Here $T^\pi(s_{t+1}|s_t,a^{\xi,S}_t)$ is defined as the transition function of the environment, considering $\pi$, from the adversary's perspective. Consequently, the policy $\pi$ is incorporated into the environment dynamics as observed by the adversary. Sampling $s_{t+1} \sim T^\pi(\cdot|(s_t,a_t),a^{\xi,S}_t)$ is performed as described in Algorithm \ref{algo:next_state_adversary_pov}.

                \vspace{5pt}
                \begin{algorithm}[H]
                    \caption{Transition $T^\pi$, using the next state perturbation function $T_\Psi^{S+}$, in the POMDP $\Omega^{\pi}$ of the adversary.}
                    \label{algo:next_state_adversary_pov}
                    \begin{algorithmic}[1]
                        \State sample $\widetilde{s}_t \sim T_\Psi^{S+}(\cdot|s_t,a^{\xi,S+}_t)$ \Comment{perturbed state}
                        \State sample $\widetilde{x}_t \sim O(\cdot|\widetilde{s}_t)$            \Comment{observation}
                        \State sample $a_t \sim \pi(\cdot|x_t)$                                  \Comment{agent action}
                        \State sample $s_{t+1} \sim T(\cdot|\widetilde{s}_t, a_t)$               \Comment{next state after transition}
                    \end{algorithmic}
                \end{algorithm}
                \vspace{5pt}

                Reversely, from the perspective of the protagonist agent $\pi$, it acts in an altered POMDP \mbox{$\Omega^{\xi}=(S,A,T^\xi,R,X,O)$}, where $s_{t+1} \sim T^\xi(\cdot|s_t,a_t)$ is performed as described in Algorithm \ref{algo:next_state_agent_pov}.

                \vspace{5pt}
                \begin{algorithm}[H]
                    \caption{Transition $T^\xi$, using the next state perturbation function $T_\Psi^{S+}$, in the POMDP $\Omega^{\xi}$ of the agent. Integration illustrated in Figure \ref{fig:next_state_perturbation}.}
                    \label{algo:next_state_agent_pov}
                    \begin{algorithmic}[1]
                        \State sample $x_t \sim O(\cdot|s_t)$                                                       \Comment{observation}
                        \State sample $\widetilde{s}_{t+1} \sim T(\cdot|s_t, a_t)$                                  \Comment{next state after transition}
                        \State sample $\widetilde{x}_{t+1} \sim O(\cdot|\widetilde{s}_{t+1})$                       \Comment{observation of the next state}
                        \State sample $a^{\xi,S+}_t \sim \xi
                        (\cdot|\widetilde{s}_{t+1},\widetilde{x}_{t+1})$   \Comment{adversary action S+}
                        \State sample $s_{t+1} \sim T_\Psi^{S+}(\cdot|\widetilde{s}_{t+1}, a^{\xi,S+}_t)$           \Comment{final next state after perturbation}
                    \end{algorithmic}
                \end{algorithm}
                \vspace{5pt}
                
                \begin{figure}[!ht]
                    \centering
                    \includegraphics[scale=1.2]{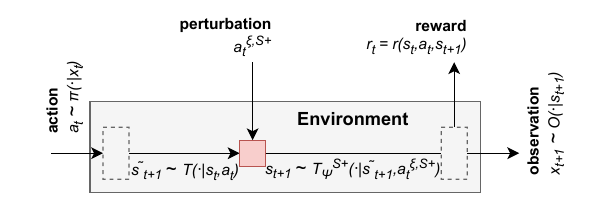}
                    \caption{Integration of the next state perturbation function $T_\Psi^{S+}$ in the environment.}
                    \label{fig:next_state_perturbation}
                \end{figure}

            \paragraph{Action Perturbation}\ \\
            
                The process starts with the agent in a given state. The agent chooses an action intended to be applied in the environment. However, before this action can be applied, it undergoes a perturbation, resulting in a perturbed action. This perturbed action is then applied, leading to transition to a new state according to the environment's transition function.
                
                For instance, consider an autonomous vehicle that decides to steer at an angle of $\alpha$ (action) based on the current traffic configuration (state). Before the steering action is executed, it is perturbed, so the actual steering angle applied to the vehicle becomes $\alpha + \epsilon$ (perturbed action). As a result, the vehicle transitions to a new traffic configuration (next state) reflecting the outcome of the perturbed steering action.
                                
                This process introduces variability into the environment dynamics without necessitating a direct modification of the environment's transition function or modification of the state of the environment. However, this approach to modifying dynamics, while introducing variability, is confined to the scope of action perturbation, limiting the diversity of potential dynamics alterations.
                
                More formally, the adversary $\xi$ acts to induce an modified next state $s_{t+1}$ by perturbing the action decided by the agent $a_t \sim \pi(\cdot|x_t)$ via the specific perturbation function $\Psi^A(a'_t|a_t,a^{\xi,A}_t)$ introduced in Section \ref{sec:perturbation}.
                
                In that case, $\xi$ can be regarded as an agent that acts by emitting adversarial actions $a^{\xi,A}_t \sim \xi(\cdot|s_t,x_t,a_t)$, given $\pi$ in a POMDP defined as  $\Omega^{\pi} = \big((S,A^\Omega),A^{\xi,A},T^\pi,R_\xi,X,O\big)$. Here $T^\pi((s_{t+1},a_{t+1})|s_t,a^{\xi,S}_t)$ is defined as the transition function of the environment, considering $\pi$, from the adversary's perspective. Consequently, the policy $\pi$ is incorporated into the environment dynamics as observed by the adversary. Sampling $(s_{t+1},a_{t+1}) \sim T^\pi(\cdot|(s_t,a_t),a^{\xi,S}_t)$ is performed as described in Algorithm \ref{algo:action_adversary_pov}.

                \vspace{5pt}
                \begin{algorithm}[H]
                    \caption{Transition $T^\pi$, using the action perturbation function $\Psi^A$,  in the POMDP $\Omega^{\pi}$ of the adversary.}
                    \label{algo:action_adversary_pov}
                    \begin{algorithmic}[1]
                        \State sample $a'_t \sim \Psi^A(\cdot|a_t,a^{\xi,A}_t)$  \Comment{perturbed agent action}
                        \State sample $s_{t+1} \sim T(\cdot|s_t, a'_t)$          \Comment{next state after transition}
                        \State sample $x_{t+1} \sim O(\cdot|s_{t+1})$            \Comment{next observation}
                        \State sample $a_{t+1} \sim \pi(\cdot|x_{t+1})$          \Comment{next agent action}
                    \end{algorithmic}
                \end{algorithm}
                \vspace{5pt}
                
                Reversely, from the perspective of the protagonist agent $\pi$, it acts in an altered POMDP \mbox{$\Omega^{\xi}=(S,A,T^\xi,R,X,O)$}, where $s_{t+1} \sim T^\xi(\cdot|s_t,a_t)$ is performed as described in Algorithm \ref{algo:action_agent_pov}.

                \vspace{5pt}
                \begin{algorithm}[H]
                    \caption{Transition $T^\xi$, using the action perturbation function $\Psi^A$, in the POMDP $\Omega^{\xi}$ of the agent. Integration illustrated in Figure \ref{fig:action_perturbation}.}
                    \label{algo:action_agent_pov}
                    \begin{algorithmic}[1]
                        \State sample $x_t \sim O(\cdot|s_t)$                                                       \Comment{observation}
                        \State sample $a^{\xi,A}_t \sim \xi
                        (\cdot|s_t,x_t,a_t)$                                   \Comment{adversary action A}
                        \State sample $a'_t \sim \Psi^A(\cdot|a_t,a^{\xi,A}_t)$                                     \Comment{perturbed action}
                        \State sample $s_{t+1} \sim T(\cdot|s_t, a'_t)$                                             \Comment{next state after transition}
                    \end{algorithmic}
                \end{algorithm}
                \vspace{5pt}
                
                \begin{figure}[!ht]
                    \centering
                    \includegraphics[scale=1.2]{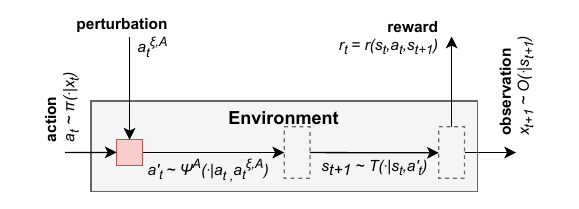}
                    \caption{Integration of the action perturbation function $\Psi^A$ in the environment.}
                    \label{fig:action_perturbation}
                \end{figure}

            From the perspective of the protagonist agent $\pi$, we can also gather all four possible perturbations of the environment dynamics just described above in a single transition, by denoting the adversaries $\xi_A$, $\xi_S$, $\xi_{T}$ and $\xi_{S+}$ respectively. The agent $\pi$ acts on an altered POMDP \mbox{$\Omega^{\xi}=(S,A,T^\xi,R,X,O)$}, where $s_{t+1} \sim T^\xi(\cdot|s_t,a_t)$ is performed as described in Algorithm \ref{algo:dynamics_agent_pov}.

            \vspace{5pt}
            \begin{algorithm}[H]
                \caption{Transition $T^\xi$ with all environment dynamics perturbations, induced by adversaries $\xi_A$, $\xi_S$, $\xi_{T}$ and $\xi_{S+}$, in the POMDP $\Omega^{\xi}$ of the agent.}
                \label{algo:dynamics_agent_pov}
                \begin{algorithmic}[1]
                    \State sample $x_t \sim O(\cdot|s_t)$                                                       \Comment{observation}
                    \State sample $a^{\xi,A}_t \sim \xi_A(\cdot|s_t,x_t,a_t)$                                   \Comment{adversary action A}
                    \State sample $a^{\xi,S}_t \sim \xi_S(\cdot|s_t,x_t,a_t)$                                   \Comment{adversary action S}
                    \State sample $a'_t \sim \Psi^A(\cdot|a_t,a^{\xi,A}_t)$                                     \Comment{perturbed action}
                    \State sample $\widetilde{s}_t \sim T_\Psi^{S}(\cdot|s_t,a^{\xi,S}_t)$                      \Comment{perturbed state}
                    \State sample $\widetilde{x}_t \sim O(\cdot|\widetilde{s}_t)$                               \Comment{observation of the perturbed state}
                    \State sample $a^{\xi,T}_t \sim \xi_T(\cdot|\widetilde{s}_t,\widetilde{x}_t,a'_t)$          \Comment{adversary action T}
                    \State sample $\widetilde{s}_{t+1} \sim T_\Psi(\cdot|\widetilde{s}_t, a'_t, a^{\xi,T}_t)$   \Comment{next state after perturbed transition}
                    \State sample $\widetilde{x}_{t+1} \sim O(\cdot|\widetilde{s}_{t+1})$                       \Comment{observation of the next state}
                    \State sample $a^{\xi,S+}_t \sim \xi_{S+}(\cdot|\widetilde{s}_{t+1},\widetilde{x}_{t+1})$   \Comment{adversary action S+}
                    \State sample $s_{t+1} \sim T_\Psi^{S+}(\cdot|\widetilde{s}_{t+1}, a^{\xi,S+}_t)$           \Comment{final next state}
                \end{algorithmic}
            \end{algorithm}
            \vspace{5pt}

            From a broader perspective, the adversary $\xi$ and the agent $\pi$ act simultaneously in a single environment $\Omega^{\pi,\xi}=(\Omega^\pi,\Omega^\xi)$ that combines the perspectives of both the adversary and the agent. Following this, the probability $P^{\Omega,\pi,\xi}(\widetilde{\tau}_t|s_t)$ of an adversary-augmented transition \mbox{$\widetilde{\tau}_t=(s_t,x_t,a_t,a^\xi_t,x'_t,a'_t,\widetilde{s}_t,\widetilde{x}_t,\widetilde{s}_{t+1},\widetilde{x}_{t+1},s_{t+1})$} given current state $s_t$, is given by:

            \begin{multline}
                P^{\Omega,\pi,\xi}(\widetilde{\tau}_t|s_t) =
                O(x_t|s_t)
                \delta_{x_t}(x'_t)
                \pi(a_t|x'_t)
                \xi_A(a^{\xi,A}_t|s_t,x_t,a_t)
                \Psi^A(a'_t|a_t,a^{\xi,A}_t)\\
                \xi_S(a^{\xi,S}_t|s_t,x_t,a_t)
                T_\Psi^S(\widetilde{s}_t|s_t,a^{\xi,S}_t)
                O(\widetilde{x}_t|\widetilde{s}_t)
                \xi_{T}(a^{\xi,T}_t|\widetilde{s}_t,\widetilde{x}_t,a'_t)
                T_\Psi(\widetilde{s}_{t+1}|\widetilde{s}_t,a'_t,a^{\xi,T}_t)\\
                O(\widetilde{x}_{t+1}|\widetilde{s}_{t+1})
                \xi_{S+}(a^{\xi,S+}_t|\widetilde{s}_{t+1},\widetilde{x}_{t+1})
                T_\Psi^{S+}(s_{t+1}|\widetilde{s}_{t+1},a^{\xi,S+}_t)
                \label{eq:prob_env}
            \end{multline}

            where $\delta_x$ stands for a Dirac distribution centered on $x$. In the case of environment dynamics alteration, the observation is not perturbed. In Equation~\eqref{eq:prob_env}, we therefore have $\delta_{x_t}(x'_t)$, which enforces the identity $x'_t = \mathrm{Id}(x_t)$.

    \subsection{Adversarial Objective}
    
        \label{sec:objective}

        Adversarial attacks in RL are strategically designed to compromise specific aspects of agent behavior or environment dynamics. In general, they aim to prevent the agent from acting optimally, but the attacks vary in their objectives and methodologies. Even if the general goal of any adversarial attack is to reduce the performance of the agent, methods to achieve this can primarily have different objective functions for specific performance reductions.

        We discuss here the optimization objectives $\Delta^{\pi,\Omega}(\xi)$ of the adversary agents $\xi$, as previously introduced in Section \ref{sec:formalization_dro_adv_training}.

        \subsubsection{Short-Term Divergence Metric}

            \label{sec:short_divergence}
            Following the literature on adversarial attacks in the supervised setting, the goal of an attack can be to induce decision divergence. Applied to RL, the primary goal is to deviate the agent from its initial, typically optimal, policy. We can deviate the policy to make it diverge from the original trajectory: in that case, the adversary $\xi$ is designed to maximize the divergence over pairs of action distributions, the first being the action of the policy given the original conditions and the second being the action of the policy given perturbed conditions. The divergence metrics $D$ are most commonly losses over the policies of the agent (e.g. Cross Entropy or MSE) but can also be any divergence metric (e.g. Kullback-Leibler or Wasserstein divergences). More generally, given a transition tuple $\tau_t = (s_t, x_t, a_t, s_{t+1})$, the divergence can be computed over any pair of elements obtained from the perturbed transition tuple $\widetilde{\tau}_t = (s_t, x_t, a_t, a_t^\xi, x'_t, a'_t, \widetilde{s}_t, \widetilde{s}_{t+1}, s_{t+1})$.
                         
            \paragraph*{Untargeted attacks}\ \\
                Using the divergence $D$ between the original and the perturbed condition as a metric to be maximized, the optimal adversary $\xi^*$ becomes:
                \begin{equation}
                    \xi^* = \arg\max_\xi \; \mathbb{E}_{s_t \sim d^{\pi,\Omega}} \; \Big[D\Big(p^{\Omega,\pi}(\cdot|s_t),p^{\Omega,\pi,\xi}(\cdot|s_t)\Big)\Big]
                \end{equation}
                with $d^{\pi,\Omega}$ being the stationary distribution on the transition tuples when following the policy $\pi$ in the environment $\Omega$.
            
            \paragraph*{Targeted attacks}\ \\
                The divergence can be formulated as $D(T,\cdot)$, where T represents a target condition. In this case, the target replaces the original condition in the divergence formulation.
                Using the divergence as a metric to be minimized, the optimal adversary $\xi^*$ becomes:
                \begin{equation}
                    \xi^* = \arg\min_\xi \; \mathbb{E}_{s_t \sim d^{\pi,\Omega}}  \; \Big[D\Big(T,p^{\Omega,\pi,\xi}(\cdot|s_t)\Big)\Big]
                \end{equation}

            In both settings, such attacker optimization can be performed on-policy, rather than off-policy by acting iteratively from a current distribution $d^{\pi,\hat{\xi},\Omega}$ (in place of $d^{\pi,\Omega}$) stating as the transition distribution on $\Omega$ following  $\pi$ given perturbations from the current attacker $\hat{\xi}$ at the considered optimization step.

        \subsubsection{Long-Term Adversarial Reward}

            \label{sec:long_reward}
            In contrast, some adversarial attacks focus on leading the agent to less favorable states or decisions, thereby minimizing the total expected reward the agent accrues, or maximizing specifically designed rewards corresponding to specify malicious goals.

            \paragraph*{Untargeted attacks}\ \\
                Adversarial Attacks can seek for reduction of the efficacy of the agent’s behavior by inducing minimization of its reward.
                In this case the goal is to reduce the rewards obtained by the agent, the optimization becomes:
                \begin{equation}
                    \xi^* \; = \; \arg\min_\xi \; \mathbb{E}_{\tilde{\tau} \sim \pi^{\xi,\Omega}} \; \Big[R(\widetilde{\tau})\Big]
                \end{equation}
                This formulation seek for the optimal adversarial strategy $\xi^*$ minimizing reward of the protagonist agent.
                
            \paragraph*{Targeted attacks}\ \\            
                Adversarial Attacks can also seek a specific agent’s behavior or target state, by designing a specific reward $R_\xi$ for the adversary to maximize (e.g. the crash of the controller vehicle). The optimization becomes:
                \begin{equation}
                    \xi^* = \arg\max_\xi \; \mathbb{E}_{\tilde{\tau} \sim \pi^{\xi,\Omega}} \; \Big[R_\xi(\widetilde{\tau})\Big]
                \end{equation}
                This formulation seeks for the optimal adversarial strategy $\xi^*$ maximizing adversarial reward designed for a specific malicious objective.

    \subsection{Knowledge Requirement}
    
        In the realm of adversarial attacks against RL agents, the extent and nature of the attacker's knowledge about the agent significantly influence the strategy and effectiveness of the attack. Broadly, these can be categorized into white-box and black-box approaches, each with its own set of strategies, challenges, and considerations.
        
        \subsubsection{White-Box}
        
            In this scenario, the adversary has complete knowledge of the agent's architecture, parameters, and training data. This scenario represents the most informed type of attack, where the adversary has access to all the inner workings of the agent, including its policy, value function, and possibly even the environment model.
            \begin{itemize}
                \item Policy and Model Access: The adversary knows the exact policy and decision-making process of the agent. This includes access to the policy's parameters, algorithm type, and architecture. In model-based RL, the attacker might also know the transition and reward functions.
                \item Optimization and Perturbation: With complete knowledge, the attacker can craft precise and potent perturbations to the agent's inputs or environment to maximize the deviation from desired behaviors or minimize rewards. They can calculate the exact gradients or other relevant information needed to optimize their attack strategy.
                \item Challenges and Implications: While white-box attacks represent an idealized scenario with maximal knowledge, they provide a comprehensive framework for testing the agent's robustness. By simulating the most extreme conditions an agent could face, developers can identify and reinforce potential vulnerabilities, leading to policies that are not only effective but also robust to a wide range of scenarios, including unexpected environmental changes. This approach is particularly valuable in safety-critical applications where ensuring reliability against all possible perturbations is crucial.
            \end{itemize}
            
        \subsubsection{Black-Box}
        
            In this scenario, the adversary has limited or no knowledge of the internal workings of the agent. They may not know the specific policy, parameters, or architecture of the RL agent. Instead, they must rely on observable behaviors or outputs to infer information and craft their attacks.
            \begin{itemize}
                \item Observational Inference: The attacker observes the agent's actions and possibly some aspects of the state transitions to infer patterns and weaknesses, or predict future actions. This process often involves probing the agent with different inputs and analyzing the outputs.
                \item Surrogate Models and Transferability: Attackers might train a surrogate model to approximate the agent's behavior or policy. If an attack is successful on the surrogate, it might also be effective on the target agent due to transferability, especially if both are trained in similar environments or tasks.
                \item Challenges and Implications: The use of black-box methods in enhancing robustness is not directly about the realism of adversarial intent but rather about preparing for a variety of uncertain conditions and environmental changes. These methods encourage the development of general defense mechanisms that improve the agent's adaptability and resilience. While the adversarial mindset might not reflect typical operational challenges, the diversity and unpredictability of black-box approaches help ensure that RL systems are robust not only against potential adversaries but also against a wide range of non-adversarial issues that can arise in dynamic and uncertain environments.
            \end{itemize}

    \subsection{Category of Approach}
        
        This section outlines the two main methodologies used to craft adversarial attacks. It is divided into direct optimization and adversarial policy learning approaches.
        
        \subsubsection{Direct Optimization}

            \label{sec:direct_optim}
            
            These approaches compute online for each sample the optimization algorithm to determine the perturbation to produce, usually to optimize a divergence metric. Many methods are derivative-based methods, called \textbf{gradient attacks}, which utilize the gradient information of the model for optimizing the adversarial objective to craft adversarial examples, efficiently targeting the model's weaknesses but requiring white-box scenarios. Other methods are \textbf{derivative-free} methods, which optimize the adversarial objective without requiring gradient information, making them suitable for black-box scenarios. Techniques include simulated annealing, genetic algorithms, random search, etc. All the methods in this category have in common that they need to perform an optimization for each sample to perturb.
                
        \subsubsection{Adversarial Policy}
            
            \label{sec:adv_policies}

            These approaches involve training an \textbf{adversarial policy} (AP) which learns an optimal attack strategy through interaction with the target system and an adversarial reward. The training of these adversarial policies is often done using RL, thus only requiring black-box access to the model of the agent. There are also \textbf{augmented adversarial policies} trained via RL augmented with white-box access to the agent's model during the training or inference phase. All the methods in this category have in common that they first do an optimization for training the adversarial policy, then this policy can be used in inference to perturb each sample.\\

    In the following sections, we will use this taxonomy as a framework to examine recent research on adversarial examples for RL. Section \ref{sec:observation_alteration} focuses on input-space perturbations, and Section \ref{sec:dynamics_alteration} on environment-space perturbations.


\section{Adversarial Attacks}
    
    \label{sec:adv_attacks}

    In this section, we conduct a comprehensive review of contemporary adversarial attacks as documented in current literature, presented in a hierarchical, tree-like structure (refer to Figure \ref{fig:categorization}). The review categorizes these attacks first based on the type of alteration induced in the POMDP: either Observation Alteration or Dynamic Alteration. Next, the categorization considers the underlying objective driving these attacks, which could be either with short-term divergence metrics or long-term adversarial reward. Lastly, the classification focuses on the computational approach employed to generate perturbation: Direct Optimization or Adversarial Policy. For each method in this classification, we will provide a detailed description, ensuring to consistently include the following critical information: the nature of the perturbation support (whether it's an observation, state, action, or transition function), the level of knowledge about the model required to execute the attack (white-box or black-box), and any specific constraints or potential limitations associated with the method.

    \subsection{Observation Attacks}
    
        \label{sec:observation_alteration}
        This section delves into the analysis of Observation Alteration Attacks targeting RL agents. These attacks specifically modify the observation function in the POMDP framework. Such methods are instrumental in simulating sensor errors in an agent, creating discrepancies between the agent's perceived observations and the actual underlying state. These techniques can be particularly beneficial during an agent's training phase, enhancing its resilience to potential observation discrepancies that might be encountered in real-world deployment scenarios. As described in Section \ref{sec:altered_observation} Observation Alteration Attacks generate a perturbation $a^{\xi,X}_\epsilon$ for a given observation $x$, resulting in a perturbed observation $x'_t \sim \Psi^X(\cdot| x_t,a^{\xi,X}_t)$. The perturbation $\|x'_t-x_t\|_p$ is usually constrained within an $\epsilon$-ball of a specified $l_p$ norm.
        \subsubsection{Attacks Driven by Short-Term Divergence Metrics}
            \label{sec:obs_short_divergence_attack}
            
            Optimizing a divergence metric (\ref{sec:short_divergence}) is a much-used way to generate perturbations on the observations. Most of the methods that apply these principles are Direct Optimization Attacks (\ref{sec:direct_optim}). They compute perturbations given any sample to craft an adversarial observation that is aimed to optimize a certain immediate divergence or loss.

            \paragraph{Gradient Attacks (White-Box)}\ \\
                \label{sec:obs_gradient_attack}
                
\begin{figure}[!ht]
                    \centering
                    \includegraphics[scale=1]{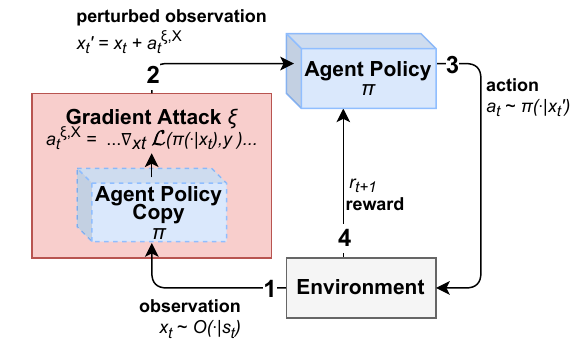}
                    \caption{Gradient Observation Attacks: The adversarial attack intercepts the observation $x_t$, computes a perturbation $a^{\xi,X}_t$ by back-propagating the gradient of a loss in the neural network of a copy of the agent $\pi$, this perturbation is used to craft a perturbed observation $x'_t$, which is sent to the agent}
                    \label{fig:gradient_observation_attack}
                \end{figure}

                Initially introduced in the context of supervised classification, gradient attacks (see section \ref{sec:dnn_attacks}) can be applied directly on RL when the policy model is available. When adapted to RL, the basic gradient-based perturbation formula \eqref{eq:supervised_grad}  essentially remains unchanged, except $f(x)$ is replaced by policy $\pi(.|x)$ given the context, and $\mathcal{L}$ is a pre-specified divergence metric for the corresponding distribution: 
                \begin{equation}
                    x' = x + a^{\xi,X} \quad \text{with} \quad  a^{\xi,X} \propto \nabla_x \mathcal{L}\big(\pi(.|x)\big)
                \end{equation}
                with $\varepsilon=\|a^{\xi,X}\|$ controlling the magnitude of the perturbation. A representation of the integration and application of Gradient Attacks in crafting observation perturbations within an RL framework is shown in Figure \ref{fig:gradient_observation_attack}.
                    
                Below, we summarize commonly used gradient methods, grouped by their objective: attacks that operate under a perturbation budget (bounded-norm attacks) and attacks that search for a minimal perturbation (minimum-norm attacks). For each method, we report the usual operational behaviour and a representative, per-example order-of-magnitude of forward and backward pass cost (for gradient evaluations and queries). All computational costs are approximate and depend strongly on the chosen iteration budget, model architecture, and implementation.
            
                \paragraph*{Attacks with a perturbation budget (bounded-norm)}\ \\
                    These methods seek an adversarial example $x'$ that changes the most the output of the model $\pi_{\theta}$ according to a metric $\mathcal{L}$ (e.g., Kullback-Leibler divergence), while satisfying the perturbation budget $\|x-x'\|_p \le \epsilon$.
                    \begin{equation}
                        \max_{x'} \mathcal{L}\big(\pi_{\theta}(.|x),\pi_{\theta}(.|x')\big)  \quad s.t. \quad \|x-x'\|_p \le \epsilon
                        \label{eq:adv_ex_bounded_norm_2}
                    \end{equation}

                    \begin{itemize}
                      \item \textbf{Fast Gradient (Sign) Method} (FGSM or FGM) \mycite{goodfellow2014explaining}.  The original formulation is a single-step attack using the sign of the input gradient: $x' = x + \varepsilon\cdot\mathrm{sign}(\nabla_x\mathcal L)$, but it is common to use the normalized natural gradient instead $x' = x + \varepsilon\cdot\frac{\nabla_x\mathcal L}{\|\nabla_x\mathcal L\|}$ depending on the needs. \textbf{Cost: 1 forward and backward pass}.
                      \item \textbf{Basic Iterative Method} (BIM) \mycite{kurakin2018adversarial}, \textbf{Projected Gradient Descent} (PGD) \mycite{madry2017towards}. Iterative extensions of FGSM; each iteration takes a gradient step and (optionally) projects back into the $\epsilon$-ball. \textbf{Cost: $\sim$10s–100s of forward and backward passes}.
                      \item \textbf{Momentum Iterative FGSM} (MI-FGSM) \mycite{dong2018boosting}. FGSM variant that accumulates a momentum term over iterations to stabilize updates and improve transferability. \textbf{Cost: $\sim$10s of forward and backward passes}.
                      \item \textbf{Frank-Wolfe} attack \mycite{chen2020frank}. A projection-free iterative method that solves the bounded-norm maximization via a linear minimization oracle at each step, avoiding the projection used in PGD while improving query/iteration efficiency. \textbf{Cost: $\sim$10s–100s of forward and backward passes}.
                      \item \textbf{AutoAttack} and \textbf{Auto-PGD} (APGD) \mycite{croce2020reliable}. An ensemble of parameter-free variants of PGD/APGD used for robust evaluation. \textbf{Cost: $\sim$1000s–10$^4$s of forward and backward passes} (APGD is a common bottleneck).
                      
                      \item \textbf{Jacobian Saliency Map Attack} (JSMA) \mycite{papernot2016limitations} iteratively perturbs features by inspecting the input–output Jacobian; suited for $\ell_0$ style attacks but computationally expensive due to the iterative Jacobian computations, which require one backward pass for each one of the $N$ output features. \textbf{Cost: $C=\sim$100s–1000s of forward passes and $C\times N$ of backward passes}.
                      \item \textbf{Voting Folded Gaussian Attack} (VFGA) \mycite{cesaire2021stochastic,hajri2022neural}, inspired by JSMA, iteratively computes the partial derivatives $\partial F_c/\partial x_i$ to identify the most promising features to perturb, and samples multiple folded-Gaussian perturbations and evaluates them with forward passes. \textbf{Cost: $\sim$10s-100s of backward and $\sim$100s-1000s of forward passes}.
                    \end{itemize}
                    
                    Several of the attacks given in the previous paragraph have been applied as is to RL in an untargeted way \mycite{behzadan2017vulnerability,huang2017adversarial} in various types of environments, and in a targeted way \mycite{pattanaik2017robust}.
                    Other gradient attacks have been developed specifically for RL:\\
                    \begin{itemize}
                        \item \textbf{Enchanting Attack} (EA) \mycite{lin2017tactics} is a method that uses a model to predict the future state, then selects the adversarial direction that leads to the worst state. And finally, apply a gradient attack on the observation targeting that adversarial direction. \textbf{Cost: Train a prediction model, then 2 forward and 1 backward passes}
                        \item \textbf{Maximal Action Difference Attack} (MAD) \mycite{zhang2020robust} is an adversarial attack that computes the gradient of the Kullback-Leibler Divergence of the policy to generate perturbations. \textbf{Cost: 1 forward and backward pass}.
                        \item \textbf{Distribution-Aware Projected Gradient Descent} (DAPGD) \mycite{duan2025rethinking} is a recent gradient attack that, rather than targeting a single action, perturbs the observation by descending the gradient of a distribution-similarity metric (Bhattacharyya distance) between the clean and perturbed policy distributions, yielding more consistent perturbations in continuous action spaces. \textbf{Cost: $\sim$10s--100s of forward and backward passes}.
                        \item \textbf{Nesterov Momentum FGM} (NM-FGM) \mycite{korkmaz2020nesterov}, inspired by MI-FGSM, is a gradient attack adapted to the sequential nature of RL, keeping momentum in the perturbations over time steps. \textbf{Cost: 1 forward and backward pass}.
                        \item \textbf{Observation Attack on the Actor Network} (OAAN) \mycite{schott2022improving} is the application of a variant of the FGM attack applied to Actor Networks of a PPO Agent. \textbf{Cost: 1 backward pass}.
                    \end{itemize}

                \paragraph*{Attacks that search for a minimal perturbation (minimum-norm)}\ \\
                    These methods aim to find the smallest $\ell_p$ perturbation that changes the policy distribution beyond a given threshold $\epsilon$,  w.r.t. a divergence metric ${\cal L}$.
                    \begin{equation}
                        \min_{x'} \|x-x'\|_p \quad s.t. \quad {\cal L}(\pi_\theta(.|x), \pi_\theta(.|x')) \geq \epsilon
                        \label{eq:adv_ex_min_norm2}
                    \end{equation}
                    
                    \begin{itemize}
                      \item \textbf{Fast geometric / linearization methods}:
                        \textbf{DeepFool} \mycite{moosavi2016deepfool}: iteratively linearizes the classifier and steps to the boundary; fast and parameter-free but suboptimal in norm. \textbf{SuperDeepFool} (SDF) \mycite{abroshan2024superdeepfool}: DeepFool family with projection/alignment steps; retains low cost while producing much smaller norms. \textbf{Cost: $\sim$10s–100s of backward passes}.
                    
                      \item \textbf{Efficient projected / iterative minimum-norm methods}:
                          \textbf{Decoupled Direction and Norm} (DDN) \mycite{rony2019decoupling}: decouples direction and norm updates with $\ell_2$ projections; designed for efficiency and used in adversarial training. Typical cost: medium (tens–hundreds of gradient passes).
                          \textbf{Fast Minimum-Norm} (FMN) \mycite{pintor2021fast}: fast minimum-norm method with adaptive constraints, improving efficiency over some iterative optimizers. \textbf{Cost: $\sim$10s–100s of backward passes}.
                    
                      \item \textbf{Optimization-based / high-accuracy minimum-norm attacks}:
                          Carlini\&Wagner (\textbf{C\&W}) \mycite{carlini2017towards}: optimization formulation that minimizes perturbation norm subject to misclassification; very accurate but computationally expensive. \textbf{Cost: $\sim$1000s–10$^4$s of backward passes}.
                          \textbf{Fast Adaptive Boundary Attack} (FAB) and \textbf{Augmented Lagrangian Adversarial Attacks} (ALMA) \mycite{croce2020minimally, rony2021augmented} approaches achieve strong minimality at the expense of high iteration counts. \textbf{Cost: $\sim$100s–1000s of backward passes}.
                    \end{itemize}

            \paragraph{Derivative-Free Attack (Black-Box)}\ \\
                \label{sec:obs_derivative-free_attack}
                
                Besides gradient-based methods, a distinct family of attacks assumes only black-box access to the model’s predicted labels. These decision-based approaches generally aim at finding minimal adversarial perturbations with as few queries as possible.

\begin{figure}[!ht]
                    \centering
                    \includegraphics[scale=1]{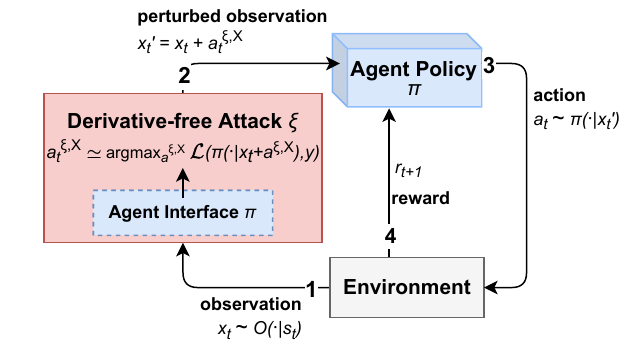}
                    \caption{Derivative-free Observation Attacks: The adversarial attack intercept the observation $x_t$, computes a perturbation $a^{\xi,X}_t$ by querying the neural network of the agent $\pi$ through an interface and applying a zeroth order optimization algorithm to maximize a loss, this perturbation is used to craft a perturbed observation $x'_t$, which is sent to the agent.}
                    \label{fig:optim_observation_attack}
                \end{figure}

                These methods seek the perturbation that maximizes the loss, to make the agent take a bad decision. Given an observation $x_t$ and a decision $y$ of the agent, a general formulation of this problem is the following:
                \begin{equation}
                    a^{\xi,X}_t = \arg \max_{a^{\xi,X}} \mathcal L ( \pi ( \cdot | x_t + a^{\xi,X}), y)
                \end{equation}
                A representation of the integration and application of derivative-free attacks in crafting observation perturbations within an RL framework is shown in Figure \ref{fig:optim_observation_attack}.
                
                The following is a non-exhaustive list of representative black-box adversarial attacks (both score-based and decision-based). Each entry gives a concise description and the setting assumed (score = access to logits/probabilities; label = only hard label).
                
                \begin{itemize}
                  \item \textbf{Zeroth-order gradient estimation.} A family of score-based methods that estimate gradients using queries. Notable examples include \textbf{Finite Difference} \\ (FD) \mycite{bhagoji2017exploring}, \textbf{Natural Evolution Strategie} (NES) \mycite{ilyas2018black}, \textbf{Simultaneous Perturbation Stochastic Approximation} (SPSA) \mycite{uesato2018adversarial}, \textbf{Bandits with Priors} \mycite{ilyas2019prior}, and \textbf{Autoencoder-based zeroth order optimization method} (AutoZOOM) \mycite{tu2019autozoom}, which differ mainly in how they reduce query cost (e.g., Gaussian perturbations, paired differences, prior-guided exploration, or dimensionality reduction). \textbf{Cost: $\sim$100s—1000s of queries}.
                  
                  \item \textbf{Coordinate and random-search methods.} These approaches perturb input coordinates or structured regions directly. \textbf{Simple Black-box Attack} (SimBA) \mycite{guo2019simple} performs greedy coordinate search, \textbf{Square Attack} \mycite{andriushchenko2020square} uses structured random perturbations, and recent extensions such as \textbf{BruSLe Attack} \mycite{vo2024brusleattack} and \textbf{Saliency Attack} \mycite{dai2023saliency} improve query efficiency and imperceptibility through sparsity or saliency priors. \textbf{Cost: $\sim$100s—1000s of queries}.
                
                  \item \textbf{Decision-based attacks.} These require only hard-label queries and iteratively approximate the decision boundary. \textbf{Boundary Attack} \mycite{brendel2018decisionbased} was the first method from this family, later improved by \textbf{HopSkipJumpAttack} (HSJA) \mycite{chen2020hopskipjumpattack}, \textbf{Sign-OPT} \mycite{cheng2019sign}, \textbf{Ray Searching} attack (RayS) \mycite{chen2020rays}, and \textbf{Geometric Decision-based Attack} (GeoDA) \mycite{rahmati2020geoda}, which progressively reduced query requirements by using binary search, sign-based optimization, ray probing, or geometric priors. \textbf{Cost: $\sim$100s—1000s of queries}.
                \end{itemize}
            
                The descriptions above are intentionally concise. Many variants and hybrids exist; attack performance (success, perturbation size) and required queries depend strongly on the threat model, perturbation budget, target model, and implementation choices. These methods are generally computationally expensive due to the number of queries required for effective perturbations.\\ \\
                Other works applied and developed derivative-free attacks specifically to RL:
                \begin{itemize}
                    \item \mycite{pan2022characterizing} applied FD in the context of RL to estimate gradients of a policy and craft perturbed observations. \textbf{Cost: $\sim$100s—1000s of queries}.
                    \item \textbf{Fractional State Attack} (FSA) \mycite{qu2020minimalistic} is a derivative-free attack that applies a genetic algorithm to select a fraction of the input space and perturb it. The fitness optimized by the GA is the KL divergence of the policy without noise w.r.t. the policy that deals with perturbed observations $D_{KL}(\pi(x_t), \pi(x_t + \eta))$. The computation of the first perturbation is computationally costly since it requires a lot a queries to maximize the fitness. But next, FSA starts the genetic process from a population containing the best solutions of the previous attacked frame, which results in quicker convergence.
                    \textbf{Cost: $\sim$100s—1000s of queries}.
                \end{itemize}
                

        \subsubsection{Attacks Driven by Long-Term Adversarial Rewards}
            
            \label{sec:obs_long_reward_attack}

            Attacks driven by long-term rewards further leverage the RL framework by exploiting its multi-step planning and return-maximization capabilities:  the aim is to steer the protagonist agent toward optimizing a cumulative objective (e.g., minimizing the victim’s return or maximizing an adversarial reward) over an episode. As a result, the attacker better plans across multiple time steps, anticipates the victim’s reactions, and exploits environment dynamics to produce more effective and transferable attacks. However, this is usually at the cost of a greatly increased cost for preparing the attack, given the protagonist agent. Still,  after this often-expensive training phase, the learned policy can be deployed as a low-cost inference procedure to generate online perturbations.

            \paragraph{Adversarial Policies (Black-Box)}\ \\
                \label{sec:obs_adv_policy_attack}
                
                Adversarial policies for observation perturbations are \textit{neural networks} that produce perturbations conditioned on the current observation. The neural network $\xi$ outputs $a^{\xi,X} = \xi(x_t)$ to craft the adversarial observation $x'_t = x_t + a^{\xi,X}$, this yields an observation-attacking policy.
                
                Representative designs include: \textbf{Optimal Attack on Reinforcement Learning Policies} (OARLP) \mycite{russo2021towards}, \textbf{Adversarial RL Generative Adversarial Network} (AdvRL-GAN) \mycite{yu2022natural}, and \textbf{Alternating Training with Learned Adversaries} (ATLA) \mycite{zhang2021robust} are concurrent works that introduced the same principle of training an Adversarial Policy with RL to generate adversarial perturbations. In these works, the adversary uses the same observation as the agent, but it could easily be extended to cases where the adversary has access to additional data of the environment or agent. Being black-box in nature, these methods only require the output of the agent model for a given input and do not need further information from the agent model. In these works, the adversarial reward to train the adversary is the opposite of the agent's reward (untargeted attacks), but it could be any other targeted adversarial reward. The \textbf{Behavior Imitation Attack} (BIA) \mycite{yamabe2024robust} is a recent variant of adversarial policy attacks that aims not to degrade performance directly but to steer the agent toward a target behavior. Unlike OARLP, AdvRL-GAN, or ATLA, which optimize adversarial rewards through reinforcement learning, BIA trains the adversarial policy via generative adversarial imitation learning (GAIL) to reproduce demonstration trajectories of a target policy, by relying on state-action occupancy matching. \textbf{Cost: First train an adversarial policy, then do 1 forward pass to attack}.
                
\begin{figure}[!ht]
                    \centering
                    \includegraphics[scale=1]{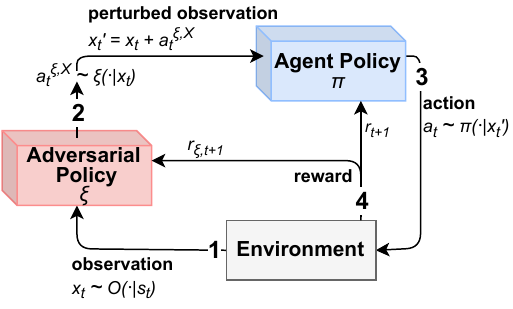}
                    \caption{Adversarial Policy Observation Attacks: The adversarial policy intercepts the observation $x_t$, computes a perturbation $a^{\xi,X}_t$ by a forward pass in its neural network, this perturbation is used to craft a perturbed observation $x'_t$, which is sent to the agent. The adversarial reward is sent to the adversarial policy to be trained.}
                    \label{fig:adversarial_policy_observation_attack}
                \end{figure}

                A representation of the integration and application of Adversarial Policies in crafting observation perturbations within an RL framework is shown in Figure \ref{fig:adversarial_policy_observation_attack}.

            \paragraph{Augmented Adversarial Policies (White-Box)}\ \\
                \label{sec:obs_adv_policy_attack_augmented}

                Adversarial Policies can also be augmented with specific white-box techniques that can improve their performance to be more effective. The first approach in this category is the \textbf{Adversarial Transformer Network} (ATN) method, as developed and studied by \mycite{baluja2018learning,tretschk2018sequential} shows that training an adversarial policy can also involve utilizing the gradients of the agent model. In this approach, the adversary is trained to maximize an adversarial reward $r_\xi$. Given the adversarial reward, the agent's loss is back-propagated to the input layer, which corresponds to the adversary's output. And the loss is subsequently back-propagated to the parameters of the adversary to be updated. The goal of this approach is to use the knowledge of the agent about the environment to improve the quality of the training of the adversary. This technique effectively trains the adversary to generate perturbations that counteract the agent's tendencies. During training, this method is considered white-box as it relies on the agent model's gradients. However, at inference time, it works as a black-box method with the simple usage of the trained policy. \textbf{Cost: First train an adversarial policy with additional backward passes, then do 1 forward pass to attack}.

\begin{figure}[!ht]
                    \centering
                    \includegraphics[scale=1]{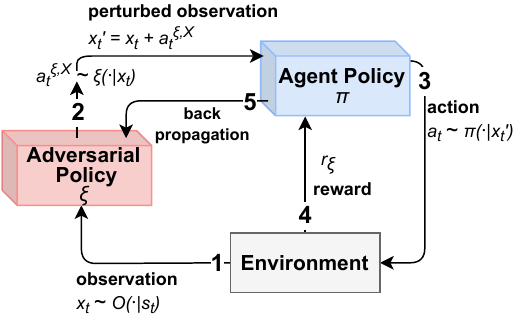}
                    \caption{ATN Observation Attacks: The adversarial policy intercepts the observation $x_t$, computes a perturbation $a^{\xi,X}_t$ by a forward pass in its neural network, and this perturbation is used to craft a perturbed observation $x'_t$, which is sent to the agent. The agent applies an action and then receives an adversarial reward $r_\xi$. The agent back-propagates the loss given $r_\xi$ to its inputs, and then it is back-propagated in the adversarial parameters to be updated.}
                    \label{fig:adversarial_policy_atn_observation_attack}
                \end{figure}

                A representation of the integration and application of the ATN in crafting observation perturbations within an RL framework is shown in Figure \ref{fig:adversarial_policy_atn_observation_attack}.

                A second augmented policy approach is the \textbf{Policy Adversarial Actor Director} (PA-AD) method \mycite{sun2022strongest}. The idea is to craft an attack in two steps. First, the adversarial direction is computed by an RL adversary agent, the \textit{director adversary}, that gives the direction of the perturbation wanted in the policy space (the target action). Then, this direction is given as a target to the \textit{actor adversary}, which uses an adversarial attack on the observation targeting that direction.
                The \textit{director adversary} is an agent trained by RL that outputs a direction (the target action), and the \textit{actor adversary} is a gradient attack following the direction given by the \textit{director adversary}. This combined approach is a white-box attack, since even if the \textit{director adversary} is trained, the \textit{actor adversary} always requires the agent's gradients to compute the perturbation. \textbf{Cost: First train an adversarial policy, then do 1 backward pass and 1 forward pass to attack}.
                
\begin{figure}[!ht]
                    \centering
                    \includegraphics[scale=1]{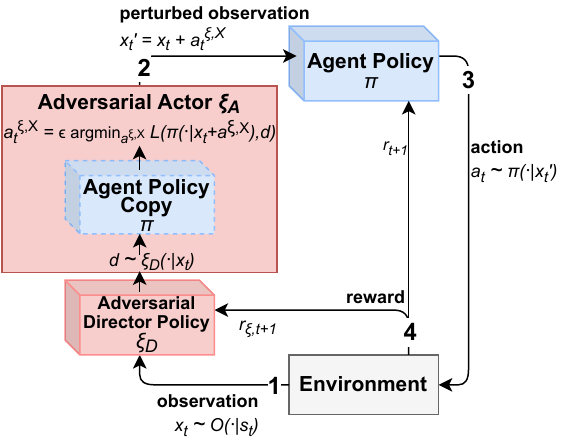}
                    \caption{PA-AD Observation Attacks: The adversarial policy intercept the observation $x_t$, the director adversary computes a direction in the policy space by forward pass in its neural network, the actor adversary computes a perturbation $a^{\xi,X}_t$ by direct optimization, this perturbation is used to craft a perturbed observation $x'_t$, which is sent to the agent. The adversarial reward is sent to the director adversary to be trained.}
                    \label{fig:adversarial_policy_paad_observation_attack}
                \end{figure}

                A representation of the integration and application of the PA-AD method in crafting observation perturbations within an RL framework is shown in Figure \ref{fig:adversarial_policy_paad_observation_attack}.

        \paragraph{Gradient Attacks (White-Box)}\ \\
            \label{sec:obs_gradient_attack_reward}
            
            Some other gradient attacks optimize an Adversarial Reward:

            \begin{itemize}
                \item \textbf{Observation Attack on the Critic Network} (OACN) \mycite{schott2022improving} is an attack inspired by FGSM. It builds a perturbation based on the estimates of the expected future rewards given by the value critic network of a PPO agent. Since the critic represents the expected cumulative reward from any state, this method allows for simulating a costly trained adversarial policy, at the cost of a single backward pass on the critic network. \textbf{Cost: 1 backward pass}.
                \item The \textbf{Static Reward Impact Map based Attack} (SRIMA) \mycite{chan2020adversarial} estimates the importance of each feature based on how much changing it affects the cumulative reward. It needs prior runs over several episodes with perturbation on subsets of features across the observation, and applies an adversarial attack. Then, the impact in terms of rewards is measured for the application of attacks for each subset. The subset that decreases the rewards the most when applying adversarial attacks is selected and will be used for the following episodes. \textbf{Cost: First run several episodes to select the subset, then use a gradient attack $\sim$10s–100s of backward passes}.
            \end{itemize}


    \subsection{Dynamics Attacks}
        
        \label{sec:dynamics_alteration}
        This section presents an overview of Dynamics Alteration Attacks for RL agents, which are methods that alter the transition function of the POMDP. These are useful for simulating the mismatch between the dynamics of a deployment environment compared to the dynamics of the training environment, and they can be used during the training of the agent to improve its robustness to unpredictable changes in the environment dynamics.
        
        As described in Section \ref{sec:altered_dynamics}, their goal is to produce an alteration of the transition function by producing a perturbation $a^{\xi}$ at a certain state $t$ with the current state being $s_t$. This perturbation $a^{\xi}$ applied to any element of the transition function will have the consequence of leading to an alternative next state $\widetilde{s}_{t+1}$, which is different than the original next state $s_{t+1}$ that would have been produced without alteration. As previously shown in Figures \ref{fig:state_perturbation}, \ref{fig:action_perturbation}, \ref{fig:transition_perturbation} and  \ref{fig:next_state_perturbation} in Section \ref{sec:altered_dynamics}, to achieve this goal the attack can either:
        \begin{itemize}
        \item compute a perturbation $a^{\xi,A}$ to craft a perturbed action $a'_t \sim \Psi^A(\cdot|a_t,a^{\xi,A}_t)$. 
        \item compute a perturbation $a^{\xi,S}$ to craft a perturbed state $\widetilde{s}_t \sim T_\Psi^S(\cdot|s_t,a^{\xi,S}_t)$.
        \item compute a perturbation $a^{\xi,T}$ to directly alter the transition function $T^\Omega$ to induce alternative next state $\widetilde{s}_{t+1} \sim T_\Psi(\cdot|s_t,a_t,a^{\xi,T}_t)$). 
        \item compute a perturbation $a^{\xi,S+}$ to craft a perturbed next state \mbox{$\widetilde{s}_{t+1} \sim T_\Psi^{S+}(\cdot|s_{t+1},a^{\xi,S+}_t)$}.
        \end{itemize}
        
        Modifying the transition dynamics represents a fundamentally different intervention from altering the observations. Observation-level perturbations affect only the time step at which they are applied: while they may lead the agent to explore different regions of the state space in the long run, their immediate effect is confined to a single perception and decision. In contrast, alterations to the transition dynamics have inherently persistent consequences: a perturbation introduced at one point in the trajectory can reshape the subsequent sequence of states, influencing the overall trajectory even if it does not immediately modify the agent’s policy.

        In the following of this section, we first discuss methods that are designed to minimize the rewards obtained by the agent by altering the transition, and then we discuss methods that are designed to deviate from the policy of the agent.
        \subsubsection{Attacks Driven by Long-Term Adversarial Rewards}
            
            \label{sec:dynamics_long_reward_attack}

            Maximizing an Adversarial Reward is the preferred way for applying dynamics perturbations in RL.

            \paragraph{Adversarial Policies (Black-Box)}\ \\
                \label{sec:dynamics_adversarial_policy_attack}

                Adversarial policies for dynamics perturbations are \textit{neural networks} that produce perturbations conditioned on the current observation. The neural network $\xi$ outputs $a^{\xi} = \xi(x_t)$ to perturb environment dynamics. Representative designs include: \textbf{Robust Adversarial Reinforcement Learning} (RARL) \mycite{pinto2017robust}, which is the first framework to apply the principle of adversarial policies to craft effective perturbations in the environment dynamics in order to maximize an adversarial reward.
                The type of perturbation permitted is defined by the action space $A^{\xi,T}$ of the adversarial policy. At each time step $t$, the adversary selects an action $a^{\xi,T}_t$, which is applied to the environment simultaneously with the agent’s action $a_t$ through a modified transition function $T_\Psi(\cdot|s_t, a_t, a^{\xi,T}_t)$. In RARL’s formulation, the training of the adversary is untargeted—that is, the adversarial reward is simply defined as the negation of the agent’s reward, $r_\xi = - r$. 
                
                \begin{figure}[!ht]
                    \centering
                    \includegraphics[scale=1]{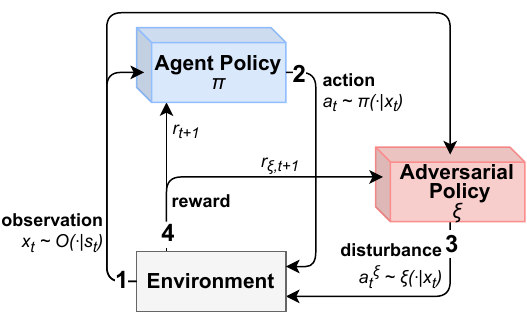}
                    \caption{Adversarial Policy Dynamics Attack: The agent and the adversary get an observation $x_t$ of the environment. The agent chooses the action $a_t$ to apply and the adversary chooses the perturbation $a^\xi_t$ to apply to alter the dynamics. The step function of the environment is run incorporating both agent and adversarial action, $s_{t+1} \sim T(\cdot | s_t,a_t,a_t^\xi)$. The adversarial reward is sent to the adversarial policy to be trained.}
                    \label{fig:adversarial_policy_dynamics_attack}
                \end{figure}

                \textbf{Risk Averse RARL} (RA-RARL) \mycite{pan2019risk} augments RARL with an explicit risk term. They use an ensemble to estimate uncertainty/risk (variance of value estimates) for the adversary to optimize a risk-seeking objective.
                
                \textbf{Env-Search} \mycite{pan2022characterizing} is another variant where the adversarial reward is not based on the actual reward of the agent but is defined as the inverse of the distance between the state after perturbation to a target state $r_\xi = 1/\|s_t-s_{target}\|_p$. 

                RARL, RA-RARL, and Env-Search have been introduced as adding an adversarial action to an augmented version of the transition function $T_\Psi$ as previously shown in Figure \ref{fig:transition_perturbation}. Note, however, that they can also be used to generate perturbations on the state $s_t$, the next state $s_{t+1}$, and also the action $a_t$ as distinct in Section \ref{sec:altered_dynamics}. A representation of the integration and usage of Adversarial Policies Attacks to add transition perturbations in an RL context is shown in Figure \ref{fig:adversarial_policy_dynamics_attack}.

                The \textbf{Probabilistic Action Robust MDP} (PR-MDP) method \mycite{tessler2019action} follows the same principle as other adversarial policy approaches, training an adversarial policy to generate perturbations, but applies it to the agent’s actions, as shown in Figure \ref{fig:action_perturbation}. In contrast, \mycite{schott2022improving} implemented the adversarial policy RARL, applying it to state perturbations, as illustrated in Figures \ref{fig:state_perturbation} and \ref{fig:next_state_perturbation}.

                All these methods are fundamentally similar: they are adversarial policies that perturb different elements depending on the adversarial action space: transition functions, states, or actions. Perturbing transitions or states requires the environment to provide “levers” that the adversarial policy can activate, which often means implementing adapted transition functions. This constraint does not exist when perturbing actions, since the actions themselves serve as direct levers. However, action-based attacks may have fewer ways to influence the environment’s dynamics compared to state- or transition-based perturbations. \textbf{Cost: All these methods require first to train an adversarial policy, then do 1 forward pass to attack}.

                Adversarial Policies have also been applied in the context of two-player games by \mycite{gleave2020adversarial,wang2022adversarial,wu2021adversarial}, and also advanced techniques as \textbf{Information-Set Monte Carlo Tree Search Best Response} (ISMCTS-BR) \mycite{timbers2022approximate}.

            \paragraph{Augmented Adversarial Policies (White-Box)}\ \\
                \label{sec:dynamics_augmented_adversarial_policy_attack}
                
                \textbf{White-Box Adversarial Policy} (WB-AP) \mycite{casper2022red} is another method for training adversarial policies, but the adversary has white-box access to the agent's internal data. For example, the adversary takes as input the same observation as the agent, concatenated with the action, the value estimates, and the latent activation (action logits) of the agent. This enables improving the attack effectiveness of the perturbations since the adversary can learn to adapt the perturbations to the internal states of the agent that is attacked. It can be applied to any observation and action spaces. \textbf{Cost: first train an adversarial policy, then do 1 backward pass and 1 forward pass to attack}.
                
\begin{figure}[!ht]
                    \centering
                    \includegraphics[scale=1]{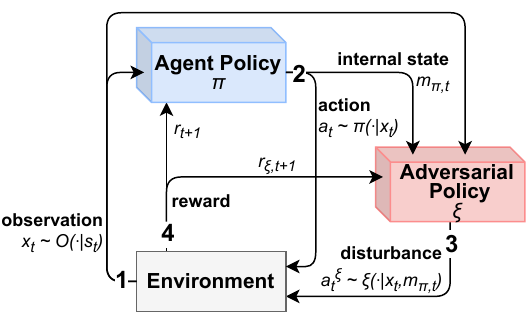}
                    \caption{White-Box Adversarial Policy Dynamics Attack: The agent gets an observation $x_t$ of the environment and chooses the action $a_t$ to apply. The adversary gets the observation and some white-box internal state of the agent and chooses the perturbation $a^\xi_t$ to apply to alter the dynamics. The step function of the environment is run incorporating both agent and adversarial action, $s_{t+1} \sim T(\cdot | s_t,a_t,a_t^\xi)$. The adversarial reward is sent to the adversarial policy to be trained.}
                    \label{fig:wb_adversarial_policy_dynamics_attack}
                \end{figure}

                A representation of the integration and usage of Adversarial Policies Attacks to add transition perturbations in an RL context is shown in Figure \ref{fig:wb_adversarial_policy_dynamics_attack}.

            \paragraph{Adversarial Domain Randomization (Black-Box)}\ \\
                \label{sec:dynamics_adversarial_domain_randomization}

                As mentioned in Section~\ref{sec:gen_purpose}, \textbf{Domain Randomization} (DR) aims to introduce perturbations into the MDP through \textbf{parametric random distributions} defined within an uncertainty set around a reference MDP. The goal is to ensure that the targeted—yet unknown—environment is included among the MDPs considered during training, similarly to our adversarial training formulation introduced in Section~\ref{sec:formalization}. 
                However, unlike adversarial training, where attackers craft perturbations on-the-fly at each step of the training episodes, DR determines a specific parametrization of the training MDP at the beginning of each episode. Classical DR typically employs uniform distributions over selected MDP parameters, but more recent works propose to adversarially learn parametric random distributions that deliberately hinder the agent’s performance. The randomized environment parameters may range from simple scalar values that scale the transition dynamics to high-dimensional vectors that significantly affect the agent’s behavior\footnote{In the latter case, these parameters may even correspond to those of a neural network that generates perturbations at each step, effectively resembling the adversarial policies described in Section~\ref{sec:dynamics_adversarial_policy_attack}, with stochastically sampled neural attackers.}. 
                
                In most DR settings, however, the uncertainty set remains limited to a small collection of domains defined over scalar parameters that modify the transition dynamics (e.g., force scales, friction/mass offsets, action delays). Following this line, \mycite{mehta2020active} introduced \textbf{Active Domain Randomization} (ADR), which replaces uniform sampling with a learned sampler over environment parameters, represented as a set of particles encoding MDP parametrizations. These particles are trained via Stein Variational Policy Gradient (SVPG) \mycite{liu2017stein} to select informative and challenging dynamics that induce significant behavioral changes in the protagonist agent. This is achieved through a discriminator-based reward function trained to distinguish between trajectories from the reference and perturbed environments.
                Intuitively, a particle configuration is rewarded for identifying regions of the randomization space that yield environment instances where the agent behaves differently from the reference case, while maintaining sufficient diversity among sampled configurations. This results in a parametric randomization distribution that concentrates probability mass on the most challenging instances, thereby improving robustness and transfer.
                
                Further focusing on worst-case parameter identification, \mycite{tanabe2022max} proposed the \textbf{Max-Min Twin Delayed Deep Deterministic Policy Gradient} (M2TD3) framework, in which the critic network is conditioned not only on the state but also on the uncertainty parameters. This augmented critic estimates expected cumulative rewards under varying dynamics conditions, enabling gradient-based refinement of the uncertainty parameters toward the most adversarial configurations within the set. \textbf{Cost: first learn the parameters of the random distributions, then sample from the distribution to attack}.

            \paragraph{Derivative-Free Attack (black-box)}\ \\
                \label{sec:dynamics_long_derivative_free_attack}

                The \textbf{Common Dominant Gradient} attack (CDG) \mycite{chen2018gradient} crafts adversarial environment configurations. It identifies sensitive regions along the agent's trajectories (so-called \textit{gradient bands}, which are not related to gradient attacks) and modifies the environment layout—e.g., adding or relocating obstacles—to steer or trap the agent and degrade its path-finding performance. Candidate modifications are generated by a constructive/search procedure and are evaluated by simulating the agent (or running a planner) to measure impact on trajectories and rewards; the strongest candidates are retained. Thus, the attack relies on environment manipulation plus repeated forward evaluation of the policy/planner. This method enables direct modification of the state of the environment but is limited to grid world or graph environments. \textbf{Cost: $\sim$10s–100s of candidates, evaluations $\sim$1000s-10$^4$s of env steps}.

            \paragraph{Gradient Attacks (White-Box)}\ \\
                \label{sec:dynamics_long_gradient_attack}

                Other approaches have been proposed to generate perturbations in the environment dynamics following a long-term reward based on gradient attacks.
                
                \textbf{Myopic Action Space} Attack  (MAS), \textbf{Look-ahead Action Space} Attack (LAS) \mycite{lee2020spatiotemporally,tan2020robustifying} are methods that apply the principle of gradient attacks to generate perturbed actions that will be applied to the environment. These methods are designed for RL agents that use an Actor and a Q-critic network that estimate the Q-value of the observation-action tuple. The idea is to apply a gradient attack on the Q-Critic network by computing the gradient of the Q-value on the input action to craft a perturbation $a^{\xi,A} = - \epsilon \nabla_a Q(x_t,a_t)$. The perturbed action $a' = a + a^{\xi,A}$ minimizes the Q-value output by the Q-network. \textbf{Cost: 1 backward pass}.
                
                LAS is an extension of the MAS method, which computes a perturbation to apply over a sequence of future states to improve the long-term impact of the attacks. LAS requires specific conditions to be applied, such as having a copy of the environment that is resettable to any state. \textbf{Cost: $\sim$10s of backward passes and env steps}.
                
\begin{figure}[!ht]
                    \centering
                    \includegraphics[scale=1]{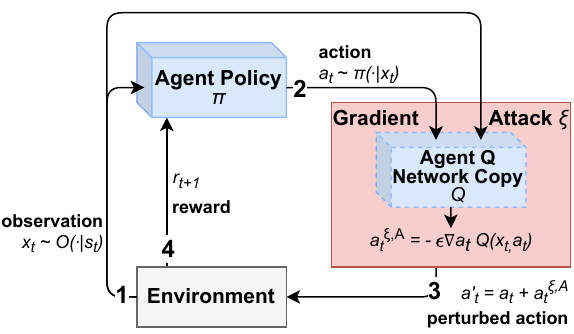}
                    \caption{MAS Dynamics Attack on the Action: The agent and the adversarial attack get the observation $x_t$ of the environment. The agent chooses the action $a_t$ to apply, the adversarial attack computes the gradient of a copy of the Q-Critic Network of the agent with respect to the action $a_t$, and this gradient is used as perturbation $a^{\xi,A}_t$ to add in the action $a'_t = a_t + a^{\xi,A}_t$. The step function of the environment is run starting from the state with the perturbed action, $s_{t+1} \sim T(\cdot|s_t,a'_t)$.}
                    \label{fig:gradient_dynamics_attack_action}
                \end{figure}

                A representation of the integration and usage of the MAS attack in an RL context is shown in Figure \ref{fig:gradient_dynamics_attack_action}.
                
                \textbf{Environment Attack based on the value-Critic Network} (EACN) \mycite{schott2022improving} applies the principle of perturbing the environment dynamics by modifying the underlying state, depending on the Critic Network. Given state $s$ and observation $x = O(s)$, EACN computes the gradient on the input of the Value-Critic Network $V$ to minimize the value estimates (expected reward). The value-Critic Network evaluates the Value function of the environment given the policy of the agent, so the method generates a perturbation of the observation $a^{\xi,X}$ which could be used to craft an adversarial observation $x' = x + a^{\xi,X}$ with $V(x') < V(x)$. This perturbation of the observation is then mapped to a perturbation $a^{\xi,S}$ of the state used to create a perturbed state $\widetilde{s} = s_t + a^{\xi,S}$ with the property $x' = O(\widetilde{s})$. The method requires a Value-Critic Network as in the PPO algorithm, but any other RL approaches can be considered by just adding the training of a Value-Critic Network on the resulting policy. \textbf{Cost: 1 backward pass}.
                        
                The main advantage of these MAS, LAS, and EACN methods over adversarial policies is that they avoid the need to train an adversary. However, MAS and LAS limited to action perturbations, while for EACN in addition to the need to have activatable levers in the environment to apply perturbations, these levers need to have a one-to-one correspondence with elements of the observations to be mapped to observations through a function $M(o) \simeq O^{-1}(o)$ that may be difficult to estimate. \mycite{schott2022improving} restricts their experiments to environments with a one-to-one correspondence between levers and some elements of the observations.\\
            
\begin{figure}[!ht]
                    \centering
                    \includegraphics[scale=1]{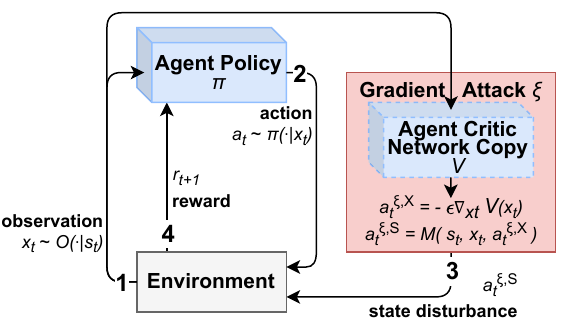}
                    \caption{EACN Dynamics Attack: The agent and the adversarial attack get the observation $x_t$ of the environment. The agent chooses the action $a_t$ to apply, and while the adversarial attack computes the gradient on the inputs of a copy of the Value-Critic Network $V$ of the agent. This gradient is used as a perturbation $a^{\xi,X}_t$ of the observation, which is mapped to a perturbation $a^{\xi,S}_t$ of the state. The adversarial state $\widetilde{s}_t = s_t + a^{\xi,S}_t$ is crafted. The step function of the environment is run starting from the perturbed state with the agent action, $s_{t+1} \sim T(\cdot|\widetilde{s}_t,a_t)$.}
                    \label{fig:gradient_long_dynamics_attack_state}
                \end{figure}

                A representation of the integration and usage of EACN for dynamic attacks in an RL context is shown in Figure \ref{fig:gradient_long_dynamics_attack_state}.
                
        \subsubsection{Attacks Driven by Short-Term Divergence Metrics}
            
            \label{sec:dynamics_short_divergence_attack}

            Optimizing a Divergence Metric is less common for applying Dynamic perturbations in RL.

            \paragraph{Gradient Attacks (White-Box)}\ \\
                \label{sec:dynamics_short_gradient_attack}
                
                \textbf{Environment Attack based on the Actor Network} (EAAN) \mycite{schott2022improving} perturbs the environment dynamics by modifying the underlying state depending on the divergence of the Actor Network. Given a state $s$ and observation $x = O(s)$, EAAN uses a gradient attack on the actor (policy) network of the agent to generate a perturbation $a^{\xi,X}$ of the observation which could be used to craft an adversarial observation $x' = x + a^{\xi,X}$ increasing a divergence metric $D\big(\pi(x'),\pi(x)\big)$. This perturbation of the observation is then mapped to a perturbation $a^{\xi,S}$ of the state used to create a perturbed state $\widetilde{s} = s + a^{\xi,S}$, ensuring the property $x' = O(\widetilde{s})$. \textbf{Cost: 1 backward pass}.
                
                Compared to EACN presented in the previous section, EAAN  does not require the knowledge of a value-critic network. However, as it is only focused on punctual changes of agent decision, it is usually less effective in the long term. Dynamics changes it induces usually lead to simple oscillations of the agent (e.g., alternating left and right directions in navigation or control tasks), without really impacting its global behavior. Defense against such attacks can nevertheless incitate the agent to reach safer states, where manipulation is more difficult (i.e., where minor state or transition changes cannot impact its decision), but tuning the level of attack can be tricky, and unexpected behaviors can emerge.  In contrast, defense against attacks such as EACN can prevent reaching undesirable states, where decisions, despite being potentially secured, inevitably lead to unwanted areas of the environment. In such, both kinds of approaches can be considered as complementary.
                
\begin{figure}[!ht]
                    \centering
                    \includegraphics[scale=1]{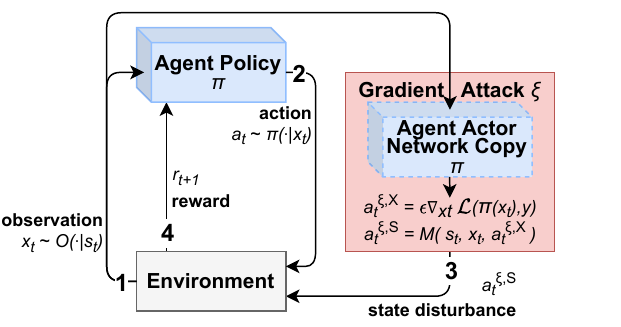}
                    \caption{EAAN Dynamics Attack: The agent and the adversarial attack get the observation $x_t$ of the environment. The agent chooses the action $a_t$ to apply and while the adversarial attack computes the gradient on the inputs of a divergence loss of a copy of the Actor Network $\pi$ of the agent. This gradient is used as a perturbation $a^{\xi,X}_t$ of the observation, which is mapped to a perturbation $a^{\xi,S}_t$ of the state, the adversarial state $\widetilde{s}_t = s_t + a^{\xi,S}_t$ is crafted. The step function of the environment is run starting from the perturbed state with the agent action, $s_{t+1} \sim T(\cdot|\widetilde{s}_t,a_t)$.}
                    \label{fig:gradient_short_dynamics_attack_state}
                \end{figure}

                A representation of the integration and usage of EAAN for dynamics attacks in an RL context is shown in Figure \ref{fig:gradient_short_dynamics_attack_state}.
                
    \subsection{Attacks Classification Summary}
        \label{sec:attack_classif}

        All attacks detailed above are classified in the following Table \ref{tab:adv_attacks}:

        \begin{table}
        \small
        \centering
        \renewcommand{\arraystretch}{1.1}
        \setlength{\tabcolsep}{0.2em}
        \begin{tabular}{|c|c|c|c|c|c|c|}
                
            \cline{1-6}
                   \textbf{\specialcell{Component\\Alteration}}       &     \textbf{Objective}      &     \textbf{Category} &  \textbf{\specialcell{Model\\Knowledge}} &  \textbf{\specialcell{Perturbed \\ Element}}  &  \textbf{Method} \\
            \cline{1-6}
            \multirow{27}{*}{\shortstack{Observations\\Alteration \\ \ref{sec:observation_alteration}}} & \multirow{21}{*}{\shortstack{Divergence\\Metrics\\ \ref{sec:obs_short_divergence_attack}}}  & \multirow{14}{*}{\shortstack{ Gradient Attacks \\ \ref{sec:obs_gradient_attack} }} & \multirow{14}{*}{white-box} & \multirow{14}{*}{\shortstack{observation }} & FGSM, FGM   \\
                                          &                             &   &  & &  BIM, PGD         \\
                                          &                             &   &  & &  MI-FGSM           \\
                                          &                             &   &  & &  AutoAttacks, APGD  \\
                                          &                             &   &  & &  FW          \\
                                          &                             &   &  & &  JSMA, VFGA          \\
            \cdashline{6-6}
                                          &                             &  &  & &  EA  \\
                                          &                             &  &  & &  MAD  \\
                                          &                             &  &  & &  DAPGD \\
                                          &                             &  &  & &  NM-FGM \\
                                          &                             &  &  & &  OAAN     \\
            \cdashline{6-6}
                                          &                             &   &  & &  DeepFool, SDF     \\
                                          &                             &   &  & &  DDN, FMN           \\
                                          &                             &   &  & &  C\&W,  FAB, ALMA    \\
            \cline{3-6}
                                          &                             & \multirow{7}{*}{\shortstack{ Derivative-Free Attacks \\ \ref{sec:obs_derivative-free_attack} }} & \multirow{7}{*}{black-box} & \multirow{7}{*}{\shortstack{observation }}  &  FD, NES, SPSA \\
                                          &                             &   &  & &  Bandit, AutoZoom    \\
            \cdashline{6-6}
                                          &                             &   &  & &  SimBA, Square    \\
                                          &                             &   &  & &  BrusLe, Saliency    \\
            \cdashline{6-6}
                                          &                             &   &  & &  Boundary, HSJA    \\
                                          &                             &   &  & &  Sign-OPT, RayS, GeoDA    \\
            \cdashline{6-6}
                                          &                             &  &  & &  FSA  \\
            \cline{2-6}
                                          & \multirow{8}{*}{\shortstack{Adversarial\\Rewards\\  \ref{sec:obs_long_reward_attack}}} & \multirow{4}{*}{\shortstack{ Adversarial Policies \\ \ref{sec:obs_adv_policy_attack} }} & \multirow{4}{*}{black-box} & \multirow{4}{*}{\shortstack{observation}}  &  OARLP     \\
                                          &                              &  &  & &  ATLA    \\
                                          &                              &  &  & &  AdvRL-GAN    \\
                                          &                              &  &  & &  BIA    \\
            \cline{3-6}
                                          &                            &   \multirow{2}{*}{\shortstack{ Augmented Adversarial \\ Policies \ref{sec:obs_adv_policy_attack_augmented}}} &  \multirow{2}{*}{white-box} & \multirow{2}{*}{\shortstack{observation}} &   ATN   \\
                                          &                             &   &  & &  PA-AD  \\
            \cline{3-6}
                                          &                            &   \multirow{2}{*}{\shortstack{ Gradient Attacks\\  \ref{sec:obs_gradient_attack_reward}}} &  \multirow{2}{*}{white-box} & \multirow{2}{*}{\shortstack{observation}} &   OACN   \\
                                          &                             &   &  & &  SRIMA  \\
            \cline{1-6}
             \multirow{16}{*}{\shortstack{Dynamics\\Alteration \\ \ref{sec:dynamics_alteration}}} & \multirow{13}{*}{\shortstack{Adversarial\\Rewards\\ \ref{sec:dynamics_long_reward_attack}}} & \multirow{4}{*}{\shortstack{ Adversarial Policies \\ \ref{sec:dynamics_adversarial_policy_attack}}} & \multirow{4}{*}{black-box} & \multirow{3}{*}{\shortstack{transition \\ state}}  & RARL \\
                                          &                             &  &  & & RA-RARL   \\
                                        &                                &  &  & & Env-Search \\
            \cline{5-6}
                                        &                               &  &  &  action   &  PR-MDP    \\
            \cline{3-6}
                                        &                               & \multirow{2}{*}{\shortstack{ Augmented Adversarial \\ Policies \ref{sec:dynamics_augmented_adversarial_policy_attack}}} &  \multirow{2}{*}{white-box}  &  \multirow{2}{*}{state} &  \multirow{2}{*}{WB-AP}     \\
                                        &                               &   &  & &      \\
            \cline{3-6}
                                        & & \multirow{3}{*}{\shortstack{Adversarial Domain \\ Randomization \\ \ref{sec:dynamics_adversarial_domain_randomization}}} & \multirow{3}{*}{black-box} & \multirow{3}{*}{\shortstack{transition}}  & \multirow{3}{*}{\shortstack{ADR \\ M2TD3}} \\
                                        &                               &  &  &     &      \\
                                        &                               &  &  &     &      \\
            \cline{3-6}
                                        &                               & \multirow{2}{*}{\shortstack{Derivative-Free Attacks\\ \ref{sec:dynamics_long_derivative_free_attack} }} &  \multirow{2}{*}{black-box}  &  \multirow{2}{*}{state} &    \multirow{2}{*}{CDG}   \\
                                        &                               &   &  & &      \\
            \cline{3-6}
                                        &                               & \multirow{2}{*}{\shortstack{  Gradient Attacks \\ \ref{sec:dynamics_long_gradient_attack} }} &  \multirow{2}{*}{white-box}  &  action &  MAS, LAS   \\
            \cline{5-6}
                                        &                               &  &   &  state &    EACN     \\
            \cline{2-6}
                    &      \multirow{3}{*}{\shortstack{Divergence\\Metrics\\ \ref{sec:dynamics_short_divergence_attack}}}      & \multirow{3}{*}{\shortstack{  Gradient Attacks \\ \ref{sec:dynamics_short_gradient_attack} }}  &  \multirow{3}{*}{white-box} &  \multirow{3}{*}{state}  &   \multirow{3}{*}{EAAN} \\
                                        &   &  &  &      &     \\
                                        &   &  &  &      &     \\
            \cline{1-6}
        
        \end{tabular}
        \caption{Characterization of Adversarial Attack Methods for Reinforcement Learning: Summary of the Content of Section \ref{sec:adv_attacks}}
        \label{tab:adv_attacks}
        \end{table}
    
        \newpage

    \subsection{Strategies for applying Attacks}
        
         Here, we discuss various strategies that extend and refine the effectiveness and usability of certain attacks, including black-box extensions, universal perturbations, stealthiness, and timing techniques, each designed to optimize the application of attacks in different contexts.
        \label{sec:strategies_attack}
        
        \subsubsection{Black-Box Extension}
        
            First, while gradient attacks are initially designed for white-box settings, strategies have been proposed to extend gradient attacks to the black-box setting. Most of these strategies use imitation learning to mimic the agent's behavior and be able to apply adversarial attacks. The \textbf{Adversarial Exploitation of Policy Imitation} (AEPI) strategy \mycite{behzadan2019adversarial} utilizes Deep Q-Learning from Demonstration to imitate the agent and use adversarial attacks on the imitated agent. Another method called \textbf{ Robust Sarsa} attack (RS) \mycite{zhang2020robust} learns the victim's Q-function without exploring, only by analyzing the victim's trajectories, enabling the application of white-box attacks on the learned Q-function. \textbf{Cost: first train an adversarial imitation policy, then the cost of the attack used}.
        
        \subsubsection{Universal Perturbations}
        
            Also, some approaches are designed for efficiency and speed that could be prioritized in strategies suitable for real-time or resource-limited applications. \textbf{CopyCAT} \mycite{hussenot2020copycat} is such a method, the idea is to pre-compute offline for a target action $a^*$ an additive perturbation mask $\delta_{a^*}$ maximizing the expected probability of the action $\mathbb{E}_{x\,\sim\,D_{\text{train}}} \big[ \pi(a^*|x + \delta_{a^*}) \big]$ over a set of pre-collected observations $D_{\text{train}}$, and then at deployment, the attacker applies the precomputed $\delta_{a^\star}$ to all observations, with no additional computation. Similar methods, like \textbf{Universal Adversarial Perturbations} (UAP) \mycite{moosavi2017universal,tekgul2022real,hirano2020simple}. In these methods, the adversary first gathers a set of observed states $D_{train}$ and sanitizes it, keeping only the ones having a critical influence on the episode. Then, the additive perturbation mask is computed using the Deepfool attack, and then applied when attacking. These methods are less effective since it does not craft specific perturbations for each observation, but are very cost efficient. \textbf{Cost: first initialize by computing the universal perturbations $\sim$10s-1000s of backward and forward passes, then 0}.
        
        \subsubsection{Sparsity and timing}
        
            Strategies from this category are characterized by their focus on attack timing, sparsity, and the preservation of stealth. Designed to reduce detectability while enhancing impact, they rely on deliberate selection of the moment of intervention.
            Additionally, \mycite{kos2017delving} showed that reducing the frequency of perturbation injections can lower computational cost while maintaining attack effectiveness. Their method \textbf{K\&S} injects adversarial perturbations only when the critic network’s estimated value exceeds a threshold, focusing on high-reward moments. In contrast, the \textbf{Strategically-Timed Attack} (STA) \mycite{lin2017tactics} uses a preference metric to identify timesteps where the policy decision is most confident, attacking only those decisive states. \textbf{Cost: 1 forward pass}.
            \textbf{Weighted Majority Algorithm} (WMA) \mycite{yang2020enhanced} gathers online statistics to select the most sensitive timeframes for attacks. These statistics are initialized with warmup episodes \textbf{Cost: several episodes for warmup, then 1 forward pass}. \textbf{Critical Point Attack} (CPA) and \textbf{Antagonist Attack} (AA) \mycite{sun2020stealthy} train models to identify critical timings for injections. CPA builds a model to predict the future states and agent’s actions, assesses the damage of each possible attack strategy, and selects the optimal one, while AA learns an adversary to discover the critical moments of attacking the agent \textbf{Cost: training of a model, then 1 forward pass}.
            More recent work further combines sparse, well-timed interventions with efficient perturbation crafting to remain stealthy while minimizing computational cost \mycite{fan2025less}. All these methods reduce the number of attacks actually computed to a fraction $\alpha$ of the trajectory lengths $T$.

    \subsection{Computational Cost of Adversarial Attacks and Strategies}
        \label{sec:cost}
        
        This section compares the various categories of attacks discussed above based on their respective computational cost.
        The computational cost of each attack can be expressed in twofold:
        
        \subsubsection*{Initialization cost in $\Omega^\pi$}
        
            This corresponds to the cost required to set up the attack (i.e., training of the attacker, prior computations, etc.).  Since attack categories rely on distinct primitive operations, we decompose this full cost into three canonical units:
            {\setstretch{0.6}
            \begin{itemize}
                \item \textbf{Init steps ($T_{\text{init}}$)}: number of steps (in terms of environment interactions or policy update) to initialize (or train) the adversary.
                \item \textbf{Forward passes ($N_{\text{fwd}}$)}: number of evaluations of a model (or query) per init step. 
                \item \textbf{Backward passes ($N_{\text{bwd}}$)}: number of back-propagations of the gradient per init step.
            \end{itemize}
            }
            We report the attack initialization cost as one multi-dimensional vector
            \begin{equation}
                C_{init}=\big(T_{\text{steps}},\,N_{\mathrm{fwd}},\,N_{\mathrm{bwd}}\big)
            \end{equation}
            
        \subsubsection*{Execution cost in $\Omega^\xi$}
        
            This corresponds to the cost of attack computation during rollouts of the policy. We decompose the cost per trajectory into three similar canonical units:
            {\setstretch{0.6}
            \begin{itemize}
                \item \textbf{Execution steps ($T_{\text{exec}}$)}: number of timesteps where perturbations are computed.
                \item \textbf{Forward passes ($N_{\text{fwd}}$)}: number of evaluations (or queries) of a model per perturbation. 
                \item \textbf{Backward passes ($N_{\text{bwd}}$)}: number of back-propagations of the gradient per perturbation.
            \end{itemize}
            }
            We report the attack execution cost as a one multi-dimensional vector
            \begin{equation}
                C_{exec}=\big(T_{\text{steps}},\,N_{\mathrm{fwd}},\,N_{\mathrm{bwd}}\big)
            \end{equation}
            Forward passes may denote white-box evaluations while queries denote black-box evaluations; here, we treat them as computationally equivalent.
            \\

        \begin{table}
        \small
        \centering
        \renewcommand{\arraystretch}{1.1}
        \setlength{\tabcolsep}{0.2em}
        \begin{tabular}{|c|c|c|c|}
            
            \hline
            \textbf{Category}   & \textbf{Attacks}     & \textbf{\specialcell{Computational \\ Cost}} &   \textbf{Decription}   \\
            \hline
                \multirow{14}{*}{\shortstack{Gradient \\ Attacks}}   & FGM, MAD, NM-FGM         & \multirow{4}{*}{\shortstack{$C_{init}=(0,0,0)$ \\ $C_{exec}=(T,1,1)$}} &    \multirow{4}{*}{\shortstack{ Single step attacks:\\  one forward  and  backward  pass \\  for gradient computation.}}  \\
                         & OAAN, OACN   &  & \\
                         & EACN, EAAN   &  & \\
                         & MAS, LAS     &  & \\
            \cdashline{2-4}
                         & BIM, PGD, MI-FGM, DAPGD    & \multirow{4}{*}{\shortstack{$C_{init}=(0,0,0)$ \\ $C_{exec}=(T,{\scriptscriptstyle +},{\scriptscriptstyle +})$}} &   \multirow{6}{*}{\shortstack{Iterative attacks: \\ several iteration with one \\ forward and backward passes.}}  \\
                         & DeepFool, SDF, DDN   &  & \\
                         & FMN, C\&W,  FAB, FW      &  & \\
                         & ALMA, VFGA           &  & \\
            \cdashline{2-3}
                         & \multirow{2}{*}{\shortstack{JSMA \\ Auto-Attacks}}  & \multirow{2}{*}{\shortstack{$C_{init}=(0,0,0)$ \\ $C_{exec}=(T,{\scriptscriptstyle ++},{\scriptscriptstyle ++})$}} & \\
                         &                   &  & \\
            \cdashline{2-4}
                         & \multirow{2}{*}{\shortstack{EA}}     &  \multirow{2}{*}{\shortstack{$C_{init}=({\scriptscriptstyle +++},1,1)$ \\ $C_{exec}=(T,1,1)$}} & \multirow{2}{*}{\shortstack{Requires first to train a  prediction model.}} \\
                         &                   &  & \\
            \cdashline{2-4}
                         & \multirow{2}{*}{\shortstack{SRIMA}}     &  \multirow{2}{*}{\shortstack{$C_{init}=({\scriptscriptstyle ++},1,0)$ \\ $C_{exec}=(T,1,1)$}} & \multirow{2}{*}{\shortstack{Requires first some setup episodes.}} \\
                         &                   &  & \\
            \hline
                \multirow{8}{*}{\shortstack{Derivative-\\free \\ Attacks}}  & FD, NES, SPSA  &  \multirow{8}{*}{\shortstack{$C_{init}=(0,0,0)$ \\ $C_{exec}=(T,{\scriptscriptstyle ++},0)$}} &  \multirow{8}{*}{\shortstack{Random search methods \\ for perturbation or gradient estimation,\\ they require multiple queries.}}  \\
                         & Bandit, AutoZoom         &   &  \\
            \cdashline{2-2}
                         & SimBA, Square            &   &  \\
                         & BruSLe, Saliency         &   &  \\
            \cdashline{2-2}
                         & Boundary, HSJA           &   &  \\
                         & Sign-OPT, RayS, GeoDA    &   &  \\
            \cdashline{2-2}
                        & FSA                       &   &  \\
            \cdashline{2-2}
                         & CDG                      &   &  \\
            \hline
                \multirow{4}{*}{\shortstack{Adversarial \\ Policies}} &  OARLP, ATLA       &  \multirow{4}{*}{\shortstack{$C_{init}=({\scriptscriptstyle +++},1,1)$ \\ $C_{exec}=(T,1,0)$}} &  \multirow{4}{*}{\shortstack{Require first to train\\ an adversarial policy,\\ then do only one forward \\ pass to attack.}} \\
                        & RARL, RA-RARL         &   &  \\
                        & Env-Search   &   &  \\
                        & PR-MDP     &   &  \\
            \hline
                \multirow{4}{*}{\shortstack{Augmented \\ Adversarial \\ Policies}}    &  \multirow{2}{*}{\shortstack{ATN}}   &  \multirow{2}{*}{\shortstack{$C_{init}=({\scriptscriptstyle +++},2,2)$ \\ $C_{exec}=(T,1,0)$}} &  \multirow{2}{*}{\shortstack{Adversarial policy that requires to compute \\ the gradient of the victim, to be trained.}}  \\
                        &    &  & \\
            \cdashline{2-4}
                        & \multirow{2}{*}{\shortstack{PA-AD, WB-AP}}  &  \multirow{2}{*}{\shortstack{$C_{init}=({\scriptscriptstyle +++},2,2)$ \\ $C_{exec}=(T,2,1)$}} & \multirow{2}{*}{\shortstack{Adversarial policies that also need \\ a gradient computation to attack.}} \\
                        &   &  &  \\
            \hline
                \multirow{3}{*}{\shortstack{Adversarial \\ Domain \\ \footnotesize{Randomization}}}    &  \multirow{3}{*}{\shortstack{ADR \\ M2TD3}}   &  \multirow{3}{*}{\shortstack{$C_{init}=({\scriptscriptstyle ++},1,1)$ \\ $C_{exec}=(T,1,0)$}} &  \multirow{3}{*}{\shortstack{Adversarial domain randomization with \\ uncertainty set parameters to learn.}}  \\
                        &    &  & \\
                        &    &  & \\
            \hline
                \multirow{8}{*}{\shortstack{Strategies \\ of \\ Application}}    &  \multirow{2}{*}{\shortstack{AEPI \\ RS}}   &  \multirow{2}{*}{\shortstack{$C_{init}=({\scriptscriptstyle +++},1,1)$ \\ $C_{exec}=(T,X,X)$}} &  \multirow{2}{*}{\shortstack{Black-box extension of white-box \\ attacks by policy imitation.}}  \\
                        &    &  & \\
            \cdashline{2-4}
                        & \multirow{3}{*}{\shortstack{CopyCat \\ UAP}}  &  \multirow{3}{*}{\shortstack{$C_{init}=({\scriptscriptstyle ++},1,1)$ \\ $C_{exec}=(T,0,0)$}} &  \multirow{3}{*}{\shortstack{\footnotesize Universal perturbations use gradient attacks \\ to pre-compute universal perturbations, which \\ are applied later without extra computations.}}  \\
                        &  &  & \\
                        &  &  & \\
            \cdashline{2-4}
                        & \multirow{1}{*}{K\&S, STA}      &  $C_{init}=(0,0,0)$                         &  \multirow{3}{*}{\shortstack{Sparsity- and timing-based strategies \\ for sparse adversarial interventions. \\ $C_{exec}=(T,1,0)+(\alpha T,X,X)$}}  \\
                        & \multirow{1}{*}{WMA}      &  $C_{init}=({\scriptscriptstyle ++},1,0)$    &  \\
                        & \multirow{1}{*}{CPA, AA}  &  $C_{init}=({\scriptscriptstyle +++},1,1)$  &  \\
            \hline
        \end{tabular}
        \caption{
        Computational Cost of Attacks. \textit{Note:} Symbols $\scriptstyle +$, $\scriptstyle ++$, and $\scriptstyle +++$ denote increasing computational effort: $\scriptstyle +$ refers to iterative methods that are not instantaneous; 
            $\scriptstyle ++$ corresponds to slower iterative methods or  requiring multiple  interaction steps; 
            and $\scriptstyle +++$ indicates methods that involve the full training of an additional model.
            }
        \label{tab:attacks_cost}
        \end{table}

        Table~\ref{tab:attacks_cost} summarizes the considered attacks and strategies, categorized by their computational approach and broken down by approximate cost. Clear patterns emerge in the balance between computational effort and flexibility. \textbf{Single-step gradient attacks} are lightweight, requiring only one forward and backward pass per perturbation, whereas \textbf{iterative methods} increase cost substantially through repeated gradient evaluations. \textbf{Derivative-free} and black-box approaches, while avoiding backpropagation, incur high query costs due to repeated model evaluations during search or gradient estimation \mycite{gluch2021query}. \textbf{Adversarial-policy} methods, in contrast, shift computation to initialization: they require extensive offline training but are inexpensive at runtime, since attacks reduce to simple forward passes. \textbf{Augmented adversarial policies} lie between these extremes, combining adversarial policies initialization and online gradient computations. \textbf{Black-box extension strategies} concentrate their cost in initialization, training surrogates, while the runtime cost depends on the specific attack that will be applied on this surrogate model. \textbf{Universal perturbation} strategies incur their cost entirely offline when computing the universal mask; at deployment, they apply the precomputed perturbation with negligible extra computation. Finally, \textbf{Sparsity and timing-based strategies} reduce the frequency of interventions, but the per-intervention cost remains that of the chosen attack; only the total execution burden is scaled down by the fraction of timesteps attacked.

        Overall, the table highlights distinct ``power relations’’ among families: iterative white-box methods dominate in precision but at the highest gradient cost; query-based and black-box attacks trade accuracy for scalability; and policy-based or precomputed approaches amortize heavy initialization over many deployments. This breakdown clarifies which attacks are computationally demanding in practice and which are better suited for real-time or resource-limited settings \mycite{croce2020reliable, cina2025attackbench, apruzzese2023real}.
        
        To compare attacks quantitatively, one might define a normalized scalar proxy of computational effort that combines forward, backward, and training components:
        \begin{equation}
            \mathcal{S}_{\text{total}}
            \;=\;
            T_{\text{exec}} \cdot
            \!\Big(
              N_{\mathrm{fwd}}\,c_{\mathrm{fwd}}
              + 
              N_{\mathrm{bwd}}\,c_{\mathrm{bwd}}
            \Big)
            +
            T_{\text{init}} \cdot
            \!\Big(
              N_{\mathrm{fwd}}^{(\text{init})}\,c_{\mathrm{fwd}}
              +
              N_{\mathrm{bwd}}^{(\text{init})}\,c_{\mathrm{bwd}}
            \Big)
            \label{eq:scalar_cost}
        \end{equation}
        where \(c_{\mathrm{fwd}}\) and \(c_{\mathrm{bwd}}\) denote the unit cost of a forward and backward pass, respectively, 
        and \(T_{\text{exec}}\) and \(T_{\text{init}}\) the number of execution and initialization steps. 
        Typically \(c_{\mathrm{bwd}}\!\approx\!2c_{\mathrm{fwd}}\), since backpropagation roughly doubles the computation of a forward pass. 
        Initialization-heavy methods (e.g., adversarial-policy or universal attacks) have large \(T_{\text{init}}\) but small \(T_{\text{exec}}\), 
        whereas iterative gradient or query-based attacks exhibit the opposite trend.

\section{Adversarial Training}
    \label{sec:adv_training}

    Adversarial training in RL improves the robustness of RL agents against adversarial attacks. The general principle involves repeatedly exposing the agent to adversarial examples during training, similar to inoculating it against real-world attacks. This method aligns with robust control principles \mycite{dorato1987historical}, which focus on maintaining stability and performance despite uncertainties. These perturbations can either be generated using adversarial attacks on the observations or the environment dynamics. Training in such challenging environments helps the RL agent perform effectively even when faced with manipulated inputs or altered state transitions, making it more robust against post-deployment manipulations. As detailed in Section \ref{sec:formalization_dro_adv_training}, the process often involves a min-max game to minimize the maximum possible loss induced by an adversary \eqref{obj_robusttl}. This approach mirrors robust control's emphasis on preparing systems for worst-case scenarios and uncertainties. While any adversarial attack can be used, some improve robustness more effectively, and a specific training strategy can better enhance robustness.

    In this section, we will first introduce the main adversarial training strategies in \ref{sec:advtrain_strategies}, then discuss how to choose the appropriate attack according to different conditions in \ref{sec:choose_attack}, and finally discuss how to find the right balance between robustness and performance \ref{sec:balance_robustnesse_and_perf}.

    \subsection{Strategies of Adversarial Training}
        \label{sec:advtrain_strategies}
        
        Adversarial training starts with an agent $\pi_1$, classically pre-trained until convergence in a given environment $\Omega$. A first step is the initialization of the adversary $\xi$, which is described in section \ref{sec:init_policy}. Next, starting from $\xi_1$ as the initialized attacker, the following of this section describes different strategies for adversarial training, ranging from a constant adversarial attacker, to attackers that improve depending on the evolution of the protagonist behavior. We note the resulting adversarially trained policy $\pi_+$, in what follows.
        
        \subsubsection{Initialization of the Adversary}
            \label{sec:init_policy}

            \paragraph*{Adversarial Policies}\ \\
                For adversaries requiring training (such as adversarial policies), one considers the pre-training of the untrained adversary $\xi_0$, until convergence within the environment $\Omega^{\pi_1}$. The procedure is described in Algorithm \ref{algo:adversary}, resulting in a trained adversary $\xi_1$.

                \begin{algorithm}[h]
                \caption{Initial Training of the Adversary $\xi$}
                \label{algo:adversary}
                \begin{algorithmic}[1]
                    \State \textbf{Initialize:} Trained agent $\pi_1$, untrained adversary $\xi_0$
                    \State $\hat{\xi} \gets \xi_0$
                    \While{not converged} \Comment{$\hat{\xi}$ trained in $\Omega^{\pi_1}$}
                        \State \textbf{Fit} $\hat{\xi}$ \textbf{on trajectories from} $\hat{\xi}^{\Omega,\pi_1}$
                    \EndWhile
                    \State $\xi_1 \gets \hat{\xi}$
                    \State \Return $\xi_1$
                \end{algorithmic}
                \end{algorithm}

            \paragraph*{Gradient or Query Attacks}\ \\
                For adversaries that do not require parameter training (e.g. gradient or query-based attacks), we define $\xi_1$ directly as the adversarial attack that computes perturbations based on the agent $\pi_1$, by computing the gradients w.r.t. the input or querying policy $\pi_1$.

        \subsubsection{Constant Adversary}

            The first approach in adversarial training involves using a constant adversary $\xi_1$, which is optimized for the agent $\pi_1$.
            
            \paragraph*{Trained Adversaries}\ \\
                For adversaries requiring training, the constant adversarial training is the default case. The constant adversary $\xi_1$ trained against $\pi_1$ is used to generate perturbations during the adversarial training of the agent $\pi$ without updating its adversarial policy $\xi_1$. The procedure is described in Algorithm \ref{algo:constant}. After this training, the adversarially trained agent becomes $\pi_+$.
            
                \begin{algorithm}[h]
                \caption{Adversarial Training of the Agent $\pi_1$ against a constant Adversary $\xi_1$}
                \label{algo:constant}
                \begin{algorithmic}[1]
                    \State \textbf{Initialize:} Agent $\pi_1$, constant adversary $\xi_1$
                    \State $\hat{\pi} \gets \pi_1$
                    \While{not converged} \Comment{$\hat{\pi}$ trained in $\Omega^{\xi_1}$}
                        \State \textbf{Fit} $\hat{\pi}$ \textbf{on trajectories from} $\hat{\pi}^{\Omega,\xi_1}$
                    \EndWhile
                    \State $\pi_+ \gets \hat{\pi}$
                    \State \Return $\pi_+$
                \end{algorithmic}
                \end{algorithm}

            \paragraph*{Gradient or Query Attacks}\ \\
                For adversaries that do not require parameter training (e.g. gradient or query-based attacks), the procedure entails, for each iteration $k$, generating perturbations targeting $\pi_k$ while relying on the initial policy $\pi_1$ as the reference for crafting those perturbations.
                Even when the agent $\pi$ is updated, the perturbations have still to be based on the original $\pi_1$ version of the agent. The procedure is the same as described in Algorithm \ref{algo:constant}.\\

            Constant adversarial training has been applied in supervised learning by \mycite{tramer2017ensemble,wong2020fast,wang2024revisiting}, and its application in RL has been discussed in \mycite{pinto2017robust}. This approach minimizes the risk of divergence far from the distribution of the environment. One major drawback of constant adversarial training is that the agent $\pi_+$ can overfit on perturbations generated by the adversary $\xi_1$ (as a kind of reward hacking), with perturbations from $\xi_1$ possibly inefficient for the finetuned policy. As a result, the policy $\pi_+$ may remain vulnerable to other attacks or stronger adversaries. For trained adversarial policies, another major limitation is the distribution shift induced by the training of the protagonist: states encountered in trajectories from $\pi_+$ may be highly different from those of $\pi_1$, on which the adversary has been trained, likely leading to unexpected behavior of the attacker.

        \subsubsection{Co-Trained Adversary}
            \label{sec:cotrained}
            
            In co-trained schemes, the agent $\pi$ and the adversary $\xi$ both improve during training. We present the \textit{alternate} update form as the core template; the widely used \textit{simultaneous} variant is a special case of this template.
            
            \paragraph*{Alternate Adversarial Training}\ \\
                Alternate training updates the agent and the adversary in separate phases (Algorithm \ref{algo:alternate}), allowing each to approximately track a moving best response while keeping the other side constant during its own update block. 

                The commonly used simultaneous training scheme can be viewed as the limit of Algorithm \ref{algo:alternate} with very small update blocks: $N_\pi = N_\xi = 1 \quad\text{per outer iteration}$,                 so that each iteration alternates a single (or very few) gradient steps of $\pi$ and $\xi$. In many implementations, both updates are even applied on the same transition batch (e.g., a shared on-policy trajectory or the same replay minibatch), effectively yielding one outer loop step per environment episode (or per minibatch). This improves data efficiency but increases non-stationarity, which can make training less stable than the larger-block alternate form.
                
                Thus, a critical factor is the relative learning rates of the protagonist and the adversary. If the adversary changes too rapidly from one epoch to the next, the overall training process can become highly unstable. Conversely, if the protagonist is over-optimized against a fixed adversary, the system can suffer from catastrophic forgetting—overfitting to the most recent adversary and exhibiting oscillatory behavior. In practice, one typically sets $N_\xi > N_\pi$.
            
                \begin{algorithm}[h]
                    \caption{Co-training of the Agent $\pi$ against an Adversarial Policy $\xi$}
                    \label{algo:alternate}
                    \begin{algorithmic}[1]
                        \State \textbf{Initialize:} Agent $\pi_1$, Adversary $\xi_1$; learning rate $\alpha$; number of policy updates per outer iteration $N_\pi$; number of adversary updates per outer iteration $N_\xi$.
                        \State $k \gets 1$
                        \While{not converged}
                            \State $\hat{\pi} \gets \pi_k$
                            \For{$i=1$ to $N_\pi$} \Comment{agent updates with adversary frozen at $\xi_k$}
                                \State \textbf{Fit} $\hat{\pi}$ \textbf{on trajectories from} $\hat{\pi}^{\Omega,\xi_k}$ \Comment{$\hat{\pi}$ trained in $\Omega^{\xi_k}$}
                            \EndFor
                            \State $\pi_{k+1} \gets \hat{\pi}$
                            \State $\hat{\xi} \gets \xi_k$
                            \For{$j=1$ to $N_\xi$} \Comment{adversary updates with agent frozen at $\pi_{k+1}$}
                                \State \textbf{Fit} $\hat{\xi}$ \textbf{on trajectories from} $\hat{\xi}^{\Omega,\pi_{k+1}}$ \Comment{$\hat{\xi}$ trained in $\Omega^{\pi_{k+1}}$}
                            \EndFor
                            \State $\xi_{k+1} \gets \hat{\xi}$
                            \State $k \gets k+1$
                        \EndWhile
                        \State \Return $\pi_+ \gets \pi_k,\ \xi_+ \gets \xi_k$
                    \end{algorithmic}
                \end{algorithm}
            
            \paragraph*{Gradient or Query Attacks}\ \\
                For adversaries that do not require parameter training (e.g. gradient or query-based attacks), the same template applies with the $\xi$-update blocks removed. In the alternate view, the adversary uses $\pi_k$ as the reference during the agent-update block; in the simultaneous-as-extreme view, the adversary adapts to the current $\pi$ at each step, possibly using the same trajectories the agent uses for its update.
                
            \medskip
            Co-trained adversarial training has been used with observation attacks~\mycite{pattanaik2017robust,zhang2020robust,korkmaz2021adversarially} and dynamics perturbations~\mycite{tan2020robustifying,schott2022improving}. Alternate schemes (e.g., ATLA) have demonstrated improved stability in several MuJoCo tasks~\mycite{zhang2021robust}, while disturbance-agent minimax variants have been explored for dynamics robustness~\mycite{pinto2017robust,ma2018improved,pan2019risk,abdullah2019wasserstein,tessler2019action}.

        \subsubsection{Mixture of Adversaries}
        \label{sec:mixture_adversaries}
        
            Maintaining diversity in adversarial attacks helps avoid robust overfitting, where $\pi$ specializes against a narrow threat model yet remains brittle to unseen perturbations. We distinguish three paradigms that differ by where the diversity comes from and how the adversary is executed at rollout time.

            \paragraph*{(1) Self-mixture}\ \\
                A single adversary $\xi$ learns best responses but executes a mixture of its historical policies. \textbf{Neural Fictitious Self-Play} (NFSP) \mycite{heinrich2016deep} maintains a best-response head $\xi$ and an average head $\overline{\xi}$ trained by supervised learning on actions taken by $\xi$. Rollouts use the mixed policy :
                \begin{equation}
                    \sigma^\xi \;=\; (1-\eta)\,\overline{\xi} \;+\; \eta\,\xi
                \end{equation}
                This reduces cycling and stabilizes training while preserving adversarial pressure. Applied to robust driving, NFSP has been used alongside RARL \mycite{ma2018improved}. The generic mixture algorithm (Alg.\ref{algo:mixture_cotraining}) reproduces NFSP by setting the population to $\Xi=\{\xi,\overline{\xi}\}$, choosing $\mathcal{S}$ to return  weights $w=((\eta),(1-\eta))$, and recording only best-response steps ($c_t^\xi{=}1$) to train $\overline{\xi}$ via $\max \log \overline{\xi}(a|s)$ on its buffer.
            
            \paragraph*{(2) Mixture over a population of adversaries}\ \\
                The idea naturally extends to a general mixture across a population of adversaries. Maintain a pool $\Xi=\{\xi^{(1)},\dots,\xi^{(m)}\}$ (disturbance policies, observation attackers, or historical snapshots) and execute a mixture over $\Xi$ during training (Alg.\ref{algo:mixture_cotraining}). The mixture can be uniform or prioritized \mycite{vinitsky2020robust}, or computed by a meta-solver such as \textbf{Policy-Space Response Oracles} (PSRO) \mycite{lanctot2017unified} or Double-Oracle \mycite{mcaleer2021xdo}, which iteratively adds best responses and solves for a meta-strategy over the current population. This reduces exploitability to any single attacker and captures multiple threat modes \mycite{zhang2024survey,shen2021robust}.
                
                \begin{algorithm}[h]
                \caption{Co-Training of the Agent $\pi$ Against a Mixture of Adversaries}
                \label{algo:mixture_cotraining}
                \begin{algorithmic}[1]
                    \State \textbf{Initialize:} Agent $\pi_1$; adversary population $\Xi=\{\xi^{(m)}\}_{m=1}^M$; scheduler/meta-strategy $\mathcal{S}$ (uniform/prioritized/PSRO); learning hyperparameters.
                    \State $k \gets 1$
                    \While{not converged}
                      \State $w_k \gets \mathcal{S}(\pi_k,\Xi)$
                      \State Define mixture adversary $\sigma^\xi_k \gets \mathrm{Mix}(\Xi, w_k)$
                      \State Sample trajectory $\widetilde{\tau} \sim \pi_k^{\Omega,\sigma^\xi_k}$ \Comment{rollout against current mixture}
                      \State $\hat{\pi} \gets \pi_k$
                      \State \textbf{Fit} $\hat{\pi}$ \textbf{on trajectories from} $\hat{\pi}^{\Omega,\sigma^\xi_k}$ \Comment{$\hat{\pi}$ trained in $\Omega^{\sigma^\xi_k}$}
                      \State $\pi_{k+1} \gets \hat{\pi}$
                      \For{each selected $\xi^{(m)} \in \Xi$ (e.g., those used in $\widetilde{\tau}$)}
                         \State $\hat{\xi}^{(m)} \gets \xi^{(m)}$
                         \State \textbf{Fit} $\hat{\xi}^{(m)}$ \textbf{on trajectories from} $(\hat{\xi}^{(m)})^{\Omega,\pi_{k+1}}$ \Comment{or reuse the same batch, if desired}
                         \State $\xi^{(m)} \gets \hat{\xi}^{(m)}$
                      \EndFor
                      \State \textbf{(Optional) Population maintenance:} expand $\Xi$ with snapshots/new adversaries; update $\mathcal{S}$ (e.g., re-solve the meta-game).
                      \State $k \gets $ $k+1$
                    \EndWhile
                    \State \Return $\pi_+ \gets \pi_k$ \text{ (and final population $\Xi$)}
                \end{algorithmic}
                \end{algorithm}
                
            \paragraph*{(3) Post-hoc worst-episode selection}\ \\
                Rather than mixing adversaries online, EPOpt \mycite{rajeswaran2017epopt} rolls out a batch of episodes across a fixed ensemble of adversaries/environments, then selects after the fact the $\varepsilon$-percentile lowest-return episodes and updates $\pi$ using only those worst-case trajectories.

    \subsection{Which Perturbation for which Robustness? }
        \label{sec:choose_attack}

        A central challenge in adversarial training is selecting attack strategies that satisfy deployment constraints (e.g., white vs. black-box, controllable components) and rely on a threat model broad enough to prevent over-specializing the defense, while still being effective and practical to generate during training.
            
        The type of perturbation plays a central role in shaping the effectiveness of adversarial training for a given problem. It determines not only how the agent perceives and reacts to adversarial inputs, but also which aspects of its policy and representation learning are made more robust. Choosing the right perturbation type is thus a key design decision that directly impacts both robustness and task performance in different ways.
        
        \paragraph*{Perturbations of the Observations}\ \\
            Adversarial training with observation perturbations primarily targets robustness at the perception level. These perturbations challenge the agent’s sensory processing, forcing it to rely on task-relevant features rather than on noisy, misleading, or superficial cues. Depending on the type of observation, the effects on robustness can differ substantially. These can be divided into two types:
            
            \begin{itemize}
                \item \textbf{Perturbations on pixel-based observations} can deceive the agent while preserving a nearly identical visual appearance, leading to policy failures despite imperceptible changes. Adversarial training with such perturbations strengthens robustness by encouraging the agent to rely on task-relevant and invariant visual features rather than on superficial pixel-level correlations.
                \item \textbf{Perturbations on feature-vector observations} (i.e., higher-level inputs that encode semantically meaningful information) can disrupt the internal representation of the environment, causing the agent to lose access to critical information and potentially rendering the task unsolvable. Training with feature-level perturbations, however, can foster representation-level robustness, pushing the agent to extract and prioritize stable, task-critical features that remain reliable even under corrupted or noisy conditions.
            \end{itemize}
        
        \paragraph*{Perturbations of the Environment Dynamics}\ \\
            Adversarial training with perturbations of the environment dynamics targets robustness at a higher level, going beyond perception. While attacks on observations primarily improve robustness to noisy or misleading sensory input, dynamic perturbations simulate challenging situations that force the agent to navigate suboptimal or risky states. This encourages the agent to develop recovery strategies, seeking ways to return to safer or more advantageous regions of the environment. 
            To be effective without destabilizing learning, the attacker should apply smooth manipulations of the dynamics, avoiding abrupt disruptions that could cause instability or catastrophic forgetting in the policy.
            
            The potential effects on agent robustness differ depending on which component of the environment is perturbed (following components detailed in section \ref{sec:perturbation}):
            \begin{itemize}
                \item \textbf{Actions ($a_t$):} Perturbing the agent’s own actions exposes it to execution noise or missteps, forcing it to adapt policies that are resilient to small deviations in its decisions. This strengthens robustness to imprecise control and encourages strategies that are stable under small action perturbations, but does not directly improve robustness to environmental variability.
                \item \textbf{Current state ($s_t$) before transition:} Perturbing the current state is applied after the agent has made its decision but before the environment transition, placing the agent in a situation different from what it expected when choosing its action. This challenges the agent to adapt to unexpected consequences of its own decisions, fostering robustness to situations where planned actions lead to unanticipated contexts.
                \item \textbf{Transition function ($T$):} Perturbing the transition function modifies the rules governing the environment itself, representing a strong form of adversarial influence. Agents trained under such conditions learn robust policies that generalize across changes in environment dynamics, enabling them to handle a variety of scenarios and recover from unforeseen shifts in state transitions (e.g., reality gap).         
                \item \textbf{Next state ($s_{t+1}$) after transition:} Perturbing the next state occurs after the environment transition but before the next decision, meaning the agent is aware of its state but faces potentially difficult or suboptimal conditions. Training with these perturbations encourages the agent to plan recovery strategies in known challenging situations, improving robustness to environmental variability and cumulative difficulties.
            \end{itemize}

            Beyond perturbing a single modality, recent work trains agents against \textit{mixed} adversaries that perturb several modalities at once. \mycite{erdem2025learning} formalize an Action-and-State-Adversarial MDP and propose ASA-PPO, where a co-trained adversary learns to distribute a shared attack budget across both the observation and the action spaces. They show that agents trained against single-modality adversaries remain vulnerable to such mixed perturbations, whereas learning to balance the perturbation across modalities yields policies that are robust under observation-only, action-only, and combined attacks simultaneously.

    \subsection{Balancing Robustness and Performance}
        \label{sec:balance_robustnesse_and_perf}

        A key challenge in adversarial training is maintaining a balance between robustness and overall task performance. Excessively focusing on defending against adversarial perturbations can sometimes hurt the agent's performance in benign environments. A good adversarial training setup should aim to minimize the performance trade-off, ensuring that robustness does not come at the expense of overall task effectiveness.

        \subsubsection{Magnitude of Perturbation}
            \label{sec:advtrain_magnitude}
            
            Adjusting the magnitude of perturbations is one of the most challenging aspects of adversarial training. The appropriate strength depends on the modality being perturbed—pixels, feature vectors, or environment dynamics—and is typically controlled either as a norm of the difference between the original component and its perturbed version, or as a divergence between distributions (e.g., KL divergence or Wasserstein distance) induced by the perturbation. If perturbations are too mild, the agent may not encounter sufficiently challenging situations to develop robust strategies, leaving it vulnerable to stronger attacks. Conversely, if they are too strong, the task can become unsolvable, degrading learning signals or causing policy failure. Choosing the right perturbation strength is therefore crucial: it must be large enough to promote robustness, yet small enough to preserve solvability and meaningful task information.

            The following summarizes common choices and practical considerations for each case:
            
            \begin{itemize}
                \item \textbf{Pixel-based observations:} Perturbations are usually constrained using $L_\infty$ or $L_2$ norms, reflecting realistic sensor noise \mycite{huang2017adversarial,behzadan2017vulnerability}, while $L_0$-based perturbations simulate discrete pixel failures \mycite{kos2017delving}. If the perturbation is too small, the agent perceives little difference from the original input, limiting robustness gains. If it is too large, the task may become unrealistically difficult, as decisions must be made from highly noisy inputs, leading to policy failure and preventing effective learning. Carefully calibrated norms enable the agent to learn robust perception strategies while retaining access to task-relevant information.
                
                \item \textbf{Feature-vector observations:} Perturbations can be defined per feature with explicit budgets if features have heterogeneous scales, $L_0$ norm can also be used to reflect the number of features that can be altered \mycite{lin2017tactics,sun2020stealthy}. $L_\infty$, $L_2$, or $L_1$ norms can also be applied when features are homogeneous \mycite{gleave2020adversarial}. Because feature vectors encode higher-level semantic information, perturbations have a stronger impact than pixel-level changes. Even more than with pixels, the perturbation strength must be chosen carefully: too weak and robustness improvements are minimal; too strong and the task may become unsolvable. Properly tuned feature perturbations promote representation-level robustness, forcing the agent to extract stable, task-critical information even under noisy or altered conditions.
                
                \item \textbf{Environment dynamics:} Perturbations are typically defined per lever with explicit budgets, representing realistic changes for each controllable aspect of the environment \mycite{pattanaik2017robust,zhang2020robust}. $L_\infty$, $L_2$, or $L_1$ norms can also be used when levers are similar \mycite{rajeswaran2017epopt}. More generally, one can quantify the divergence between the perturbed and original transition distributions, which bounds the induced changes in expected trajectories. The magnitude of the perturbation strongly affects robustness:
            
                \begin{itemize}
                    \item If the perturbation is too weak, the agent rarely encounters challenging situations and cannot develop effective recovery strategies, limiting robustness gains.
                    \item If the perturbation is too strong, the environment may become unrealistic or unsolvable, preventing the agent from learning any meaningful policy and destabilizing training.
                    \item Carefully tuned perturbation strengths per lever allow the agent to learn policies that are robust to plausible dynamic variations, adaptively recover from suboptimal states, and generalize across a wide range of environmental conditions without overfitting to extreme scenarios.
                \end{itemize}
            \end{itemize}
                        
            The remainder of this section discusses different strategies for carefully adjusting the adversarial training process with respect to perturbation magnitude. These strategies range from simple but rigid and computationally expensive grid searches for effective bounds, to more advanced methods that dynamically adapt perturbation limits during training, or introduce regularization mechanisms to prevent catastrophic policy updates in the course of adversarial learning.

            \paragraph{Grid Search over Magnitudes}\ \\
            
                Most existing works on adversarial training rely on a grid search to adjust the magnitude of perturbations \mycite{schott2022improving} across a predefined range of values to find the optimal level. While grid search offers a systematic way to explore different perturbation strengths, it is computationally expensive and inefficient. Each level of perturbation must be evaluated multiple times across various training episodes, requiring significant computational resources, especially when the search space is large. Moreover, grid search does not adapt dynamically during training, which means that the chosen perturbation level remains fixed, even as the agent's performance and robustness evolve. This lack of flexibility often leads to suboptimal solutions, as the ideal perturbation magnitude may change depending on the agent’s progress.
            
            \paragraph{Dynamic Regulation of Magnitude}\ \\

                A more effective approach involves dynamically regulating the magnitude of perturbations based on the agent’s current performance and learning progress.
                
                Following this idea, \mycite{liu2024robust} proposed an \textbf{Adaptive Adversarial Policy} training framework (A2P-SAC) for reinforcement learning, where perturbations are applied to the agent’s actions. The framework dynamically modulates the strength of the perturbations according to the agent’s progress, aiming to balance robustness and task performance. The core idea is to adjust the attacker’s influence throughout training: when the agent performs well, the attack strength is increased to expose new vulnerabilities; when the agent’s performance deteriorates, it is reduced to maintain stable learning. In practice, the authors co-train an attacker policy jointly with the agent (implemented using SAC), while an adaptive coefficient automatically regulates the effective attack level instead of relying on manual tuning. This mechanism eliminates tedious hyperparameter adjustments, stabilizes training, and produces policies with improved transferability under dynamic shifts in MuJoCo benchmarks.

                Several other approaches for dynamic regulations of adversarial policies have been explored, \mycite{ma2018improved} introduced the approach \textbf{Semi-competitive RARL} (SC-RARL) that incorporates a constraint in the adversarial policy's loss function, encouraging the learning of smaller perturbations while optimizing for more effective attacks. In SC-RARL, the reward of the adversary is the opposite of the reward of the agent, with a penalization term that discourages large perturbations $r_\xi =  -r + \|a^{\xi}\|_p$.
                
                Other research in both classification and RL has explored the approach of dynamically regulating the magnitude of perturbations. In \textbf{Wasserstein Robust RL} (WR\textsuperscript{2}L) \mycite{abdullah2019wasserstein} the adversarial policy is updated to decrease the agent’s return while constrained to remain within an $\varepsilon$-Wasserstein ball around the previous parameters, yielding adversarial yet plausible dynamics. Related methods adopt the same principle with different update rules: \textbf{Distributionally Robust policy learning via Adversarial Generation of ENvironment} (DRAGEN) \mycite{ren2022distributionally} learns a generative model of environments and performs latent-space ascent under a distributional-robustness constraint to produce hard-but-plausible dynamics for training; in \textbf{Unsupervised Environment Design} (UED) and \textbf{Protagonist Antagonist Induced Regret Environment Design} (PAIRED) \mycite{dennis2020emergent}, a teacher policy proposes environments (parameter settings/layouts) that maximize regret between a protagonist and an antagonist, effectively acting as an adversarial policy over $\phi$. Conceptually, these approaches instantiate an embedded adversary in the transition function: the trainable environment parameters $\phi$ are the adversary’s policy parameters. \textbf{Time-Constrained Robust MDP} (TC-RMDP) \mycite{zouitine2024time} models the environment as an adversarially controlled dynamics parameter $\phi_t$ inside $T_\Psi(\cdot;\phi_t)$, with a bounded rate-of-change constraint: $\phi_t=\phi_{t-1}+u_t$, $\|u_t\|\le \rho$. The adversary is a learned policy that selects $u_t$  each step, while the constraint induces temporally correlated disturbances. Finally, the recent work \mycite{schott2026reward} introduces \textbf{Reward-Preserving attacks}, as a class of attacks that ensure that a given portion of the initial reward remains reachable, in order to avoid  overly destructive or too conservative adversarial learning strategies.

            By maintaining a balance in the magnitude of perturbations, the agent can avoid catastrophic forgetting and maintain a steady learning trajectory, even in adversarial environments.

        \subsubsection{Frequency of Perturbation}
            \label{sec:advtrain_frequency}

            The frequency of adversarial attacks is another important factor that affects both robustness and the agent’s ability to adapt. Frequent attacks can overwhelm the agent and cause it to overfit to the adversarial environment, neglecting performance in normal conditions. On the other hand, infrequent attacks may not provide enough pressure for the agent to develop resilience.
    
            One approach is to dynamically adjust the attack frequency based on the agent’s performance. Early in training, a lower attack frequency can help the agent establish a baseline level of performance before introducing adversarial conditions. As training progresses, the frequency can gradually increase, allowing the agent to adapt to more challenging environments. This incremental increase helps prevent catastrophic forgetting, where the agent might lose learned behaviors due to excessively abrupt or frequent adversarial attacks. Additionally, alternating periods of adversarial attacks with benign environments ensures that the agent is exposed to both hostile and normal conditions, which can improve generalization. Section \ref{sec:strategies_attack} discusses methods like \mycite{kos2017delving,lin2017tactics,sun2020stealthy,yang2020enhanced} for attack sparification aimed at achieving both stealthiness and efficiency; these techniques can also help balance robustness and performance in adversarial training by adjusting the frequency and timing of the attacks.

        \subsubsection{Regularized Adversarial Training}
            \label{sec:advtrain_regularization}
                     
            To further enhance stability during adversarial training, several approaches introduce regularization mechanisms that mitigate abrupt policy updates under strong adversarial perturbations. 
            
            One notable method, \textbf{Robust Adversarial Training with Langevin Dynamics} (RAT-LD) \mycite{kamalaruban2020robust}, addresses instability and over-conservatism in min–max robust RL by introducing stochasticity into the adversary’s updates. Instead of deterministically seeking the worst-case perturbations, the adversary follows Langevin dynamics—gradient ascent steps perturbed with Gaussian noise—effectively sampling from a smoothed landscape of adversarial disturbances. This stochastic regularization stabilizes the joint optimization between the protagonist and adversary, preventing collapse to overly pessimistic solutions and leading to more balanced robustness in continuous-control tasks. 
            
            Building on this idea of smoothing training dynamics, \textbf{Time-Discounted Robust Training} (TDRT) \mycite{yamabe2024robust} augments the agent’s objective with a temporally weighted KL-divergence regularizer. This time-discounted penalty constrains early trajectory updates—where instability has the largest long-term effect—while allowing more flexibility later in the episode. The adversary remains jointly trainable, but this regularization smooths the protagonist’s learning dynamics, improving convergence and robustness without hindering adaptation.

            Instead of perturbing the MDP during training, which can cause instability from sudden policy changes or out-of-distribution states, \textbf{Worst-Case-Aware Robust RL} (WocaR-RL) \mycite{liang2022efficient} achieves robustness with adversarial regularization in the original MDP. It uses a worst-attack critic network to estimate the policy’s worst-case Q-value under any bounded observation perturbation. Convex-relaxation techniques bound the policy’s outputs within a small neighborhood of each state, and actor–critic updates are performed with this critic as a regularizer, encouraging consistent behavior in that neighborhood. However, because training fully relies on an unperturbed MDP, the behavior distribution remains unchanged, leaving some parts of the state space that the agent might encounter under attack unexplored. As a result, the policy can remain potentially vulnerable in critical situations that lie far from its original state distribution.

            \begin{table}
            \small
            \centering
            \renewcommand{\arraystretch}{1.2}
            \setlength{\tabcolsep}{0.2em}
            \begin{tabular}{|c|c|c|c|}
            \hline
            \textbf{Topic} & \textbf{\shortstack{Category}} & \textbf{Description} \\
            \hline
            
            \multirow{10}{*}{\shortstack{Main Strategy\\\ref{sec:advtrain_strategies}}}
            & \multirow{2}{*}{\shortstack{Constant Training}} & \multirow{2}{*}{\shortstack{Agent learns against a fixed adversary;\\fast but limited.}} \\
            & & \\
            \cdashline{2-3}
            & \multirow{2}{*}{\shortstack{Co-Training}} & \multirow{2}{*}{\shortstack{Agent and adversary learn jointly;\\effective but can be unstable.}} \\
            & & \\
            \cdashline{2-3}
            & \multirow{2}{*}{\shortstack{Self-Mixture}} & \multirow{2}{*}{\shortstack{Single adversary plays a mixture of its own\\historical policies (NFSP-style) to stabilize learning.}} \\
            & & \\
            \cdashline{2-3}
            & \multirow{2}{*}{\shortstack{Mixture or Population}} & \multirow{2}{*}{\shortstack{Train against a pool of adversaries; sample or\\meta-solve a mixture to reduce exploitability.}} \\
            & & \\
            \cdashline{2-3}
            & \multirow{2}{*}{\shortstack{Worst-case Selection}} & \multirow{2}{*}{\shortstack{Pick the hardest environment/adversary each round\\(e.g., $\epsilon$-worst or learned selector) for targeted robustness.}} \\
            & & \\
            \hline

            \multirow{6}{*}{\shortstack{Component\\to Perturb\\\ref{sec:choose_attack}}}
            & \multirow{2}{*}{\shortstack{Pixel-based\\Observations}} & \multirow{2}{*}{\shortstack{Perturb raw images; improves robustness\\to sensor noise, superficial cues.}} \\
            & & \\
            \cdashline{2-3}
            & \multirow{2}{*}{\shortstack{Feature-vector\\Observations}} & \multirow{2}{*}{\shortstack{Perturb abstract features; risks semantic\\corruption but stresses task-critical cues.}} \\
            & & \\
            \cdashline{2-3}
            & \multirow{2}{*}{\shortstack{Environment\\Dynamics}} & \multirow{2}{*}{\shortstack{Modify transitions via actionable levers;\\teaches adaptability to changing physics.}} \\
            & & \\
            \hline
            
            \multirow{7}{*}{\shortstack{Perturbation\\Norm\\\ref{sec:advtrain_magnitude}}}
            & \multirow{2}{*}{\shortstack{Pixels}} & \multirow{2}{*}{\shortstack{$L_\infty$, $L_2$ (sensor-like noise);\\$L_0$ for sparse pixel failures.}} \\
            & & \\
            \cdashline{2-3}
            & \multirow{3}{*}{\shortstack{Features}} & \multirow{3}{*}{\shortstack{Per-features budget if they are heterogeneous. \\ $L_0$ for selective corruption;\\$L_1, L_2, L_\infty$ if all features are similar.}} \\
            & & \\
            & & \\
            \cdashline{2-3}
            & \multirow{2}{*}{\shortstack{Dynamics}} & \multirow{2}{*}{\shortstack{Per-lever budget to reflect physical \\ constraints; optional global $L_p$.}} \\
            & & \\
            \hline
            
            \multirow{5}{*}{\shortstack{Magnitude of\\Perturbation\\\ref{sec:advtrain_magnitude}}}
            & \multirow{2}{*}{\shortstack{Grid Search}} & \multirow{2}{*}{\shortstack{Fixed attack strength chosen from a\\predefined grid; stable but costly.}} \\
            & & \\
            \cdashline{2-3}
            & \multirow{3}{*}{\shortstack{Dynamic Regulation: }} & \multirow{3}{*}{\shortstack{Adapt attack strength according to the \\ agents progress : A2P-SAC, SC-RARL, \\ WR\textsuperscript{2}L, DRAGEN, UED, PAIRED, TC-RMDP.}} \\
            & & \\
            & & \\
            \hline

            \multirow{2}{*}{\shortstack{Frequency \\ \ref{sec:advtrain_frequency}}}
            & \multirow{2}{*}{\shortstack{Low $\to$ High Schedule}} & \multirow{2}{*}{\shortstack{Sparse perturbations to stabilize and prevent overfitting.}} \\
            & & \\
            \hline
                        
            \multirow{4}{*}{\shortstack{Regularization\\\ref{sec:advtrain_regularization}}}
            & \multirow{2}{*}{\shortstack{Smoothing Policy Updates}} & \multirow{2}{*}{\shortstack{Add regularization of policy under attacks to prevent \\ overfitting on strong perturbations : RAT-LD, TDRT.}} \\
            & & \\
            \cdashline{2-3}
            & \multirow{2}{*}{\shortstack{Worst-Case Regularization}} & \multirow{2}{*}{\shortstack{Worst-case value surrogate; regularizes the  policy to be \\ locally consistent within an $\varepsilon$-ball : WocaR-RL.}}\\
            & & \\
            \hline
            
            \end{tabular}
            \caption{Adversarial Training in RL, Summary of design choices and practical guidance.}
            \label{tab:adv_training_summary_multirow}
            \end{table}

\newpage

\section{Tools for Robust Reinforcement Learning}
    \label{sec:tools}
    
    This section presents a range of tools, implementations, and libraries widely used for developing, testing, and analyzing the robustness of RL agents. These tools are organized from the most general libraries to those more specific to adversarial robustness in RL.
    
    \subsection{Libraries for Adversarial Attacks and Robustness}
    
        A variety of libraries have been developed to create adversarial examples, deploy defenses, and benchmark model robustness. While many of these tools were initially designed for supervised learning, several have been adapted or extended to support RL applications. Some libraries offer pre-trained models and standardized evaluation frameworks, providing a foundation for researchers to test models under adversarial attacks. Additionally, RL-specific tools are allowing for comprehensive testing and improvement of RL models.

        \subsubsection{General ML Robustness Libraries}
        
            \begin{itemize}
                \item \textbf{CleverHans} \mycite{papernot2016technical} \href{https://github.com/cleverhans-lab/cleverhans}{github.com/cleverhans-lab/cleverhans} is one of the earliest libraries for adversarial attacks, originally focused on supervised learning. It provides various common gradient attacks, and while it supports RL, it is more lightweight and primarily designed for generating adversarial examples in supervised learning tasks. CleverHans is straightforward and easy to use, making it ideal for quick prototyping and testing adversarial attacks. However, it lacks extensive support for defensive mechanisms and robustness evaluation, and its focus on supervised learning may limit its out-of-the-box application to more complex RL tasks, requiring additional modifications for effective use in RL.
                
                \item \textbf{Adversarial Robustness Toolbox} (ART) \mycite{nicolae2018adversarial}\\ \href{https://github.com/Trusted-AI/adversarial-robustness-toolbox}{github.com/Trusted-AI/adversarial-robustness-toolbox} is a comprehensive library that provides a wide and up-to-date range of adversarial attacks, defenses, and evaluation tools. While ART was primarily developed for supervised learning, it includes functionalities that can be adapted for RL models. ART supports multiple frameworks (e.g., TensorFlow, PyTorch, Keras) and offers gradient attacks as well as some black-box attacks, but it contains no attacks specific of the RL domain. ART’s versatility makes it well-suited for both adversarial attacks and defenses, allowing researchers to perform comprehensive evaluations across different ML domains. However, ART’s wide scope may introduce complexity, especially for users focused solely on RL, as its features are not explicitly tailored for the unique challenges of RL. Adapting its functions for RL may require extra customization.
                
                \item \textbf{TorchAttacks} \mycite{kim2020torchattacks}\\ \href{https://github.com/Harry24k/adversarial-attacks-pytorch}{github.com/Harry24k/adversarial-attacks-pytorch} is a PyTorch-based library that is optimized for adversarial attacks on models implemented in PyTorch. It includes several common gradient attacks, but it is solely focused on adversarial attack generation and does not provide defensive mechanisms or robustness evaluation. TorchAttacks is highly efficient for PyTorch users, offering a simple plug-and-play solution for adding adversarial attacks. However, its lack of support for defenses and robustness evaluation means it is limited to attack generation, and researchers working with RL may need to complement it with other tools to achieve a full evaluation of their models’ robustness.
                
                \item \textbf{RobustBench} \mycite{croce2021robustbench} \href{https://github.com/RobustBench/robustbench}{github.com/RobustBench/robustbench} is a widely used and up-to-date benchmark for adversarial robustness in machine learning. It provides pre-trained models, evaluations, and benchmarks for adversarial robustness, primarily designed for supervised learning. However, it does not offer such resources for RL, though its standardized framework could inspire the development of similar benchmarks for RL.

                \item \textbf{AttackBench} \mycite{cina2025attackbench} \href{https://github.com/attackbench/attackbench}{github.com/attackbench/attackbench} is an open-source benchmarking framework designed to provide fair and reproducible evaluation of gradient-based adversarial attacks on supervised tasks. It standardizes comparisons by enforcing common evaluation protocols and shared computational budgets (e.g., forward/backward passes), integrates implementations from major robustness libraries described above, and supplies a diverse model zoo including both robust and non-robust networks. The tool is intended as a plug-and-play platform for systematically ranking and analyzing attacks, and supports a public leaderboard to encourage transparent and ongoing community benchmarking. AttackBench is limited to supervised learning tasks and does not natively support reinforcement learning. However, its evaluation principles—such as standardized computational budgets and optimality-based ranking—could inspire the design of analogous benchmarks for adversarial attacks in RL.
            \end{itemize}
            
        These well-established tools, such as ART, CleverHans, and TorchAttacks, provide a strong foundation for adversarial research, but they were primarily designed for supervised learning and therefore require adaptation for use in RL.
    
        \subsubsection{RL Specific Robustness Libraries}
            \begin{itemize}
                \item \textbf{Robust Reinforcement Learning Suite} (RRLS) \mycite{zouitine2024rrls}\\
                \href{https://github.com/SuReLI/RRLS}{github.com/SuReLI/RRLS} is a standardized benchmark suite for robust RL, built on MuJoCo environments. RRLS provides continuous control tasks with uncertainty sets and various wrappers (e.g., domain randomization, probabilistic action robustness, adversarial dynamics) to evaluate the robustness of RL agents against adversarial perturbations and environmental uncertainties.
                
                \item \textbf{Robust-Gymnasium} \mycite{gu2025robust}\\
                \href{https://robust-gym.github.io/}{robust-gym.github.io/} is an open-source benchmark suite for robust RL. Its goal is to provide a unified, modular, extensible platform to evaluate how RL algorithms perform under a wide variety of perturbations, uncertainties, and disruptions across observations, actions, reward signals, and environment dynamics. It is designed to enable systematic comparison of robust RL methods or standard RL methods under many different robustness challenges.
            \end{itemize}
            
            RRLS and Robust-Gymnasium are recent libraries dedicated specifically to robustness in RL. Compared to the more established general-purpose robustness libraries in machine learning, they are still relatively young and less mature, with more limited ecosystems and community adoption. This reflects the broader gap in tooling for adversarial training and robustness evaluation in RL, where standardized frameworks are still emerging. Nevertheless, both initiatives represent important and promising steps toward filling this gap, providing tailored benchmarks and experimental settings that could play a central role in advancing robust RL research.

\section{Next Steps Toward Robust RL}
\label{sec:discussion}

    RL agents being vulnerable to alteration in the environments and to adversarial attacks, our taxonomy of these attacks focuses on how they target observations or environment dynamics. Adversarial training emerges as a key strategy to enhance robustness by exposing agents to adversarial conditions during training. However, challenges remain, such as balancing generalization and robustness and managing the computational demands of adversarial defenses.
    
    Now, we discuss key considerations and next steps needed to further advance the development of robust RL techniques, focusing on addressing existing challenges and exploring new avenues for improving performance and security in dynamic environments.

    \subsection{Stability}
    
        Adversarial training often faces significant stability issues, especially when simultaneously training attackers and protagonist agents. While this allows for continuous adaptation of attacks with respect to the  evolving behavior of the agent, thus improving its ability to challenge the agent with stronger or more impactful perturbations, this induces non-stationary dynamics for training, hurting learning stability.  

        These stability and convergence issues in adversarial RL are similar to challenges encountered in other domains like Generative Adversarial Networks (GANs) and Multi-Agent Reinforcement Learning (MARL). Several techniques developed in these fields, which also look at reaching an effective Nash equilibrium between competitive policies,  could be adapted to RL to address these issues effectively.
            
            \paragraph*{GANs}\ \\
                which involve a generator and a discriminator in a competitive setting, face significant stability challenges during training. GAN-like methods for generating sequences in dynamic environments share common challenges with adversarial RL, particularly due to non-stationary reward distributions. This is similar to how adversarial RL agents face the issue of changing access to rewards or feedback based on the adversary's actions, which leads to non-stationary dynamics. Both MALIGAN \mycite{che2017maximum}  and GCN \mycite{lamprier2022generative}, for instance,  address the challenge of stability in sequence generation under dynamic or non-stationary reward structures, which directly parallels adversarial RL scenarios where agents must adapt to changing environments. MALIGAN \mycite{che2017maximum} combines maximum-likelihood estimation with GAN objectives to stabilize training by reducing the variance in updates, which helps in cases where rewards (or sequences) evolve over time. This technique helps the generator to learns effectively, even under fluctuating reward conditions, by using importance sampling and variance reduction. On the other hand, GCN \mycite{lamprier2022generative} introduces a cooperative framework where the generator and discriminator work together, rather than in opposition, to ensure stable learning. In GCN, the discriminator acts as a guide to help the generator adjust to the non-stationary environment, which aligns well with adversarial RL's need to handle dynamic accessibility to rewards. Both methods focus on mitigating the instability caused by evolving conditions, making them particularly relevant for improving stability in adversarial RL environments where adversaries introduce shifts in reward dynamics.

            \paragraph*{MARL}\ \\
                where multiple agents interact in shared environments, they also encounter convergence and stability issues. In adversarial settings, opponent modeling \mycite{foerster2018learning} helps agents predict and counter the strategies of other agents. This concept has been demonstrated in \mycite{gleave2020adversarial} and \mycite{wu2021adversarial}, where adversaries can destabilize and exploit weaknesses in victim policies, even in complex environments like robotics or Go \mycite{wang2022adversarial}. In Adversarial RL, agents could use this to anticipate adversarial attacks and perturbations, reducing the likelihood of oscillatory behavior and improving robustness. A paradigm of centralized learning for decentralized execution is often considered in MARL, which enables more stable optimization and limits non-stationarity by making agents aware of others' decisions during training \mycite{lowe2017multi}. This could serve as a significant source of inspiration for robust adversarial training, with protagonist agents informed by attack plans at training time to converge towards more stable policies that better anticipate unknown perturbations in their final deployment environment. Recently \mycite{standen2025adversarial} provided a large-scale survey of Adversarial Attacks in MARL, classifying attacks and defences, introducing a modelling framework, and identifying research gaps for building resilient multi-agent systems.

    \subsection{Explainability}
    
        Explainable Reinforcement Learning (XRL) enhances robustness by making agents' decision-making processes transparent, enabling easier detection and correction of errors and ensuring reliable performance across diverse conditions. This transparency facilitates better human oversight, allowing interventions and defenses against adversarial attacks. However, a key challenge is balancing explainability with model complexity, as more interpretable models may sacrifice some performance in complex environments. Surveys such as \mycite{qing2022survey,milani2022survey} review XRL methods, while \mycite{sequeira2023ixdrl} presents tools for understanding agent behaviors. \mycite{cheng2023statemask} identifies critical states in decision-making, and \mycite{lee2023adaptive} proposes a framework for adaptive robotic skills. Finally, \mycite{avery2022measuring} links explainability to robustness via Interventional Robustness (IR), a crucial property for generating reliable counterfactual explanations.
        
    \subsection{Human In the Loop}

        Human-in-the-Loop Reinforcement Learning (HILRL) and Reinforcement Learning from Human Feedback (RLHF) significantly enhance robustness and safety in RL systems by incorporating continuous human feedback during training. This real-time human intervention helps detect and correct errors, ensuring agents avoid risky actions while adapting to diverse, unpredictable scenarios. Human feedback refines decision-making, enabling better handling of edge cases, promoting safer behaviors, and improving generalization to avoid overfitting. However, challenges such as inconsistent human feedback, overfitting to specific preferences, and scaling human involvement can hinder achieving full robustness. RLHF, as discussed in \mycite{dai2024safe}, uses human preferences to guide learning, balancing helpfulness and harmlessness in RL agents. By leveraging human feedback, RLHF and HILRL help build robust and dependable systems, as outlined in \mycite{retzlaff2024human,kaufmann2025a}. Furthermore, \mycite{christiano2017deep} emphasizes the importance of shaping policies through human feedback, while \mycite{chen2022human} introduces an algorithm that processes human preferences to refine policy learning.

    \subsection{Large Language Models}
                
        Large Language Models (LLMs) provide valuable insights to RL agents by interpreting natural language instructions and feedback, offering real-time guidance and corrections that enhance learning. This high-level knowledge helps RL agents adapt to diverse scenarios, handle challenges, and improve resilience. LLMs also facilitate communication between humans and agents, improving oversight and intervention for more robust RL systems. Recent works study how LLMs can be used to help RL explore the environment \mycite{du2023guiding}, guiding RL agents towards human-meaningful and contextually relevant behaviors. This requires having \textit{grounded} the common sense of the LLM in the target environment, which is the subject of numerous current research studies \mycite{carta2023grounding}.
        
        Despite their benefits, LLMs are vulnerable to adversarial attacks that exploit weaknesses in their learned representations. Perturbations or manipulations of input data can destabilize RL systems by causing LLMs to generate unintended outputs, as discussed in \mycite{casper2022red}. Moreover, research by \mycite{wang2024reinforcement} shows that automated RL-driven attacks can guide malicious prompt generation to exploit LLM vulnerabilities further. Conversely, LLMs can themselves be turned against RL agents: \mycite{jiang2025policy} use an LLM within a reward-optimization loop to synthesize adversarial rewards and identify the critical states where forcing suboptimal actions most degrades the victim's performance, illustrating how LLMs may serve as a tool for crafting adversarial attacks on RL policies. A key challenge in this context is developing effective defenses against such adversarial manipulations, especially in black-box settings where model internals are inaccessible. Addressing these vulnerabilities is crucial for maintaining the security and reliability of LLM-integrated RL systems.

\section{Conclusion}
\label{sec:conclusion}

    This survey provides an extensive examination of the robustness challenges in RL and the various adversarial training methods aimed at addressing them. Our work highlighted the weaknesses of RL agents to dynamics and observation alterations, underscoring a significant gap in their application in real-world scenarios where reliability and safety are paramount.
    
    We have presented a novel taxonomy of adversarial attacks, categorizing them based on their impact on the dynamics and observations within the RL environment. This classification system not only aids in understanding the nature of these attacks but also serves as a guide for researchers and practitioners in identifying appropriate adversarial training strategies tailored to specific types of vulnerabilities.
    
    Our formalization of the robustness problem in RL, drawing from the principles of distributionally robust optimization for both observation and dynamics alterations, provides a foundational framework for future research. By considering the worst-case scenarios within a controlled uncertainty set, we can develop RL agents that are not only robust to known adversarial attacks but also equipped to handle unexpected variations in real-world environments.
    
    The exploration of adversarial training strategies in this survey emphasizes the importance of simulating realistic adversarial conditions during the training phase. By doing so, RL agents can be better prepared for the complexities and uncertainties of real-world operations, leading to more reliable and effective performance.
    
    While our work sheds light on the current state of adversarial methods in RL and their role in enhancing agent robustness, it also opens the door for further exploration. Future research should focus on refining adversarial training techniques, exploring new forms of attacks, and expanding the taxonomy as the field evolves. Additionally, there is a need to develop more sophisticated models that can balance the trade-off between robustness and performance efficiency. As RL continues to evolve, the pursuit of robust, reliable, and safe autonomous agents remains a critical objective, ensuring their applicability and trustworthiness in a wide range of real-world applications.

\subsection*{Acknowledgements}
    This work has been supported by the French government under the \say{France 2030} program, as part of the SystemX Technological Research Institute within the Confiance.ai program.

\vskip 0.2in
\bibliography{ref}
\bibliographystyle{apalike}

\end{document}